Florentin Smarandache

# Plithogeny, Plithogenic Set, Logic, Probability, and Statistics

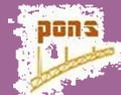

Florentin Smarandache

**Plithogeny,
Plithogenic Set,**
    **Logic,**
        **Probability,**
     **and Statistics**

*In the memory of my beloved parents*
*Maria (born Mitroescu) (1930-2015)*
*and*
*Gheorghe Smarandache (1930-1994)*

*15 January 2017*

Florentin Smarandache

# Plithogeny, Plithogenic Set, Logic, Probability, and Statistics

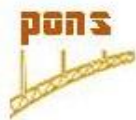

Brussels, Belgium, 2017

# Contents

# Extensions of Fuzzy Set/Logic, Intuitionistic Fuzzy Set/Logic, and Neutrosophic Set/Logic/Probability/Statistics to Plithogenic Set/Logic/Probability/Statistics

## (*Preface*)

We introduce for the first time the concept of *plithogeny* in philosophy and, as a derivative, the concepts of *plithogenic set / logic / probability / statistics* in mathematics and engineering – and the *degrees of contradiction (dissimilarity) between the attributes' values* that contribute to a more accurate construction of *plithogenic aggregation operators* and to the *plithogenic relationship of inclusion* (partial ordering).

They resulted from practical needs in our everyday life, and we present several examples and applications of them.

*Plithogeny* is the genesis or origination, creation, formation, development, and evolution of *new entities* from dynamics and organic fusions of contradictory and/or neutrals and/or non-contradictory *multiple old entities*.

Plithogeny pleads for the connections and unification of theories and ideas in any field.

As "entities" in this study, we take the "knowledges" in various fields, such as soft sciences, hard sciences, arts and letters theories, etc.

Plithogeny is the dynamics of many types of opposites, and/or their neutrals, and/or non-opposites and their organic fusion.

Plithogeny is a generalization of *dialectics* (dynamics of one type of opposites: <A> and <antiA>), *neutrosophy* (dynamics of one type of opposites and their neutrals: <A> and <antiA> and <neutA>), since *plithogeny* studies the dynamics of many types of opposites and their neutrals and non-opposites (<A> and <antiA> and <neutA>, <B> and <antiB> and <neutB>, etc.), and many non-opposites (<C>, <D>, etc.) all together.

As application and particular derivative case of plithogeny is the **Plithogenic Set**, that we'll present in the book, since it is an extension of crisp set, fuzzy set, intuitionistic fuzzy set, and neutrosophic set, and has many scientific applications.

## Plithogenic Set

A *plithogenic set P* is a set whose elements are characterized by one or more attributes, and each attribute may have many values.

Each attribute's value *v* has a corresponding (fuzzy, intuitionistic fuzzy, or neutrosophic) *degree of appurtenance d(x, v)* of the element *x*, to the set *P*, with respect to some given criteria.

In order to obtain a better accuracy for the plithogenic aggregation operators, a (fuzzy, intuitionistic fuzzy, or neutrosophic) *contradiction (dissimilarity) degree* is defined between each attribute value and the dominant (most important) attribute value.

{However, there are cases when such dominant attribute value may not be taking into consideration or may not exist, or there may be many dominant attribute values. In such cases, either the contradiction degree function is suppressed, or another relationship function between attribute values should be designed by the experts according to the application they need to solve.}

The *plithogenic aggregation operators* (intersection, union, complement, inclusion, equality) are linear combinations of the fuzzy operators $t_{norm}$ and $t_{conorm}$.

Plithogenic set is a generalization of the crisp set, fuzzy set, intuitionistic fuzzy set, and neutrosophic set, since these four types of sets are characterized by a single attribute (*appurtenance*): which has one value (membership) – for the crisp set and for fuzzy set, two values (membership, and nonmembership) – for intuitionistic fuzzy set, or three values (membership, nonmembership, and indeterminacy) – for neutrosophic set.

A plithogenic set, in general, may have its elements characterized by attributes with four or more attribute values.

## Plithogenic Logic

A *plithogenic logic proposition P* is a proposition that is characterized by degrees of many truth-values with respect to the corresponding attributes' values.

With respect to each attribute's value *v* there is a corresponding *degree of truth-value d(P, v)* of *P* with respect to the attribute value v.

Plithogenic logic is a generalization of the classical logic, fuzzy logic, intuitionistic fuzzy logic, and neutrosophic logic, since these four types of logics are characterized by a single attribute value (*truth-value*): which has one value (*truth*) – for the classical logic and fuzzy logic, two values (*truth,* and *falsehood*) – for intuitionistic fuzzy logic, or three values (*truth, falsehood,* and *indeterminacy*) – for neutrosophic logic.

A plithogenic logic proposition *P*, in general, may be characterized by more than four degrees of truth-values resulted from under various attributes.

## Plithogenic Probability

In the *plithogenic probability* each event *E* from a probability space *U* is characterized by many chances of the event to occur [not only one chance of the event to occur: as in classical probability, imprecise probability, and neutrosophic probability], chances of occurrence calculated with respect to the corresponding attributes' values that characterize the event *E*.

We present into the book the discrete/continuous finite/infinite n-attribute-values plithogenic probability spaces.

The plithogenic aggregation probabilistic operators (conjunction, disjunction, negation, inclusion, equality) are based on contradiction degrees between attributes' values, and the first two are linear combinations of the fuzzy logical operators' $t_{norm}$ and $t_{conorm}$.

Plithogenic probability is a generalization of the classical probability [ since a single event may have more crisp-probabilities of occurrence ], imprecise probability [ since a single event may have more subunitary subset-probabilities of occurrence ], and neutrosophic probability [ since a single event may have more triplets of: subunitary subset-probabilities of occurrence, subunitary subset-probabilities of indeterminacy (not clear if occurring or not occurring), and subunitary subset-probabilities of nonoccurring) ].

## Plithogenic Statistics

As a generalization of classical statistics and neutrosophic statistics, the *Plithogenic Statistics* is the analysis of events described by the plithogenic probability.

*

Formal definitions of plithogenic set / logic / probability / statistics are presented into the book, followed by plithogenic aggregation operators, theorems, and then examples and applications of these new concepts into our everyday life.

*The Author*

# I. INTRODUCTION TO PLITHOGENY

## I.1. Etymology of Plithogeny

**Plitho-geny** etymologically comes from:
(Gr.) *πλήθος (plithos)* = crowd, large number of, multitude, plenty of,
and
-**geny** < (Gr.) -γενιά *(-geniá)* = generation, the production of something & *γένεια (géneia)* = generations, the production of something < *-γένεση (-génesi)* = genesis, origination, creation, development,
according to Translate Google Dictionaries [ https://translate.google.com/ ] and Webster's New World Dictionary of American English, Third College Edition, Simon & Schuster, Inc., New York, pp. 562-563, 1988.

Therefore, **plithogeny** is the genesis or origination, creation, formation, development, and evolution of *new entities* from dynamics and organic fusions of contradictory and/or neutrals and/or non-contradictory *multiple old entities*.

Plithogeny pleads for the connections and unification of theories and ideas in any field.

As "entities" in this study we take the "knowledges" in various fields, such as soft sciences, hard sciences, arts and letters theories, etc.

Plithogeny is the dynamics of many types of opposites, and/or their neutrals, and/or non-opposites and their organic fusion.

Plithogeny is a generalization of *dialectics* (dynamics of one type of opposites: <A> and <antiA>), *neutrosophy* (dynamics of one type of opposites and their neutrals: <A> and <antiA> and <neutA>), since *plithogeny* studies the dynamics of many types of opposites and their neutrals and non-opposites (<A> and <antiA> and <neutA>, <B> and <antiB> and <neutB>, etc.), and many non-opposites (<C>, <D>, etc.) all together.

### I.1.1. Etymology of Plithogenic

While **plithogenic** means what is pertaining to plithogeny.

## I.2. Plithogenic Multidisciplinarity

While Dialectics is the dynamics of opposites *(<A> and <antiA>)*, Neutrosophy is the dynamic of opposites and their neutrality *(<A>, <neutA>, <antiA>)*, Plithogeny [as a generalization of the above two] is the dynamic and fusion of many opposites and their neutralities, as well as other non-opposites *(<A>, <neutA>, <antiA>; <B>, <neutB>, <antiB>; <C>; <D>;...)*.

Plithogeny organically melts opposites and neutrals and non-opposites entities as a *melting pot* of hybrid or related ideas and concepts.

Plithogeny is multidynamic.

Plithogeny is a MetaScience.

Plithogeny is a hyper-hermeneutics, consisting of hyper-interpretation of literature and arts, soft and hard sciences, and so on. Unconventional research / theory / ideas…

Of course, the dynamics and fusions do not work for all kind of opposites, neutrals, or non-opposites.

Plithogeny is referred to the fluctuation of opposites, neutrals, and non-opposites that all converge towards the same point, then diverge back, and so on. A continuum process of merging, and splitting, or integration and disintegration.

Plithogeny is the basement of plithogenic set, plithogenic logic, plithogenic probability, and plithogenic statistics.

## I.3. Applications of Plithogeny

Of a particular interest is the **plithogenic set**, that we'll present next, since it is an extension of *fuzzy set, intuitionistic fuzzy set,* and *neutrosophic set*, and has many scientific applications.

Similarly plithogenic logic, plithogenic probability and plithogenic statistics.

Let $\oplus$ be a plithogenic aggregation operator [organic fusion].

### *I.3.1. Three entities: Religion (<A>) $\oplus$ Science (<antiA>) $\oplus$ Body (<B>).*

The religious prayer (part of <A>) has been proven be a psychological therapy (hence a scientific method that belongs to <antiA>), which means a non-empty intersection of <A> and <antiA>, or <neutA>, pleading to a spiritual health, which positively influences the physical body *(<B>)*. Therefore, a fusion between <A>, <antiA>, and <B>, or better – a neutrosophic implication: <A> ∩ <antiA> ⇒ <B>.

### I.3.2. Materialism (<A>) ⊕ Idealism (<antiA>) ⊕ Ontology (<B>).

*Materialism* is a philosophical doctrine stating that the only reality is the matter. The emotions, the feelings, and the thoughts can also be explained in terms of matter.

*Idealism* is a philosophical theory that things (reality) exist in our mind only, as ideas; they are dependent of our mind. The things are not material objects.

According to *Neutrosophy*, the reality is the matter (independent of our mind) and our ideas, feelings and thoughts (dependent of our mind). Thus, the combination of *<A>* and *<antiA>* gives *<neutA>*.

*Ontology* is a branch of metaphysics that studies the common traits and principles of being, of reality (ultimate substance). But, *<A>* ∩ *<B>*, which is the <u>Materialist Ontology</u>, studies the structure, determination and development of respective reality. While *<antiA>* ∩ *<B>*, which is <u>Idealist Ontology</u> (or *Gnoseology*), studies the knowledge process: its structure, general conditions, and validity. Whence *<A>* ∩ *<antiA>* ∩ *<B>* studies both the objective and subjective realities of being.

Etymologically, *Gnoseology* < Germ. *Gnoseologie* < Gr. γνώσης *(gnósis)* = knowledge < Gr. γιγνώσκειν *(gignóskein)* = to look with wide open eyes, to goggle, to know.

We introduce now for the first time this philosophical term from German and Romanian languages to English.

*Gnoseology* is the theory of cognition; philosophical theory refereed to the human's capacity of knowing the reality and getting to the truth. { *Dicționarul Explicativ al Limbii Române*, Editura Academiei, Bucharest, p. 377, 1975. }

### I.3.3. Conscious Motivation (<A>) ⊕ Aconscious Motivation (<neutA>) ⊕ Unconscious Motivation (<antiA>) ⊕ Optimism (<B>) ⊕ Pesimism (<antiB>).

All above five concepts [*Conscious Motivation, Aconscious Motivation, Unconscious Motivation, Optimism,* and *Pesimism*] are from the neutrosophic philosophical assumptions and neutrosophic personality traits.

The fusion of these five philosophical entities using plithogeny ends up in assuming: a degree of optimism and a degree of pessimism regarding the motivation at each of the memory's neutrosophic levels: conscious, aconscious, and unconscious.

## I.4. Notations for Crisp, Fuzzy, Intuitionistic Fuzzy, Neutrosophic, and Plithogenic Set/Logic/Probability Operators

We use the notations for intersection, union, complement, less than or equal to, greater than or equal to, and equal to respectively - as follows:

Crisp (Classical): $\wedge, \vee, \neg, \leq, \geq, =$

Fuzzy: $\wedge_F$ ($t_{norm}$), $\vee_F$ ($t_{conorm}$), $\neg_F, \leq_F, \geq_F, =_F$

Refined Fuzzy: $\wedge_{RF}, \vee_{RF}, \neg_{RF}, \leq_{RF}, \geq_{RF}, =_{RF}$

Intuitionistic Fuzzy: $\wedge_{IF}, \vee_{IF}, \neg_{IF}, \leq_{IF}, \geq_{IF}, =_{IF}$

Refined Intuitionistic Fuzzy: $\wedge_{RIF}, \vee_{RIF}, \neg_{RIF}, \leq_{RIF}, \geq_{RIF}, =_{RIF}$

Neutrosophic: $\wedge_N, \vee_N, \neg_N, \leq_N, \geq_N, =_N$

Refined Neutrosophic: $\wedge_{RN}, \vee_{RN}, \neg_{RN}, \leq_{RN}, \geq_{RN}, =_{RN}$

Plithogenic: $\wedge_p, \vee_p, \neg_p, \leq_P, \geq_P, =_P$.

# II. PLITHOGENIC SET

## II.1. Informal Definition of Plithogenic Set

A plithogenic set *P* is a set whose elements are characterized by one or more attributes, and each attribute may have many values.

Each attribute's value *v* h as a corresponding degree of appurtenance *d(x, v)* of the element *x,* to the set *P*, with respect to some given criteria.

In order to obtain a better accuracy for the plithogenic aggregation operators, a contradiction (dissimilarity) degree is defined between each attribute value and the dominant (most important) attribute value.

{However, there are cases when such dominant attribute value may not be taking into consideration or may not exist [therefore it is considered zero by default], or there may be many dominant attribute values. In such cases, either the contradiction degree function is suppressed, or another relationship function between attribute values should be established.}

The plithogenic aggregation operators (intersection, union, complement, inclusion, equality) are based on contradiction degrees between attributes' values, and the first two are linear combinations of the fuzzy operators' $t_{norm}$ and $t_{conorm}$.

Plithogenic set is a generalization of the crisp set, fuzzy set, intuitionistic fuzzy set, and neutrosophic set, since these four types of sets are characterized by a single attribute value (*appurtenance*): which has one value (*membership*) – for the crisp set and fuzzy set, two values (*membership,* and *nonmembership*) – for intuitionistic fuzzy set, or three values (*membership, nonmembership,* and *indeterminacy*) – for neutrosophic set.

A plithogenic set, in general, may have elements characterized by attributes with four or more attribute values.

## II.2. Formal Definition of Single (Uni-Dimensional) Attribute Plithogenic Set

Let $U$ be a universe of discourse, and $P$ a non-empty set of elements, $P \subseteq U$.

### II.2.1. Attribute Value Spectrum

Let $\mathscr{A}$ be a non-empty set of uni-dimensional attributes

$\mathscr{A} = \{\alpha_1, \alpha_2, ..., \alpha_m\}$,

$m \geq 1$; and $\alpha \in \mathscr{A}$ be a given attribute whose spectrum of all possible values (or states) is the non-empty set $S$, where $S$ can be a finite discrete set, $S = \{s_1, s_2, ..., s_l\}$, $1 \leq l < \infty$, or infinitely countable set $S = \{s_1, s_2, ..., s_\infty\}$, or infinitely uncountable (continuum) set $S = ]a, b[$, $a < b$, where $]...[$ is any open, semi-open, or closed interval from the set of real numbers or from other general set.

### II.2.2. Attribute Value Range

Let $V$ be a non-empty subset of $S$, where $V$ is the *range of all attribute's values* needed by the experts for their application. Each element $x \in P$ is characterized by all attribute's values in $V = \{v_1, v_2, ..., v_n\}$, for $n \geq 1$.

## II.2.3. Dominant Attribute Value

Into the attribute's value set *V*, in general, there is a <u>dominant attribute value</u>, which is determined by the experts upon their application. Dominant attribute value means the most important attribute value that the experts are interested in.

{However, there are cases when such dominant attribute value may not be taking into consideration or not exist, or there may be many dominant (important) attribute values - when different approach should be employed.}

## II.2.4. Attribute Value Appurtenance Degree Function

Each attributes value $v \in V$ has a corresponding degree of appurtenance $d(x, v)$ of the element *x,* to the set *P*, with respect to some given criteria.

The degree of appurtenance may be: a fuzzy degree of appurtenance, or intuitionistic fuzzy degree of appurtenance, or neutrosophic degree of appurtenance to the plithogenic set.

Therefore, the attribute value appurtenance degree function is:

$$\forall x \in P, d: P \times V \rightarrow \mathcal{P}([0, 1]^z), \qquad (1)$$

so $d(x, v)$ is a subset of $[0, 1]^z$, where $\mathcal{P}([0, 1]^z)$ is the power set of the $[0, 1]^z$, where $z = 1$ (for fuzzy degree of appurtenance), $z = 2$ (for intuitionistic fuzzy degree of appurtenance), or $z = 3$ (for neutrosophic degree de appurtenance).

## II.2.5. Attribute Value Contradiction (Dissimilarity) Degree Function

Let the cardinal $|V| \geq 1$.

Let $c: V \times V \to [0, 1]$ be the <u>attribute value contradiction degree function</u> (that we introduce now for the first time) between any two attribute values $v_1$ and $v_2$, denoted by

$c(v_1, v_2)$, and satisfying the following axioms:

$c(v_1, v_1) = 0$, the contradiction degree between the same attribute values is zero;

$c(v_1, v_2) = c(v_2, v_1)$, commutativity.

For simplicity, we use a <u>fuzzy attribute value contradiction degree function</u> ($c$ as above, that we may denote by $c_F$ in order to distinguish it from the next two), but an <u>intuitionistic attribute value contradiction function</u> ($c_{IF} : V \times V \to [0, 1]^2$), or more general a <u>neutrosophic attribute value contradiction function</u> ($c_N : V \times V \to [0, 1]^3$) may be utilized increasing the complexity of calculation but the accuracy as well.

We mostly compute the contradiction degree between *uni-dimensional attribute values*. For *multi-dimensional attribute values* we split them into corresponding uni-dimensional attribute values.

The attribute value contradiction degree function helps the plithogenic aggregation operators, and the plithogenic inclusion (partial order) relationship to obtain a more accurate result.

The attribute value contradiction degree function is designed in each field where plithogenic set is used in accordance with the application to solve. If it is ignored, the aggregations still work, but the result may lose accuracy.

Several examples will be provided into this book.

Then $(P, a, V, d, c)$ is called a plithogenic set:

● where "*P*" is a set, "*a*" is a (multi-dimensional in general) attribute, "*V*" is the range of the attribute's values, "*d*" is the degree of appurtenance of each element *x*'s attribute value to the set *P* with respect to some given criteria ($x \in P$), and "*d*" stands for "$d_F$" or "$d_{IF}$" or "$d_N$", when dealing with fuzzy degree of appurtenance, intuitionistic fuzzy degree of appurtenance, or neutrosophic degree of appurtenance respectively of an element *x* to the plithogenic set *P*;

● and "*c*" stands for "$c_F$" or "$c_{IF}$" or "$c_N$", when dealing with fuzzy degree of contradiction, intuitionistic fuzzy degree of contradiction, or neutrosophic degree of contradiction between attribute values respectively.

The functions $d(\cdot,\cdot)$ and $c(\cdot,\cdot)$ are defined in accordance with the applications the experts need to solve.

One uses the notation:
$$x(d(x,V)),$$
where $d(x,V) = \{d(x,v), \text{for all } v \in V\}, \forall x \in P$.

## II.2.6. About the Plithogenic Aggregation Set Operators

The attribute value contradiction degree is calculated between each attribute value with respect to the dominant attribute value (denoted $v_D$) in special, and with respect to other attribute values as well.

The attribute value contradiction degree function *c* between the attribute's values is used into the definition of plithogenic aggregation operators {*Intersection (AND), Union (OR), Implication ( $\Rightarrow$ ), Equivalence ( $\Leftrightarrow$ ), Inclusion Relationship (Partial Order),* and other plithogenic aggregation operators that combine two or more attribute value degrees - that *t_norm* and *t_conorm* act upon}.

Most of the plithogenic aggregation operators are linear combinations of the fuzzy $t_{norm}$ (denoted $\wedge_F$), and fuzzy $t_{conorm}$ (denoted $\vee_F$), but non-linear combinations may as well be constructed.

If one applies the $t_{norm}$ on dominant attribute value denoted by $v_D$, and the contradiction between $v_D$ and $v_2$ is $c(v_D, v_2)$, then onto attribute value $v_2$ one applies:

$[1 - c(v_D, v_2)] \cdot t_{norm}(v_D, v_2) + c(v_D, v_2) \cdot t_{conorm}(v_D, v_2)$, (2)

Or, by using symbols:

$[1 - c(v_D, v_2)] \cdot (v_D \wedge_F v_2) + c(v_D, v_2) \cdot (v_D \vee_F v_2)$. (3)

Similarly, if one applies the $t_{conorm}$ on dominant attribute value denoted by $v_D$, and the contradiction between $v_D$ and $v_2$ is $c(v_D, v_2)$, then onto attribute value $v_2$ one applies:

$[1 - c(v_D, v_2)] \cdot t_{conorm}(v_D, v_2) + c(v_D, v_2) \cdot t_{norm}(v_D, v_2)$, (4)

Or, by using symbols:

$[1 - c(v_D, v_2)] \cdot (v_D \vee_F v_2) + c(v_D, v_2) \cdot (v_D \wedge_F v_2)$. (5)

## II.3. Plithogenic Set as Generalization of other Sets

The plithogenic set is an extension of all: crisp set, fuzzy set, intuitionistic fuzzy set, and neutrosophic set.

For examples:

Let $U$ be a universe of discourse, and a non-empty set $P \subseteq U$. Let $x \in P$ be a generic element.

### II.3.1. Crisp (Classical) Set (CCS)

The *attribute* is α = "*appurtenance*";
the *set of attribute values* V = *{membership, nonmembership}*, with cardinal $|V| = 2$;
the *dominant attribute value* = *membership*;

the *attribute value appurtenance degree function*:

$$d: P \times V \to \{0, 1\}, \quad (6)$$

$d(x, membership) = 1, \ d(x, nonmembership) = 0,$

and the *attribute value contradiction degree function*:

$$c: V \times V \to \{0, 1\}, \quad (7)$$

*c(membership, membership) = c(nonmembership, nonmembership) = 0,*

*c(membership, nonmembership) = 1.*

### II.3.1.1. Crisp (Classical) Intersection

$$a \wedge b \in \{0, 1\} \quad (8)$$

### II.3.1.2. Crisp (Classical) Union

$$a \vee b \in \{0, 1\} \quad (9)$$

### II.3.1.3. Crisp (Classical) Complement (Negation)

$$\neg a \in \{0, 1\}. \quad (10)$$

## II.3.2. Single-Valued Fuzzy Set (SVFS)

The *attribute* is α = "*appurtenance*";

the *set of attribute values V = {membership}*, whose cardinal $|V| = 1$;

the *dominant attribute value = membership;*

the *appurtenance attribute value degree function:*

$$d: P \times V \to [0, 1], \quad (11)$$

with $d(x, membership) \in [0, 1]$;

and the *attribute value contradiction degree function*:

$$c: V \times V \to [0, 1], \quad (12)$$

*c(membership, membership) = 0.*

### II.3.2.1. Fuzzy Intersection

$$a \wedge_F b \in [0, 1] \quad (13)$$

### II.3.2.2. Fuzzy Union

$$a \vee_F b \in [0, 1] \tag{14}$$

### II.3.2.3. Fuzzy Complement (Negation)

$$\neg_F a = 1 - a \in [0, 1]. \tag{15}$$

## II.3.3. Single-Valued Intuitionistic Fuzzy Set (SVIFS)

The *attribute* is α = "*appurtenance*";
the *set of attribute values* V = *{membership, nonmembership}*, whose cardinal |V| = 2;
the *dominant attribute value* = *membership*;
the *appurtenance attribute value degree function*:
$$d: P \times V \to [0, 1], \tag{16}$$
$d(x, membership) \in [0, 1]$, $d(x, nonmembership) \in [0, 1]$,
with $d(x, membership) + d(x, nonmembership) \leq 1$,
and the *attribute value contradiction degree function*:
$$c: V \times V \to [0, 1], \tag{17}$$
$c(membership, membership) = c(nonmembership, nonmembership) = 0$,
$c(membership, nonmembership) = 1$,

which means that for SVIFS aggregation operators' intersection (AND) and union (OR), if one applies the $t_{norm}$ on membership degree, then one has to apply the $t_{conorm}$ on nonmembership degree – and reciprocally.

Therefore:

### II.3.3.1. Intuitionistic Fuzzy Intersection

$$(a_1, a_2) \wedge_{\text{IFS}} (b_1, b_2) = (a_1 \wedge_F b_1, a_2 \vee_F b_2) \tag{18}$$

### II.3.3.2. Intuitionistic Fuzzy Union

$$(a_1, a_2) \vee_{\text{IFS}} (b_1, b_2) = (a_1 \vee_F b_1, a_2 \wedge_F b_2), \tag{19}$$

and

### II.3.3.3. Intuitionistic Fuzzy Complement (Negation)

$$\neg_{IFS}(a_1, a_2) = (a_2, a_1). \tag{20}$$

where $\wedge_F$ and $\vee_F$ are the fuzzy $t_{norm}$ and fuzzy $t_{conorm}$ respectively.

### II.3.3.4. Intuitionistic Fuzzy Inclusions (Partial Orders)

*II.3.3.4.1. Simple Intuitionistic Fuzzy Inclusion (the most used by the intuitionistic fuzzy community):*

$$(a_1, a_2) \leq_{IFS} (b_1, b_2) \tag{21}$$

iff $a_1 \leq b_1$ and $a_2 \geq b_2$.

*II.3.3.4.2. Plithogenic (Complete) Intuitionistic Fuzzy Inclusion (that we now introduce for the first time):*

$$(a_1, a_2) \leq_P (b_1, b_2) \tag{22}$$

iff $a_1 \leq (1-c_v) \cdot b_1, a_2 \geq (1-c_v) \cdot b_2$,

where $c_v \in [0, 0.5)$ is the contradiction degree between the attribute dominant value and the attribute value $v$ { the last one whose degree of appurtenance with respect to Expert A is *(a₁, a₂)*, while with respect to Expert B is *(b₁, b₂)* }. If $c_v$ does not exist, we take it by default as equal to zero.

## II.3.4. Single-Valued Neutrosophic Set (SVNS)

The *attribute* is α = "*appurtenance*";
the *set of attribute values V = {membership, indeterminacy, nonmembership}*, whose cardinal |V| = 3;
the *dominant attribute value* = *membership*;
the *attribute value appurtenance degree function*:
$$d: P \times V \to [0, 1], \tag{23}$$
*d(x, membership)* ∈ [0, 1], *d(x, indeterminacy)* ∈ [0, 1],
*d(x, nonmembership)* ∈ [0, 1],

with $0 \leq d(x, membership) + d(x, indeterminacy) + d(x, nonmembership) \leq 3$;

and the *attribute value contradiction degree function*:

$c: V \times V \rightarrow [0, 1]$, (24)

$c(membership, membership) = c(indeterminacy, indeterminacy) = c(nonmembership, nonmembership) = 0$,

$c(membership, nonmembership) = 1$,

$c(membership, indeterminacy) = c(nonmembership, indeterminacy) = 0.5$,

which means that for the SVNS aggregation operators (Intersection, Union, Complement etc.), if one applies the $t_{norm}$ on membership, then one has to apply the $t_{conorm}$ on nonmembership {and reciprocally}, while on indeterminacy one applies the average of $t_{norm}$ and $t_{conorm}$, as follows:

### II.3.4.1. Neutrosophic Intersection:

*II.3.4.1.1. Simple Neutrosophic Intersection (the most used by the neutrosophic community):*

$$(a_1, a_2, a_3) \wedge_{NS} (b_1, b_2, b_3) = (a_1 \wedge_F b_1, a_2 \vee_F b_2, a_3 \vee_F b_3) \quad (25)$$

*II.3.4.1.2. Plithogenic Neutrosophic Intersection*

$$(a_1, a_2, a_3) \wedge_P (b_1, b_2, b_3) =$$

$$\left( a_1 \wedge_F b_1, \frac{1}{2}\left[ (a_2 \wedge_F b_2) + (a_2 \vee_F b_2) \right], a_3 \vee_F b_3 \right) \quad (26)$$

### II.3.4.2. Neutrosophic Union:

*II.3.4.2.1. Simple Neutrosophic Union (the most used by the neutrosophic community):*

$$(a_1, a_2, a_3) \vee_{NS} (b_1, b_2, b_3) =$$

$$(a_1 \vee_F b_1, a_2 \wedge_F b_2, a_3 \wedge_F b_3) \quad (27)$$

*II.3.4.2.2. Plithogenic Neutrosophic Union*

$(a_1, a_2, a_3) \vee_P (b_1, b_2, b_3)$

$$= \left( a_1 \vee_F b_1, \frac{1}{2}\left[(a_2 \wedge_F b_2) + (a_2 \vee_F b_2)\right], a_3 \wedge_F b_3 \right). \quad (28)$$

In other way, with respect to what one applies on the membership, one applies the opposite on non-membership, while on indeterminacy one applies the average between them.

### *II.3.4.3. Neutrosophic Complement (Negation):*

$$\neg_{NS} (a_1, a_2, a_3) = (a_3, a_2, a_1). \quad (29)$$

### *II.3.4.4. Neutrosophic Inclusions (Partial-Orders)*

*II.3.4.4.1. Simple Neutrosophic Inclusion (the most used by the neutrosophic community):*

$(a_1, a_2, a_3) \leq_{NS} (b_1, b_2, b_3)$     (30)

iff *$a_1 \leq b_1$ and $a_2 \geq b_2$, $a_3 \geq b_3$.*

*II.3.4.4.2. Plithogenic Neutrosophic Inclusion (defined now for the first time):*

Since the degrees of contradiction are

*$c(a_1, a_2) = c(a_2, a_3) = c(b_1, b_2) = c(b_2, b_3) = 0.5$,*     (31)

one applies: *$a_2 \geq [1- c(a_1, a_2)]b_2$ or $a_2 \geq (1-0.5)b_2$ or $a_2 \geq 0.5 \cdot b_2$* while

*$c(a_1, a_3) = c(b_1, b_3) = 1$*     (32)

{having *$a_1 \leq b_1$* one does the opposite for *$a_3 \geq b_3$*},

whence

*$(a_1, a_2, a_3) \leq_P (b_1, b_2, b_3)$*     (33)

iff *$a_1 \leq b_1$ and $a_2 \geq 0.5 \cdot b_2$, $a_3 \geq b_3$.*

## II.3.5. Single-Valued Refined Fuzzy Set (SVRFS)

For the first time the fuzzy set was refined by Smarandache [2] in 2016 as follows:

A SVRFS number has the form:

$(T_1, T_2, ..., T_p)$,

where $p \geq 2$ is an integer, and all $T_j \in [0, 1]$, for $j \in \{1, 2, ..., p\}$.

The *attribute* α = "*appurtenance*";

the *set of attribute values* $V = \{m_1, m_2, ..., m_p\}$, where "m" means *submembership*;

the *dominant attribute values* = $m_1, m_2, ..., m_p$;

the *attribute value appurtenance degree function*:

$d: P \times V \rightarrow [0, 1]$, (34)

$d(x, m_j) \in [0, 1]$, for all $j$,

and $\sum_{j=1}^{p} d_x(m_j) \leq 1$ ; (35)

and the *attribute value contradiction degree function*:

$c(m_{j_1}, m_{j_2}) = 0$, (36)

for all $j_1, j_2 \in \{1, 2, ..., p\}$.

Aggregation operators on SVRFS:

Let $(a_j, 1 \leq j \leq p)$, with all $a_j \in [0, 1]$, be a SVRFS number, which means that the sub-truths $T_j = a_j$ for all $1 \leq j \leq p$.

### II.3.5.1. Refined Fuzzy Intersection:

$(a_j, 1 \leq j \leq p) \wedge_{RFS} (b_j, 1 \leq j \leq p)$

$= (a_j \wedge_F b_j, 1 \leq j \leq p)$. (37)

### II.3.5.2. Refined Fuzzy Union:

$(a_j, 1 \leq j \leq p) \vee_{RFS} (b_j, 1 \leq j \leq p)$

$= (a_j \vee_F b_j, 1 \leq j \leq p)$. (38)

### II.3.5.3. Refined Fuzzy Complement (Negation):

$$\neg_{RFS} (T_j = a_j, 1 \leq j \leq p) = (F_j = a_j, 1 \leq j \leq p), \quad (39)$$

where $F_j$ are the sub-falsehoods, for all $1 \leq j \leq p$.

### II.3.5.4. Refined Fuzzy Inclusion (Partial-Order)

$$(a_j, 1 \leq j \leq p) \leq_{RFS} (b_j, 1 \leq j \leq p) \quad (40)$$

iff $a_j \leq b_j$ for all $1 \leq j \leq p$.

## II.3.6. Single-Valued Refined Intuitionistic Fuzzy Set (SVRIFS)

For the first time, the intuitionistic fuzzy set was refined by Smarandache [2] in 2016, as follows:

A SVRIFS number has the form:

*($T_1$, $T_2$, …, $T_p$; $F_1$, $F_2$, …, $F_s$),*

where $p, r \geq 1$ are integers, and $p + r \geq 3$, and all $T_j, F_l \in [0, 1]$, for $j \in \{1, 2, …, p\}$ and $l \in \{1, 2, …, s\}$.

The *attribute* α = "*appurtenance*";

the *set of attribute values* $V = \{m_1, m_2, …, m_p; nm_1, nm_2, …, nm_p\}$, where "m" means *submembership*, and "nm" *subnonmembership*;

the *dominant attribute values* = $m_1, m_2, …, m_p$;

the *attribute value appurtenance degree function*:

$d: P \times V \to [0, 1],$ \quad (41)

$d(x, m_j) \in [0, 1]$, for all $j$, and $d(x, nm_l) \in [0, 1]$, for all $l$, where

$$\sum_{j=1}^{p} d_x(m_j) + \sum_{l=1}^{s} d_x(nm_l) \leq 1; \quad (42)$$

and the *attribute value contradiction degree function*:

$$c(m_{j_1}, m_{j_2}) = c(nm_{l_1}, nm_{l_2}) = 0, \quad (43)$$

for all $j_1, j_2 \in \{1, 2, …, p\}$, and $l_1, l_2 \in \{1, 2, …, s\}$,

while $c(m_j, nm_l) = 1$ for all $j$ and $l$.

Aggregation operators on SVRIFS:

### II.3.6.1. Refined Intuitionistic Set Intersection:

$$(a_j, 1 \leq j \leq p; b_l, 1 \leq l \leq s) \wedge_{RIFS}$$
$$(c_j, 1 \leq j \leq p; d_l, 1 \leq l \leq s) =$$
$$(a_j \wedge_F c_j, 1 \leq j \leq p; b_l \vee_F d_l, 1 \leq l \leq s). \tag{44}$$

### II.3.6.2. Refined Intuitionistic Set Union

$$(a_j, 1 \leq j \leq p; b_l, 1 \leq l \leq s) \vee_{RIFS}$$
$$(c_j, 1 \leq j \leq p; d_l, 1 \leq l \leq s) =$$
$$(a_j \vee_F c_j, 1 \leq j \leq p; b_l \wedge_F d_l, 1 \leq l \leq s). \tag{45}$$

### II.3.6.3. Refined Intuitionistic Complement (Negation)

$$\neg_{RIFS}(T_j = a_j, 1 \leq j \leq p; F_j = b_l, 1 \leq l \leq s) =$$
$$(T_j = b_l, 1 \leq l \leq s; F_j = a_j, 1 \leq j \leq p). \tag{46}$$

### II.3.6.4. Refined Intuitionistic Inclusions (Partial Orders)

#### II.3.6.4.1. Simple Refined Intuitionistic Inclusion

$$(a_j, 1 \leq j \leq p; b_l, 1 \leq l \leq s) \leq_{RIFS}$$
$$(u_j, 1 \leq j \leq p; w_l, 1 \leq l \leq s) \tag{47}$$

iff

$a_j \leq u_j$ for all $1 \leq j \leq p$ and $w_l \geq d_l$ for all $1 \leq l \leq s$.

#### II.3.6.4.2. Plithogenic Refined Intuitionistic Inclusion

$$(a_j, 1 \leq j \leq p; b_l, 1 \leq l \leq s) \leq_P$$
$$(u_j, 1 \leq j \leq p; w_l, 1 \leq l \leq s) \tag{48}$$

iff $a_j \leq (1 - c_v) \cdot u_j$ for all $1 \leq j \leq p$ and $b_l \geq (1 - c_v) \cdot w_l$ for all $1 \leq l \leq s$,

where similarly $c_v \in [0, 0.5)$ is the contradiction degree between the attribute dominant value and the attribute value $v$. If $c_v$ does not exist, we take it by default as equal to zero.

## II.3.7. Single-Valued Finitely Refined Neutrosophic Set (SVFRNS)

The Single-Valued Refined Neutrosophic Set and Logic were first defined by Smarandache [3] in 2013.

A SVFRNS number has the form:

$(T_1, T_2, ..., T_p; I_1, I_2, ..., I_r; F_1, F_2, ..., F_s)$,

where $p, r, s \geq 1$ are integers, with $p + r + s \geq 4$,

and all $T_j$, $I_k$, $F_l \in [0, 1]$, for $j \in \{1, 2, ..., p\}$, $k \in \{1, 2, ..., r\}$, and $l \in \{1, 2, ..., s\}$.

The *attribute* α = "*appurtenance*";

the *set of attribute values* $V = \{m_1, m_2, ..., m_p; i_1, i_2, ..., i_r; f_1, f_2, ..., f_s\}$, where "*m*" means submembership, "*i*" subindeterminacy, and "*f*" sub-nonmembership;

the *dominant attribute values* = $m_1, m_2, ..., m_p$;

the *attribute value appurtenance degree function*:

$d: P \times V \to [0, 1]$, $\hspace{2cm}$ (49)

$d(x, m_j) \in [0, 1]$, $d(x, i_k) \in [0, 1]$, $d(x, f_l) \in [0, 1]$, for all $j, k, l$, with

$$0 \leq \sum_{j=1}^{p} d_x(m_j) + \sum_{k=1}^{r} d_x(i_k) + \sum_{l=1}^{s} d_x(f_l) \leq p + r + s;$$
(50)

and the *attribute value contradiction degree function*:

$c(m_{j_1}, m_{j_2}) = c(i_{k_1}, i_{k_2}) = c(f_{l_1}, f_{l_2}) = 0$, $\hspace{1cm}$ (51)

for all $j_1, j_2 \in \{1, 2, ...., p\}$, $k_1, k_2 \in \{1, 2, ...., r\}$, and $l_1, l_2 \in \{1, 2, ...., s\}$;

$c(m_j, f_l) = 1$, $\hspace{5cm}$ (52)

$$c(m_j, i_k) = c(f_l, i_k) = 0.5, \qquad (53)$$

for all $j$, $k$, $l$.

Aggregation operators on SVFRNS:

### II.3.7.1. Refined Neutrosophic Set Intersection:

$$\left(a_j, 1 \leq j \leq p; b_k, 1 \leq k \leq r; g_l, 1 \leq l \leq s\right) \wedge_{RNS}$$
$$\left(u_j, 1 \leq j \leq p; o_k, 1 \leq k \leq r; w_l, 1 \leq l \leq s\right) =$$
$$\left(a_j \wedge_F u_j, 1 \leq j \leq p; \frac{1}{2}\left[(b_k \wedge_F o_k) + (b_k \vee_F o_k)\right], g_l \vee_F w_l, \atop 1 \leq l \leq s\right); \qquad (54)$$

and

### II.3.7.2. Refined Neutrosophic Set Union:

$$\left(a_j, 1 \leq j \leq p; b_k, 1 \leq k \leq r; g_l, 1 \leq l \leq s\right) \vee_{RNS}$$
$$\left(u_j, 1 \leq j \leq p; o_k, 1 \leq k \leq r; w_l, 1 \leq l \leq s\right) =$$
$$\left(a_j \vee_F u_j, 1 \leq j \leq p; \frac{1}{2}\left[(b_k \wedge_F o_k) + (b_k \vee_F o_k)\right], \atop 1 \leq k \leq r; g_l \wedge_F w_l, 1 \leq l \leq s\right). \qquad (55)$$

### II.3.7.3. Refined Neutrosophic Complement (Negation):

$$\neg_{NS} \begin{pmatrix} T_j = a_j, 1 \leq j \leq p; I_k = b_k, \\ 1 \leq k \leq r; F_l = g_l, 1 \leq l \leq s \end{pmatrix} =$$
$$= \begin{pmatrix} T_j = g_l, 1 \leq l \leq s; I_k = b_k, 1 \leq k \leq r; \\ g_l, 1 \leq l \leq s; F_l = a_j, 1 \leq j \leq p \end{pmatrix}, \qquad (56)$$

where all $T_j$ = sub-truths, all $I_k$ = sub-indeterminacies, and all $F_l$ = sub-falsehoods.

### II.3.7.4. Refined Neutrosophic Inclusions (Partial-Orders):
### II.3.7.4.1. Simple Refined Neutrosophic Inclusion
$$(a_j, 1 \leq j \leq p; b_k, 1 \leq k \leq r; g_l, 1 \leq l \leq s) \leq_{RNS}$$
$$(u_j, 1 \leq j \leq p; o_k, 1 \leq k \leq r; w_l, 1 \leq l \leq s) \quad (57)$$
iff all $a_j \leq u_j$, all $b_k \geq o_k$ and all $g_l \geq w_l$.

### II.3.7.4.2. Plithogenic Refined Neutrosophic Inclusion
$$\begin{aligned}&(a_j, 1 \leq j \leq p; b_k, 1 \leq k \leq r; g_l, 1 \leq l \leq s) \\ &\leq_P (u_j, 1 \leq j \leq p; o_k, 1 \leq k \leq r; w_l, 1 \leq l \leq s)\end{aligned} \quad (58)$$

iff all $a_j \leq (1-c_v) \cdot u_j$, all $b_k \geq (1-c_v) \cdot o_k$ and all $g_l \geq (1-c_v) \cdot w_l$, where $c_v \in [0, 0.5)$ is the contradiction degree between the attribute dominant value and the attribute value $v$. If $c_v$ does not exist, we take it by default as equal to zero.

## II.4. One-Attribute-Value Plithogenic Single-Valued Set Operators

If onto the dominant attribute value $v_D$ one applies the plithogenic t$_{norm}$, then on an attribute value $v_1$ whose contradiction degree with respect to $v_D$ is 1, one applies the opposite, i.e. the plithogenic t$_{conorm}$.

While onto an attribute value $v_2$ whose contradiction degree with respect to $v_D$ belongs to (0, 1), one applies a linear combination of the t$_{norm}$ and t$_{conorm}$:

$$\alpha \cdot \text{t}_{\text{norm}}[d_A(v_2), d_B(v_2)] + \beta \cdot \text{t}_{\text{conorm}}[d_A(v_2), d_B(v_2)], \quad (59)$$
with $\alpha, \beta \in (0, 1)$, and $\alpha + \beta = 1$.

When doing a *plithogenic intersection*: the closer is $c(v_D, v_2)$ to 0, the larger is the percentage of t$_{norm}$ added and the smaller is the percentage of t$_{conorm}$ added.

And reciprocally, when doing a *plithogenic union*: the closer is $c(v_D, v_2)$ to 0, the smaller is the percentage of t$_{norm}$ added and the bigger is the percentage of t$_{conorm}$ added.

If $c(v_D, v_2) = \frac{1}{2}$, then the plithogenic intersection coincides with the plithogenic union:
$$d_A(v_2) \wedge_p d_B(v_2) =$$
$$\frac{1}{2} \cdot [d_A(v_2) \wedge_F d_B(v_2)] + \frac{1}{2} \cdot [d_A(v_2) \vee_F d_B(v_2)], \quad (60)$$
while
$$d_A(v_2) \vee_p d_B(v_2) =$$
$$\frac{1}{2} \cdot [d_A(v_2) \vee_F d_B(v_2)] + \frac{1}{2} \cdot [d_A(v_2) \wedge_F d_B(v_2)]. \quad (61)$$

If onto $v_D$ one applies $\wedge_p$, then on all $v$'s with $c(v_D, v) < 0.5$ one also applies $\wedge_p$, while on those $v$'s with $c(v_D, v) \geq 0.5$ one applies the opposite ($\vee_p$).

And reciprocally: if on $v_D$ one applies $\vee_p$, then on all $v$'s with $c(v_D, v) < 0.5$ one also applies $\vee_p$, while on those $v$'s with $c(v_D, v) \geq 0.5$ one applies the opposite ($\wedge_p$).

## II.4.1. One-Attribute-Value Plithogenic Single-Valued Fuzzy Set Operators

Let $U$ be a universe of discourse, and a subset of it $P$ be a plithogenic set, and $x \in P$ an element. Let α be a uni-dimensional attribute that characterize *x*, and *v* an attribute value, $v \in V$, where $V$ is set of all attribute's α values used into solving an application.

The degree of contradiction $c(v_D, v) = c_0 \in [0, 1]$ between the dominant attribute value $v_D$ and the attribute value $v$.

Let's consider two experts, A and B, each evaluating the single-valued fuzzy degree of appurtenance of attribute value $v$ of $x$ to the set $P$ with respect to some given criteria:

$d_A^F(v) = a \in [0, 1]$, and
$d_B^F(v) = b \in [0, 1]$.

Let $\wedge_F$ and $\vee_F$ be a fuzzy t$_{norm}$ and respectively fuzzy t$_{conorm}$.

## II.4.2. One-Attribute-Value Plithogenic Single-Valued Fuzzy Set Intersection

$$a \wedge_p b = (1 - c_0) \cdot [a \wedge_F b] + c_0 \cdot [a \vee_F b]. \quad (62)$$

If $c(v_D, v) = c_0 \in [0, 0.5)$ then more weight is assigned onto the t$_{norm}$(a, b) = $a \wedge_F b$ than onto t$_{conorm}$(a,b) = $a \vee_F b$; this is a proper plithogenic intersection.

If $c(v_D, v) = c_0 \in (0.5, 1]$ then less weight is assigned onto the t$_{norm}$(a, b) = $a \wedge_F b$ than onto t$_{conorm}$(a,b) = $a \vee_F b$; this becomes (rather) an improper plithogenic union.

If $c(v_D, v) = c_0 \in 0.5$ then the same weight {0.5} is assigned onto the

t$_{norm}$(a, b) = $a \wedge_F b$ and on t$_{conorm}$(a,b) = $a \vee_F b$.

## II.4.3. One-Attribute-Value Plithogenic Single-Valued Fuzzy Set Union

$$a \vee_p b = (1 - c_0) \cdot [a \vee_F b] + c_0 \cdot [a \wedge_F b]. \quad (63)$$

If $c(v_D, v) = c_0 \in [0, 0.5)$ then more weight is assigned onto the t$_{conorm}$(a, b) = $a \vee_F b$ than onto t$_{norm}$(a,b) = $a \wedge_F b$; this is a proper plithogenic union.

If $c(v_D, v) = c_0 \in (0.5, 1]$ then less weight is assigned onto the t<sub>conorm</sub>(a, b) = $a \vee_F b$ than onto t<sub>norm</sub>(a,b) = $a \wedge_F b$; this is (rather) an improper plithogenic intersection.

If $c(v_D, v) = c_0 \in 0.5$ then the same weight {0.5} is assigned onto the

t<sub>conorm</sub>(a, b) = $a \wedge_F b$ and on t<sub>norm</sub>(a,b) = $a \vee_F b$.

## II.4.4. One-Attribute-Value Plithogenic Single-Valued Fuzzy Set Complements (Negations)

### II.4.4.1. Denying the Attribute Value

$$\neg_p(v) = anti(v), \qquad (64)$$

i.e. the opposite of $v$, where $anti(v) \in V$ or $anti(v) \in RefinedV$ (refined set of $V$).

So, we get:
$$d_A^F(anti(v)) = a. \qquad (65)$$

### II.4.4.2. Denying the Attribute Value Degree

$$\neg_p(a) = 1 - a, \text{ or } \neg_p d_A^F(v) = 1 - a. \qquad (66)$$

$$\binom{v}{a} \xrightarrow{negation} \binom{anti(v)}{a} \text{ or } \binom{v}{1-a}. \qquad (67)$$

## II.5. One-Attribute-Value Plithogenic Single-Valued Intuitionistic Fuzzy Set Operators

Let's consider the single-valued intuitionistic fuzzy degree of appurtenance of attribute value $v$ of $x$ to the set $P$ with respect to some given criteria:
$$d_A^{IF}(v) = (a_1, a_2) \in [0, 1]^2, \qquad (68)$$
and
$$d_B^{IF}(v) = (b_1, b_2) \in [0, 1]^2. \qquad (69)$$

### II.5.1. One-Attribute-Value Plithogenic Single-Valued Intuitionistic Fuzzy Set Intersection

$$(a_1, a_2) \wedge_p (b_1, b_2) = \left(a_1 \wedge_p a_2, b_1 \vee_p b_2 \right). \quad (70)$$

### II.5.2. One-Attribute-Value Plithogenic Single-Valued Intuitionistic Fuzzy Set Union

$$(a_1, a_2) \vee_p (b_1, b_2) = \left(a_1 \vee_p a_2, b_1 \wedge_p b_2 \right). \quad (71)$$

### II.5.3. One-Attribute-Value Plithogenic Single-Valued Intuitionistic Complements Set (Negations)

$$\neg_p(a_1, a_2) = (a_2, a_1) \quad (72)$$
$$\neg_p(a_1, a_2) = (1 - a_1, 1 - a_2) \quad (73)$$
Etc.

### II.5.4. One-Attribute-Value Plithogenic Single-Valued Intuitionistic Fuzzy Set Inclusions (Partial Orders)

$$(a_1, a_2) \leq_p (b_1, b_2) \text{ if } a_1 \leq (1-c_v) \cdot b_1,\ a_2 \geq (1-c_v) \cdot b_2, \quad (74)$$

where $c_v \in [0, 0.5)$ is the contradiction degree between the attribute dominant value and the attribute value $v$.

### II.5.5. One-Attribute-Value Plithogenic Single-Valued Intuitionistic Fuzzy Set Equality

$$(a_1, a_2) =_p (b_1, b_2) \text{ if } (a_1, a_2) \leq_p (b_1, b_2) \quad (75)$$
and
$$(b_1, b_2) \leq_p (a_1, a_2). \quad (76)$$

## II.6. One-Attribute Value Plithogenic Single-Valued Neutrosophic Set Operators

Let's consider the single-valued neutrosophic degree of appurtenance of attribute value $v$ of $x$ to the set $P$ with respect to some given criteria:

$$d_A^N(v) = (a_1, a_2, a_3) \in [0, 1]^3, \quad (77)$$

and

$$d_B^N(v) = (b_1, b_2, b_3) \in [0, 1]^3. \quad (78)$$

### II.6.1. One-Attribute-Value Plithogenic Single-Valued Neutrosophic Set Intersection

$$(a_1, a_2, a_3) \wedge_p (b_1, b_2, b_3) =$$
$$= \left(a_1 \wedge_p b_1, \frac{1}{2}(a_2 \wedge_F b_2 + a_2 \wedge_F b_2), a_3 \vee_p b_3\right). \quad (79)$$

### II.6.2. One-Attribute-Value Plithogenic Single-Valued Neutrosophic Set Union

$$(a_1, a_2, a_3) \vee_p (b_1, b_2, b_3) =$$
$$\left(a_1 \vee_p b_1, \frac{1}{2}(a_2 \wedge_F b_2 + a_2 \vee_F b_2), a_3 \wedge_p b_3\right). \quad (80)$$

### II.6.3. One-Attribute-Value Plithogenic Single-Valued Neutrosophic Set Complements (Negations)

$$\neg_p(a_1, a_2, a_3) = (a_3, a_2, a_1) \quad (81)$$
$$\neg_p(a_1, a_2, a_3) = (a_3, 1 - a_2, a_1) \quad (82)$$
$$\neg_p(a_1, a_2, a_3) = (1 - a_1, a_2, 1 - a_3) \quad (83)$$

Etc.

## II.6.4. One-Attribute-Value Plithogenic Single-Valued Neutrosophic Inclusions Set (Partial Orders)

### II.6.4.1. Simple Neutrosophic Inclusion

$$(a_1, a_2, a_3) \leq_N (b_1, b_2, b_3) \tag{84}$$

if $a_1 \leq b_1$, $a_2 \geq b_2$, $a_3 \geq b_3$.

### II.6.4.2. Complete Neutrosophic Inclusion

$$(a_1, a_2, a_3) \leq_p (b_1, b_2, b_3) \tag{85}$$

if $a_1 \leq b_1$, $a_2 \geq 0.5 \cdot b_2$, $a_3 \geq b_3$.

## II.6.5. One-Attribute-Value Plithogenic Single-Valued Neutrosophic Set Equality

$$(a_1, a_2, a_3) =_N (b_1, b_2, b_3) \tag{86}$$

if $(a_1, a_2, a_3) \leq_N (b_1, b_2, b_3)$
and $(b_1, b_2, b_3) \leq_N (a_1, a_2, a_3)$.
And similarly:

$$(a_1, a_2, a_3) =_p (b_1, b_2, b_3) \tag{87}$$

if $(a_1, a_2, a_3) \leq_p (b_1, b_2, b_3)$
and $(b_1, b_2, b_3) \leq_p (a_1, a_2, a_3)$.

## II.7. n-Attribute-Values Plithogenic Single-Valued Set Operators

The easiest way to apply the plithogenic operators on a multi-attribute plithogenic set is to split back the *m*-dimensional attribute, $m \geq 1$, into *m* uni-dimensional attributes, and separately apply the plithogenic operators on the set of all values (needed by the application to solve) of each given attribute.

Therefore, let α be a given attribute, characterizing each element $x \in P$, whose set of values are:

$$V = \{v_1, v_2, \ldots, v_n\} \equiv \{v_D, v_2, \ldots, v_n\}, n \geq 1, \quad (88)$$

where $v_D$ = dominant attribute value, and $c(v_D, v_i) = c_i \in [0, 1]$ the contradiction degrees. Without restricting the generality, we consider the values arranged in an increasing order with respect to their contradiction degrees, i.e.:

$$c(v_D, v_D) = 0 \leq c_1 \leq c_2 \leq \cdots \leq c_{i_0} < \frac{1}{2} \leq$$
$$\leq c_{i_0+1} \leq \cdots \leq c_n \leq 1. \quad (89)$$

## II.7.1. n-Attribute-Values Plithogenic Single-Valued Fuzzy Set Operators

Let's consider two experts, A and B, which evaluate an element $x$, with respect to the fuzzy degree of the values $v_1, \ldots, v_n$ of appurtenance to the set $P$, upon some given criteria:

$$d_A^F : P \times V \to [0, 1], d_A^F(x, v_i) = a_i \in [0, 1], \quad (90)$$
$$d_B^F : P \times V \to [0, 1], d_B^F(x, v_i) = b_i \in [0, 1], \quad (91)$$

for all $i \in \{1, 2, \ldots, n\}$.

## II.7.2. n-Attribute-Values Plithogenic Single-Valued Fuzzy Set Intersection

$$(a_1, a_2, \ldots, a_{i_0}, a_{i_0+1}, \ldots, a_n) \wedge_p (b_1, b_2, \ldots, b_{i_0}, b_{i_0+1}, \ldots, b_n) =$$
$$\begin{pmatrix} a_1 \wedge_p b_1, a_2 \wedge_p b_2, \ldots, a_{i_0} \wedge_p b_{i_0}, \\ a_{i_0+1} \wedge_p b_{i_0+1}, \ldots, a_n \wedge_p b_n \end{pmatrix} \quad (92)$$

The first $i_0$ intersections are proper plithogenic intersections (the weights onto the $t_{norm}$'s are bigger than onto $t_{conorm}$'s):

$$a_1 \wedge_p b_1, a_2 \wedge_p b_2, \ldots, a_{i_0} \wedge_p b_{i_0} \quad (93)$$

whereas the next $n - i_0$ intersections

$$a_{i_0+1} \wedge_p b_{i_0+1}, \ldots, a_n \wedge_p b_n \tag{94}$$

are improper plithogenic unions (since the weights onto the $t_{norm}$'s are less than onto $t_{conorm}$'s):

## II.7.3. n-Attribute-Values Plithogenic Single-Valued Fuzzy Set Union

$$(a_1, a_2, \ldots, a_{i_0}, a_{i_0+1}, \ldots, a_n) \vee_p (b_1, b_2, \ldots, b_{i_0}, b_{i_0+1}, \ldots, b_n)$$
$$= (a_1 \vee_p b_1, a_2 \vee_p b_2, \ldots, a_{i_0} \vee_p b_{i_0}, a_{i_0+1} \vee_p b_{i_0+1}, \ldots, a_n \vee_p b_n) \tag{95}$$

The first $i_0$ unions are proper plithogenic unions (the weights onto the $t_{conorm}$'s are bigger than onto $t_{norm}$'s):

$$a_1 \vee_p b_1, a_2 \vee_p b_2, \ldots, a_{i_0} \vee_p b_{i_0} \tag{96}$$

whereas the next $n - i_0$ unions

$$a_{i_0+1} \vee_p b_{i_0+1}, \ldots, a_n \vee_p b_n \tag{97}$$

are improper plithogenic intersections (since the weights onto the $t_{conorm}$'s are less than onto $t_{norm}$'s):

## II.7.4. n-Attribute-Values Plithogenic Single-Valued Fuzzy Set Complements (Negations)

In general, for a generic $x \in P$, characterized by the uni-dimensional attribute α, whose values are $V = (v_D, v_2, \ldots, v_n), n \geq 2$, whose attribute value contradiction degrees (with respect to the dominant attribute value $v_D$) are respectively: $0 \leq c_2 \leq \cdots \leq c_{n-1} \leq c_n \leq 1$, and their attribute value degrees of appurtenance to the set $P$ are respectively $a_D, a_2, \ldots, a_{n-1}, a_n \in [0,1]$, then the plithogenic fuzzy complement (negation) of $x$ is:

$$\neg_p [ x \begin{pmatrix} 0 & c_2 & & c_{n-1} & c_n \\ v_D, & v_2, & \dots, & v_{n-1}, & v_n \\ a_D & a_2 & & a_{n-1} & a_n \end{pmatrix} ] =$$

$$\neg_p x \begin{pmatrix} 1-c_n & 1-c_{n-1} & & 1-c_2 & 1-c_D \\ anti(v_n) & anti(v_{n-1}) & \dots & anti(v_2) & anti(v_D) \\ a_n & a_{n-1} & & a_2 & a_D \end{pmatrix}. \quad (98)$$

Some $anti(V_i)$ may coincide with some $V_j$, whereas other $anti(V_i)$ may fall in between two consecutive $[v_k, v_{k+1}]$ or we may say that they belong to the *Refined* set V;

or

$$= \begin{Bmatrix} v_n & v_{n-1} & \dots & \dots & v_1 \\ a_1, & a_2, & \dots, & \dots, & a_n \end{Bmatrix} \quad (99)$$

{This version gives an exact result when the contradiction degrees are equi-distant (for example: *0, 0.25, 0.50, 0.75, 1*) or symmetric with respect to the center *0.5* (for example: *0, 0.4, 0.6, 1*), and an approximate result when they are not equi-distant and not symmetric to the center (for example: *0, 0.3, 0.8, 0.9, 1*);}

or

$$\begin{Bmatrix} v_1 & v_2 & \dots & v_{i_0} & v_{i_0+1} & \dots & v_n \\ 1-a_1 & 1-a_2 & \dots & 1-a_{i_0} & 1-a_{i_0+1} & \dots & 1-a_n \end{Bmatrix} \quad (100)$$

where $anti(v_i) \in V$ or $anti(v_i) \in RefinedV$, for all $i \in \{1, 2, \dots, n\}$.

## II.7.5. n-Attribute-Values Plithogenic Single-Valued Fuzzy Set Inclusions (Partial Orders)

### II.7.5.1. Simple Inclusion (Partial Order)

$$(a_1, a_2, \dots, a_{i_0}, a_{i_0+1}, \dots, a_n) \leq_N$$
$$(b_1, b_2, \dots, b_{i_0}, b_{i_0+1}, \dots, b_n) \quad (101)$$

if

$a_1 \leq b_1, a_2 \leq b_2, \dots,$

$$a_{i_0} \leq b_{i_0}, a_{i_0+1} \geq b_{i_0+1}, \ldots, a_n \geq b_n. \tag{102}$$

**II.7.5.2. Plithogenic Inclusion (Partial Order)**

$$\begin{aligned}(a_1, a_2, \ldots, a_{i_0}, a_{i_0+1}, \ldots, a_n) \leq_p \\ (b_1, b_2, \ldots, b_{i_0}, b_{i_0+1}, \ldots, b_n)\end{aligned} \tag{103}$$

if

$a_1 \leq (1-c_1) \cdot b_1,$

$a_2 \leq (1-c_2) \cdot b_2, \ldots,$

$a_{i_0} \leq (1-c_{i_0}) \cdot b_{i_0},$

$a_{i_0+1} \geq (1-c_{i_0+1}) \cdot b_{i_0+1}, \ldots,$

$a_n \geq (1-c_n) \cdot b_n$

## II.7.6. n-Attribute-Values Plithogenic Single-Valued Fuzzy Set Equality

$$\begin{aligned}(a_1, a_2, \ldots, a_{i_0}, a_{i_0+1}, \ldots, a_n) =_p \\ (b_1, b_2, \ldots, b_{i_0}, b_{i_0+1}, \ldots, b_n)\end{aligned} \tag{104}$$

if

$$\begin{aligned}(a_1, a_2, \ldots, a_{i_0}, a_{i_0+1}, \ldots, a_n) \leq_p \\ (b_1, b_2, \ldots, b_{i_0}, b_{i_0+1}, \ldots, b_n)\end{aligned} \tag{105}$$

and

$$\begin{aligned}(b_1, b_2, \ldots, b_{i_0}, b_{i_0+1}, \ldots, b_n) \leq_p \\ (a_1, a_2, \ldots, a_{i_0}, a_{i_0+1}, \ldots, a_n).\end{aligned} \tag{106}$$

Similarly for "$=_N$".

## II.8. n-Attribute-Values Plithogenic Single-Valued Intuitionistic Fuzzy Set Operators

Let the intuitionistic fuzzy degree functions be:

$$d_A^{IF}: P \times V \to [0,1]^2, \; d_A^{IF}(x, v_i) = (a_{i_1}, a_{i_2}) \in [0,1]^2, \quad (107)$$
$$d_B^{IF}: P \times V \to [0,1]^2, \; d_B^{IF}(x, v_i) = (b_{i_1}, b_{i_2}) \in [0,1]^2, \quad (108)$$
for all $i \in \{1, 2, \ldots, n\}$.

## II.8.1. n-Attribute-Values Plithogenic Single-Valued Intuitionistic Fuzzy Set Intersection

$$\left((a_{11}, a_{12}), (a_{21}, a_{22}), \ldots, (a_{i_0 1}, a_{i_0 2}), (a_{i_0+1,1}, a_{i_0+1,2}), \ldots, (a_{n1}, a_{n2})\right) \wedge_p$$
$$\left((b_{11}, b_{12}), (b_{21}, b_{22}), \ldots, (b_{i_0 1}, b_{i_0 2}), (b_{i_0+1,1}, b_{i_0+1,2}), \ldots, (b_{n1}, b_{n2})\right) =$$
$$\begin{pmatrix} (a_{11} \wedge_p b_{11}, a_{12} \vee_p b_{12}), (a_{21} \wedge_p b_{21}, a_{22} \vee_p b_{22}), \ldots, \\ (a_{i_0 1} \wedge_p b_{i_0 1}, a_{i_0 2} \vee_p b_{i_0 2}), (a_{i_0+1,1} \wedge_p b_{i_0+1,1}, a_{i_0 1,2} \vee_p b_{i_0 1,2}), \ldots, \\ (a_{n1} \wedge_p b_{n1}, a_{n2} \vee_p b_{n2}) \end{pmatrix}.$$
(109)

Similarly, the first $i_0$ intersections (of first duplet components) are proper plithogenic intuitionistic fuzzy intersections

$$(a_{11} \wedge_p b_{11}, a_{12} \vee_p b_{12}), (a_{21} \wedge_p b_{21}, a_{22} \vee_p b_{22}), \ldots,$$
$$(a_{i_0 1} \wedge_p b_{i_0 1}, a_{i_0 2} \vee_p b_{i_0 2}), \quad (110)$$

since for each duplet, for the first component, the weights onto the $t_{norm}$'s are bigger than onto $t_{conorm}$'s, while for the second component the weights onto the $t_{norm}$'s are smaller than onto $t_{conorm}$'s,

And the next $n - i_0$ intersections (of first duplet components) are rather plithogenic intuitionistic fuzzy unions:
$$(a_{i_0+1,1} \wedge_p b_{i_0+1,1}, a_{i_0 1,2} \vee_p b_{i_0 1,2}), \ldots, (a_{n1} \wedge_p b_{n1}, a_{n2} \vee_p b_{n2})$$
(111)

since for each duplet, for the first component, the weights onto the $t_{norm}$'s are smaller than onto $t_{conorm}$'s, while for the second component the weights onto the $t_{norm}$'s are bigger than onto $t_{conorm}$'s,

## II.8.2. Attribute-Values Plithogenic Single-Valued Intuitionistic Fuzzy Set Union

Let's use simpler notations of the elements:
$$\big((a_{i1}, a_{i2}), 1 \leq i \leq n\big) \vee_p \big((b_{i1}, b_{i2}), 1 \leq i \leq n\big) = \big((a_{i1} \vee_p b_{i1}, a_{i2} \wedge_p b_{i2}), 1 \leq i \leq n\big). \quad (112)$$

Analogously, the first $i_0$ unions are proper plithogenic intuitionistic fuzzy unions, since for each duplet, for the first duplet component, the weights onto the $t_{conorm}$'s are bigger than onto $t_{norm}$'s, while for the second duplet component the weights onto the $t_{conorm}$'s are smaller than onto $t_{norm}$'s,

And the next $n - i_0$ unions are improper plithogenic intuitionistic fuzzy intersections, since for each duplet, for the first duplet component, the weights onto the $t_{conorm}$'s are smaller than onto $t_{norm}$'s, while for the second duplet component the weights onto the $t_{conorm}$'s are bigger than onto $t_{norm}$'s,

## II.8.3. n-Attribute-Values Plithogenic Single-Valued Intuitionistic Fuzzy Set Complements (Negations)

$$\neg_p \left\{ \genfrac{}{}{0pt}{}{v_i}{(a_{i1}, a_{i2})} \middle| i \in \{1, 2, \ldots, n\} \right\} = \left\{ \genfrac{}{}{0pt}{}{anti(v_i)}{(a_{i1}, a_{i2})} \middle| i \in \{1, 2, \ldots, n\} \right\}$$
$$(113)$$

$$\text{or} \left\{ \genfrac{}{}{0pt}{}{v_i}{(a_{i2}, a_{i1})} \middle| i \in \{1, 2, \ldots, n\} \right\} \quad (114)$$

$$\text{or} \left\{ \genfrac{}{}{0pt}{}{v_i}{(1 - a_{i1}, 1 - a_{i2})} \middle| i \in \{1, 2, \ldots, n\} \right\}. \quad (115)$$

## II.8.4. n-Attribute-Values Plithogenic Single-Valued Intuitionistic Fuzzy Set Inclusions (Partial Orders)

Let
$$A = \begin{pmatrix} (a_{11}, a_{12}), (a_{21}, a_{22}), \ldots, (a_{i_0 1}, a_{i_0 2}), \\ (a_{i_0+1,1}, a_{i_0+1,2}), \ldots, (a_{n1}, a_{n2}) \end{pmatrix} \quad (116)$$

and
$$B = \begin{pmatrix} (b_{11}, b_{12}), (b_{21}, b_{22}), \ldots, (b_{i_0 1}, b_{i_0 2}), \\ (b_{i_0+1,1}, b_{i_0+1,2}), \ldots, (b_{n1}, b_{n2}) \end{pmatrix}. \quad (117)$$

### II.8.4.1. Simple Intuitionistic Fuzzy Inclusion

$A \leq_{IF} B$      (118)

iff $(a_{i1}, a_{i2}) \leq_{IF} (b_{i1}, b_{i2})$ for $1 \leq i \leq i_0$,

and $(b_{j1}, b_{j2}) \leq_{IF} (a_{j1}, a_{j2})$ for $i_0 + 1 \leq j \leq n$.

### II.8.4.2. Plithogenic Intuitionistic Fuzzy Inclusion

$A \leq_P B$      (119)

iff $(a_{i1}, a_{i2}) \leq_P (b_{i1}, b_{i2})$ for $1 \leq i \leq i_0$,

and $(b_{j1}, b_{j2}) \leq_p (a_{j1}, a_{j2})$ for $i_0 + 1 \leq j \leq n$.

## II.8.5. n-Attribute-Values Plithogenic Single-Valued Intuitionistic Fuzzy Set Equality

### II.8.5.1. Simple Intuitionistic Fuzzy Equality

$A =_{IF} B$ iff $A \leq_{IF} B$ and $B \leq_{IF} A$.      (120)

### II.8.5.2. Plithogenic-mple Intuitionistic Fuzzy Equality

$A =_P B$ iff $A \leq_P B$ and $B \leq_P A$.      (121)

## II.9. n-Attribute-Values Plithogenic Single-Valued Neutrosophic Set Operators

Let the neutrosophic degree functions be:
$$d_A^N: P \times V \to [0,1]^3, d_A^N(x, v_i) = (a_{i_1}, a_{i_2}, a_{i_3}) \in [0,1]^3, (122)$$
$$d_B^N: P \times V \to [0,1]^3, d_B^N(x, v_i) = (b_{i_1}, b_{i_2}, b_{i_3}) \in [0,1]^3, (123)$$
for all $i \in \{1, 2, \ldots, n\}$.

## II.9.1. n-Attribute-Values Plithogenic Single-Valued Neutrosophic Set Intersection

$$\begin{aligned}&\left((a_{11},a_{12},a_{13}),(a_{21},a_{22},a_{23}),\ldots,(a_{i_0 1},a_{i_0 2},a_{i_0 3}),(a_{i_0+1,1},a_{i_0+1,2},a_{i_0+1,3}),\ldots,(a_{n1},a_{n2},a_{n3})\right) \wedge_P \\ &\left((b_{11},b_{12},b_{13}),(b_{21},b_{22},b_{23}),\ldots,(b_{i_0 1},b_{i_0 2},b_{i_0 3}),(b_{i_0+1,1},b_{i_0+1,2},b_{i_0+1,3}),\ldots,(b_{n1},b_{n2},b_{n3})\right) = \\ &\begin{pmatrix}(a_{11},a_{12},a_{13}) \wedge_P (b_{11},b_{12},b_{13}),(a_{21},a_{22},a_{23}) \wedge_P (b_{21},b_{22},b_{23}),\ldots,(a_{i_0 1},a_{i_0 2},a_{i_0 3}) \wedge_P (b_{i_0 1},b_{i_0 2},b_{i_0 3}), \\ (a_{i_0+1,1},a_{i_0+1,2},a_{i_0+1,3}) \wedge_P (b_{i_0+1,1},b_{i_0+1,2},b_{i_0+1,3}),\ldots,(a_{n1},a_{n2},a_{n3}) \wedge_P (b_{n1},b_{n2},b_{n3})\end{pmatrix}\end{aligned}$$
(124)

With simpler notations:
$$\begin{aligned}&\left((a_{i1}, a_{i2}, a_{i3}), 1 \leq i \leq n\right) \wedge_p \left((b_{i1}, b_{i2}, b_{i3}), 1 \leq i \leq n\right) \\ &= \left(\left(a_{i1} \wedge_p b_{i1}, \frac{1}{2}(a_{i2} \wedge_F b_{i2}) + \frac{1}{2}(a_{i2} \vee_F b_{i2}), a_{i3} \vee_p b_{i3}\right),\right. \\ &\quad\quad\quad\quad\quad\quad\quad\quad\quad 1 \leq i \leq n \end{aligned}$$
(125)

Analogously, the first $i_0$ intersections are proper plithogenic intersections, since for each triplet, for the first triplet component, the weights onto the *t*<sub>norm</sub>'s are bigger than onto *t*<sub>conorm</sub>'s, while for the third triplet component the weights onto the *t*<sub>norm</sub>'s are smaller than onto *t*<sub>conorm</sub>'s.

And the next $n - i_0$ intersections are improper plithogenic unions, since for each triplet, for the first triplet component, the weights onto the *t*<sub>norm</sub>'s are smaller than onto *t*<sub>conorm</sub>'s, while for the third triplet component the weights onto the $t_{norm}$'s are bigger than onto $t_{conorm}$'s,

## II.9.2. n-Attribute-Values Plithogenic Single-Valued Neutrosophic Set Union

$$\begin{pmatrix}(a_{11},a_{12},a_{13}),(a_{21},a_{22},a_{23}),...,(a_{i_01},a_{i_02},a_{i_03}),(a_{i_0+1,1},a_{i_0+1,2},a_{i_0+1,3}),...,(a_{n1},a_{n2},a_{n3})\end{pmatrix} \vee_P$$
$$\begin{pmatrix}(b_{11},b_{12},b_{13}),(b_{21},b_{22},b_{23}),...,(b_{i_01},b_{i_02},b_{i_03}),(b_{i_0+1,1},b_{i_0+1,2},b_{i_0+1,3}),...,(b_{n1},b_{n2},b_{n3})\end{pmatrix} =$$
$$\begin{pmatrix}(a_{11},a_{12},a_{13}) \vee_P (b_{11},b_{12},b_{13}),(a_{21},a_{22},a_{23}) \vee_P (b_{21},b_{22},b_{23}),...,(a_{i_01},a_{i_02},a_{i_03}) \vee_P (b_{i_01},b_{i_02},b_{i_03}), \\ (a_{i_0+1,1},a_{i_0+1,2},a_{i_0+1,3}) \vee_P (b_{i_0+1,1},b_{i_0+1,2},b_{i_0+1,3}),...,(a_{n1},a_{n2},a_{n3}) \vee_P (b_{n1},b_{n2},b_{n3})\end{pmatrix}$$

(126)

With simpler notations:
$$\left((a_{i1}, a_{i2}, a_{i3}), 1 \leq i \leq n\right) \vee_p \left((b_{i1}, b_{i2}, b_{i3}), 1 \leq i \leq n\right)$$
$$= \left(\left(a_{i1} \vee_p b_{i1}, \frac{1}{2}(a_{i2} \wedge_F b_{i2}) + \frac{1}{2}(a_{i2} \vee_F b_{i2}), a_{i3} \wedge_p b_{i3}\right),\right.$$
$$\left. 1 \leq i \leq n \right)$$

(127)

Analogously, the first $i_0$ unions are proper plithogenic unions, since for each triplet, for the first triplet component, the weights onto the $t_{conorm}$'s are bigger than onto $t_{norm}$'s, while for the third triplet component the weights onto the $t_{conorm}$'s are smaller than onto $t_{norm}$'s.

And the next $n - i_0$ unions are rather plithogenic intersections, since for each triplet, for the first triplet component, the weights onto the $t_{conorm}$'s are smaller than onto $t_{norm}$'s, while for the third triplet component the weights onto the $t_{conorm}$'s are bigger than onto $t_{norm}$'s,

## II.9.3. n-Attribute-Values Plithogenic Single-Valued Neutrosophic Set Complements (Negations)

$$\neg_p \{(a_{i1}, a_{i2}, a_{i3}), 1 \leq i \leq n\} = \left\{ \begin{array}{c} anti(v_i) \\ (a_{i1}, a_{i2}, a_{i3}) \end{array} \middle| i \in \{1,, 2, \ldots, n\} \right\}$$
(128)

or

$$\left\{ \begin{array}{c} v_i \\ (a_{i3}, a_{i2}, a_{i1}) \end{array} \middle| i \in \{1,, 2, \ldots, n\} \right\}$$
(129)

or

$$\left\{ \begin{array}{c} v_i \\ (a_{i3}, 1 - a_{i2}, a_{i1}) \end{array} \middle| i \in \{1,, 2, \ldots, n\} \right\}$$
(130)

Etc.

## II.9.4. n-Attribute-Values Plithogenic Single-Valued Neutrosophic Set Inclusions (Partial Orders)

### II.9.4.1. Simple Neutrosophic Inclusion

$$\big((a_{11}, a_{12}, a_{13}), (a_{21}, a_{22}, a_{23}), \ldots, (a_{i_01}, a_{i_02}, a_{i_03}), (a_{i_0+1,1}, a_{i_0+1,2}, a_{i_0+1,3}), \ldots, (a_{n1}, a_{n2}, a_{n3})\big) \leq_N$$
$$\big((b_{11}, b_{12}, b_{13}), (b_{21}, b_{22}, b_{23}), \ldots, (b_{i_01}, b_{i_02}, b_{i_03}), (b_{i_0+1,1}, b_{i_0+1,2}, b_{i_0+1,3}), \ldots, (b_{n1}, b_{n2}, b_{n3})\big)$$
(131)

iff

$(a_{11}, a_{12}, a_{13}) \leq_N (b_{11}, b_{12}, b_{13}), (a_{21}, a_{22}, a_{23}) \leq_N (b_{21}, b_{22}, b_{23}), \ldots, (a_{i_01}, a_{i_02}, a_{i_03}) \leq_N (b_{i_01}, b_{i_02}, b_{i_03}),$
$(a_{i_0+1,1}, a_{i_0+1,2}, a_{i_0+1,3}) \geq_N (b_{i_0+1,1}, b_{i_0+1,2}, b_{i_0+1,3}), \ldots, (a_{n1}, a_{n2}, a_{n3}) \geq_N (b_{n1}, b_{n2}, b_{n3}).$
(132)

### II.9.4.2. Plithogenic Neutrosophic Inclusion

$$\big((a_{11}, a_{12}, a_{13}), (a_{21}, a_{22}, a_{23}), \ldots, (a_{i_01}, a_{i_02}, a_{i_03}), (a_{i_0+1,1}, a_{i_0+1,2}, a_{i_0+1,3}), \ldots, (a_{n1}, a_{n2}, a_{n3})\big) \leq_P$$
$$\big((b_{11}, b_{12}, b_{13}), (b_{21}, b_{22}, b_{23}), \ldots, (b_{i_01}, b_{i_02}, b_{i_03}), (b_{i_0+1,1}, b_{i_0+1,2}, b_{i_0+1,3}), \ldots, (b_{n1}, b_{n2}, b_{n3})\big)$$
(133)

iff

$$(a_{11}, a_{12}, a_{13}) \leq_P (b_{11}, b_{12}, b_{13}), (a_{21}, a_{22}, a_{23}) \leq_P (b_{21}, b_{22}, b_{23}), ..., (a_{i_0 1}, a_{i_0 2}, a_{i_0 3}) \leq_P (b_{i_0 1}, b_{i_0 2}, b_{i_0 3}),$$
$$(a_{i_0+1,1}, a_{i_0+1,2}, a_{i_0+1,3}) \geq_P (b_{i_0+1,1}, b_{i_0+1,2}, b_{i_0+1,3}), ..., (a_{n1}, a_{n2}, a_{n3}) \geq_P (b_{n1}, b_{n2}, b_{n3})$$
(134)

## II.9.5. n-Attribute-Values Plithogenic Single-Valued Neutrosophic Set Equality

$$\left((a_{11}, a_{12}, a_{13}), (a_{21}, a_{22}, a_{23}), ..., (a_{i_0 1}, a_{i_0 2}, a_{i_0 3}), (a_{i_0+1,1}, a_{i_0+1,2}, a_{i_0+1,3}), ..., (a_{n1}, a_{n2}, a_{n3})\right) =_P$$
$$\left((b_{11}, b_{12}, b_{13}), (b_{21}, b_{22}, b_{23}), ..., (b_{i_0 1}, b_{i_0 2}, b_{i_0 3}), (b_{i_0+1,1}, b_{i_0+1,2}, b_{i_0+1,3}), ..., (b_{n1}, b_{n2}, b_{n3})\right)$$
(135)

if

$$\left((a_{11}, a_{12}, a_{13}), (a_{21}, a_{22}, a_{23}), ..., (a_{i_0 1}, a_{i_0 2}, a_{i_0 3}), (a_{i_0+1,1}, a_{i_0+1,2}, a_{i_0+1,3}), ..., (a_{n1}, a_{n2}, a_{n3})\right) \leq_P$$
$$\left((b_{11}, b_{12}, b_{13}), (b_{21}, b_{22}, b_{23}), ..., (b_{i_0 1}, b_{i_0 2}, b_{i_0 3}), (b_{i_0+1,1}, b_{i_0+1,2}, b_{i_0+1,3}), ..., (b_{n1}, b_{n2}, b_{n3})\right)$$
(136)

and

$$\left((b_{11}, b_{12}, b_{13}), (b_{21}, b_{22}, b_{23}), ..., (b_{i_0 1}, b_{i_0 2}, b_{i_0 3}), (b_{i_0+1,1}, b_{i_0+1,2}, b_{i_0+1,3}), ..., (b_{n1}, b_{n2}, b_{n3})\right) \leq_P$$
$$\left((a_{11}, a_{12}, a_{13}), (a_{21}, a_{22}, a_{23}), ..., (a_{i_0 1}, a_{i_0 2}, a_{i_0 3}), (a_{i_0+1,1}, a_{i_0+1,2}, a_{i_0+1,3}), ..., (a_{n1}, a_{n2}, a_{n3})\right)$$
(137)

And similarly for "$=_N$".

## II.10. Theorems Related to One-Attribute-Value Plithogenic Single-Valued Fuzzy Set Intersections and Unions

Let $v_D$ be an attribute dominant value, and $v$ any attribute value.

## II.10.1. Theorem 1.

If $c(v_D, v) = 0$, then:

1.1. if on $v_D$ one applies the $t_{norm}$, on $v$ one also applies the $t_{norm}$;

1.2. if on $v_D$ one applies the $t_{conorm}$, on $v$ one also applies the $t_{conorm}$.

## II.10.2. Theorem 2.

If $c(v_D, v) = 1$, then:

2.1. if on $v_D$ one applies the $t_{norm}$, on $v$ one applies the $t_{conorm}$;

2.2. if on $v_D$ one applies the $t_{conorm}$, on $v$ one applies the $t_{norm}$.

## II.10.3. Theorem 3.

If $0 < c(v_D, v) < 1$, then on $v$ one applies a linear combination of $t_{norm}$ and $t_{conorm}$.

## II.10.4. Theorem 4.

Let $a, b$ be the fuzzy degrees of appurtenance of the attribute value $v$ with respect to Experts A and B. Then:

$$a \wedge_p b + a \vee_p b = a \wedge_F b + a \vee_F b. \qquad (138)$$

*Proof.*

Let $c_0 = c(v_D, v) \in [0, 1]$. Then

$$\begin{aligned}
a \wedge_p b &+ a \vee_p b \\
&= (1 - c_0) \cdot [a \wedge_F b] + c_0 \cdot [a \vee_F b] \\
&\quad + (1 - c_0) \cdot [a \vee_F b] + c_0 \cdot [a \wedge_F b] \\
&= a \wedge_F b + a \vee_F b. \qquad (139)
\end{aligned}$$

## II.10.5. Theorem 5.

Let $a, b$ be the fuzzy degrees of appurtenance of the attribute value $v$ with respect to Experts A and B.

If the degree of contradiction of $a$ and $b$, with respect to their corresponding dominant attribute values is equal to 0.5, then

$$a \wedge_p b = a \vee_p b. \tag{140}$$

*Proof:*

Since $c_0 = 0.5$, then $1 - c_0 = c_0 = 0.5$ and therefore the definitions of $a \wedge_p b$ and $a \vee_p b$ become the same.

## II.10.6. Theorem 6.

Let $a, b$ be the fuzzy degrees of appurtenance of the attribute value $v$ with respect to Experts A and B respectively, and the contradiction degree of attribute value $v$ with the attribute dominant value $v_D$ by $c_0$.

Let $a', b'$ be the fuzzy degrees of appurtenance of the attribute value $v'$ with respect to Experts A and B respectively, where $a' = a$ and $b' = b$, and the contradiction degree of attribute value $v'$ with the attribute dominant value $v_D$ be $1 - c_0$.

Then:

$$a \wedge_p b = a' \vee_p b'. \tag{141}$$

*Proof:*

$$a \wedge_p b = (1-c_0) \cdot [a \wedge_F b] + c_0 \cdot [a \vee_F b] =$$
$$c_0 \cdot [a \vee_F b] + (1-c_0) \cdot [a \wedge_F b] =$$
$$[1-(1-c_0)] \cdot [a \vee_F b] + (1-c_0) \cdot [a \wedge_F b] =$$
$$[1-(1-c_0)] \cdot [a' \vee_F b'] + (1-c_0) \cdot [a' \wedge_F b'] =$$
{ since a = a' and b = b'}

$$= a' \vee_p b' \quad (142)$$

{ since the contradiction degree between v' and $v_D$ is $1 - c_0$ }.

## II.10.7. Theorem 7.

Let's consider a plithogenic neutrosophic set $P$, and $x$ an element from $P$. The neutrosophic degrees of appurtenance of the element x's attribute values $v$ with respect to Experts A and B are respectively $(T_1, I_1, F_1)$ and $(T_2, I_2, F_2)$. The *interior degrees of contradiction* between the neutrosophic components $T, I, F$, or truth, indeterminacy, and falsehood respectively, are:

$$c(T, I) = \frac{1}{2}, c(I, F) = \frac{1}{2} \text{ and } c(T, F) = 1,$$

where $T$ is considered the dominant neutrosophic component.

If one applies the $t_{norm}$ on $T_1$ and $T_2$, then one has to apply the opposite, i.e. the $t_{conorm}$ on $F_1$ and $F_2$ - since $T$ and $F$ are 100% opposite. Similarly, if one applies the $t_{conorm}$ on $T_1$ and $T_2$, then one has to apply the opposite, i.e. the $t_{norm}$ on $F_1$ and $F_2$.

But $I$ is only half (50%) opposite to both $T$ and $F$, therefore no matter if $t_{norm}$ or $t_{conorm}$ were applied on $T_1$ and $T_2$, on $I_1$ and $I_2$ one applies:

$$\frac{1}{2}(I_1 \wedge_p I_2) + \frac{1}{2}(I_1 \vee_p I_2) = \frac{1}{2}(I_1 \vee_p I_2 + I_1 \wedge_p I_2) =$$
$$\frac{1}{2}(I_1 \wedge_F I_2 + I_1 \vee_F I_2). \quad (143)$$

If the *exterior degree of contradiction* between $v$ and its corresponding dominant attribute value $v_D$ is $c(v_D, v) = c_0$, then again one applies it on the above $\wedge_p$ and $\vee_p$ plithogenic operators. But, according to *Theorem 4* (where it was proved that: for any $c(v_D, v)$ one has $a \wedge_p b + a \vee_p b = a \wedge_F b + a \vee_F b$ ), no matter the exterior contradiction degree, we always get:

$$(T_1, I_1, F_1) \wedge_p (T_2, I_2, F_2) =$$
$$\left(T_1 \wedge_p T_2, \frac{1}{2}(I_1 \wedge_F I_2 + I_1 \vee_F I_2), F_1 \vee_p F_2\right), \quad (144)$$

and respectively

$$(T_1, I_1, F_1) \vee_p (T_2, I_2, F_2) =$$
$$\left(T_1 \vee_p T_2, \frac{1}{2}(I_1 \wedge_F I_2 + I_1 \vee_F I_2), F_1 \wedge_p F_2\right). \quad (145)$$

## II.11. First Classification of the Plithogenic Set

### II.11.1. Refined Plithogenic Set

If at least one of the attribute's values $v_k \in V$ is split (refined) into two or more attribute sub-values: $v_{k1}, v_{k2}, \ldots \in V$, with the attribute sub-value appurtenance degree function: $d(x, v_{ki}) \in \mathcal{P}([0, 1])$, for $i = 1, 2, \ldots$, then $(P_r, \alpha, V, d, c)$ is called a *Refined Plithogenic Set*, where "*r*" stands for "refined".

### II.11.2. Plithogenic Overset / Underset / Offset

If for at least one of the attribute's values $v_k \in V$, of at least one element

$x \in P$, has the attribute value appurtenance degree function $d(x, v_k)$ exceeding *1*, then $(P_o, \alpha, V, d, c)$ is called a *Plithogenic Overset*, where "*o*" stands for "overset"; but if $d(x, v_k)$ is below *0*, then $(P_u, \alpha, V, d, c)$ is called a *Plithogenic Underset*, where "*u*" stands for "underset"; while if $d(x, v_k)$ exceeds *1*, and $d(y, s_j)$ is below *0* for the attribute values $v_k, v_j \in V$ that may be the same or different attribute values corresponding to the same element or to two different elements $x, y \in P$, then

*(P_off, α, V, d, c)* is called a *Plithogenic Offset*, where "off" stands for "offset" (or plithogenic set that is both overset and underset).

## II.11.3. Plithogenic Multiset

A plithogenic set *P* that has at least an element $x \in P$, which repeats into the set P with the same plithogenic components

$x(a_1, a_2, ..., a_m), x(a_1, a_2, ..., a_m)$  (146)

or with different plithogenic components

$x(a_1, a_2, ..., a_m), x(b_1, b_2, ..., b_m),$  (147)

then *(P_m, α, V, d, c)* is called a *Plithogenic Multiset*, where "*m*" stands for "multiset".

## II.11.4. Plithogenic Bipolar Set

If $\forall x \in P$, d: $P \times V \to \{\mathcal{P}[-1, 0]) \times \mathcal{P}[0, 1])\}^z$, then (P_b, α, V, d, c) is called a *Plithogenic Bipolar Set*, since *d(x, v)*, for $v \in V$, associates an appurtenance negative degree (as a subset of *[-1, 0]*) and a positive degree (as a subset of *[0, 1]*) to the value *v*; where *z = 1* for fuzzy degree, *z = 2* for intuitionistic fuzzy degree, and *z = 3* for neutrosophic fuzzy degree.

## II.11.5-6. Plithogenic Tripolar Set & Plitogenic Multipolar Set

Similar definitions for Plithogenic Tripolar Set and Plitogenic Multipolar Set (extension from Neutrosophic Tripolar Set and respectively Neutrosophic Multipolar Set {[4], 123-125}.

## II.11.7. Plithogenic Complex Set

If, for any $x \in P$, d: $P \times V \to \{\mathcal{P}[0, 1]) \times \mathcal{P}[0, 1])\}^z$, and for any $v \in V$, $d(x, v)$ is a complex value, i.e. $d(x, v) = M_1 \cdot e^{jM_2}$, where $M_1 \subseteq [0, 1]$ is called amplitude, and $M_2 \subseteq [0, 1]$ is called phase, and the appurtenance degree may be fuzzy ($z = 1$), intuitionistic fuzzy ($z = 2$), or neutrosophic ($z = 3$), then ($P_{com}$, α, V, d, c) is called a *Plithogenic Complex Set*.

## II.12. Second Classification of Multi-Attribute Plithogenic Fuzzy Sets

Upon the values of the appurtenance degree function, one has:

### II.12.1. Single-Valued Plithogenic Fuzzy Set, if

$\forall x \in P$, d: $P \times V \to [0, 1]$, (148)

and $\forall v \in V$, *d(x, v)* is a single number in *[0, 1]*.

### II.12.2. Hesitant Plithogenic Fuzzy Set, if

$\forall x \in P$, *d: P×V→ $\mathcal{P}$([0, 1])*, (149)

and $\forall v \in V$, *d(x, v)* is a discrete finite set of the form *{n₁, n₂, ..., n_p}*, where

$1 \leq p < \infty$, included in [0, 1].

### II.12.3. Interval-Valued Plithogenic Fuzzy Set, if

$\forall x \in P$, *d: P×V→ $\mathcal{P}$([0, 1])*, (150)

and $\forall v \in V$, *d(x, v)* is an (open, semi-open, closed) interval included in *[0, 1]*.

## II.13. Applications of Uni-Dimensional Attribute Plithogenic Single-Valued Fuzzy Set

Let *U* be a universe of discourse, and a non-empty plithogenic set $P \subseteq U$. Let $x \in P$ be a generic element.

For simplicity, we consider the uni-dimensional attribute and the single-valued fuzzy degree function.

### II.13.1. Small Discrete-Set of Attribute-Values

If the *attribute* is "*color*", and we consider only a discrete *set of attribute values V*, formed by the following six pure colors:

V = *{violet, blue, green, yellow, orange, red}*,

the *attribute value appurtenance degree function*:

$d: P \times V \to [0, 1]$,  (151)

$d(x, violet) = v \in [0, 1]$, $d(x, blue) = b \in [0, 1]$, $d(x, green) = g \in [0, 1]$,

$d(x, yellow) = y \in [0, 1]$, $d(x, orange) = o \in [0, 1]$, $d(x, red) = r \in [0, 1]$,

then one has:

*x(v, b, g, y, o, r)*,

where *v, b, g, y, o, r* are fuzzy degrees of violet, blue, green, yellow, orange, and red, respectively, of the object *x* with respect to the set of objects *P*, where

*v, b, g, y, o, r* ∈ [0, 1].

The cardinal of the set of attribute values *V* is *6*.

The other colors are blends of these pure colors.

## II.13.2. Large Discrete-Set of Attribute-Values

If the *attribute* is still "*color*" and we choose a more refined representation of the color values as:

$x\{d_{390}, d_{391}, \ldots, d_{699}, d_{700}\}$,

measured in nanometers, then we have a discrete finite set of attribute values, whose cardinal is: *700 – 390 + 1 = 311*.

Where for each $j \in V = \{390, 391, \ldots, 699, 700\}$, $d_j$ represents the degree to which the object *x*'s color, with respect to the set of objects *P*, is of "*j*" nanometers per wavelength, with $d_i \in [0, 1]$. A nanometer (nm) is a billionth part of a meter.

## II.13.3. Infinitely-Uncountable-Set of Attribute-Values

But if the *attribute* is again "*color*", then one may choose a continuous representation:

$x(d([390, 700]))$,

having $V = [390, 700]$ a closed real interval, hence an infinitely uncountable (continuum) set of attribute values. The cardinal of the *V* is ∞.

For each $j \in [390, 700]$, $d_j$ represents the degree to which the object *x*'s color, with respect to the set of objects *P*, is of "*j*" nanometers per wavelength, with $d_i \in [0, 1]$. And $d([390, 700]) = \{d_j, j \in [390, 700]\}$.

The light, ranging from *390* (violet color) to *700* (red color) nanometers per wavelengths is visible to the eye of the human. The cardinal of the set of attribute values *V* is continuum infinity.

## II.14. Multi-Attribute Plithogenic General Set

## II.14.1. Definition of Multi-Attribute Plithogenic General Set

The Multi-Attribute Plithogenic Set is obviously a generalization of the One-Attribute Plithogenic Set.

While the one-attribute plithogenic set has each element $x \in P$ characterized by a single attribute $\alpha \in \mathscr{A}$, in a more general form we extend it to the *multi-attribute plithogenic set*, defined as follows.

Let $U$ be a universe of discourse, and a non-empty plithogenic set $P \subseteq U$.

Let $\mathscr{A}$ be a set of $m \geq 2$ attributes: $\alpha_1, \alpha_2, \ldots, \alpha_m$, whose corresponding spectra of values are the non-empty sets $S_1, S_2, \ldots, S_m$ respectively.

Let $V_1 \subseteq S_1, V_2 \subseteq S_2, \ldots, V_m \subseteq S_m$ be subsets of attribute values of the attributes $\alpha_1, \alpha_2, \ldots, \alpha_m$ respectively needed by experts in their given application.

For each $j \in \{1, 2, \ldots, m\}$, the *set of attribute values* $V_j$ means the *range of attribute $\alpha_j$'s values*, needed by the experts in a specific application or in a specific problem to solve.

Each element $x \in P$ is characterized by all $m$ attributes' values.

Let the *m-dimensional attribute value degree function* be:
$$d_{[m]}: (P, V_1 \times V_2 \times \ldots \times V_m) \to \mathscr{P}([0,1])^m. \quad (152)$$

For any $x \in P$, and any $v_j \in V_j$ with $j \in \{1, 2, \ldots, m\}$, one has:
$$d_{[m]}(x(v_1, v_2, \ldots, v_m)) \subseteq \mathscr{P}([0,1])^m. \quad (153)$$

## II.14.2. m-Dimensional Attribute

Instead of working with *m uni-dimensional distinct attributes*, we may use only *a single m-dimensional attribute*, employing the following notations:

$\alpha_{[m]} = (\alpha_1, \alpha_2, \ldots, \alpha_m)$, $V_{[m]} = V_1 \times V_2 \times \ldots \times V_m$,
and $v_{[m]} = (v_1, v_2, \ldots, v_m)$, (154)
whence:
$d_{[m]}: (P, V_{[m]}) \to \mathcal{P}([0,1])^m$ (155)
with $d_{[m]}(x(v_{[m]})) \subseteq \mathcal{P}([0,1])^m$ (156)

and the set of values of the attribute $\alpha_j$, for $1 \leq j \leq m$, being $V_j = \{v_{jk}\}_{k \in W_j}$, where $W_j$ is the (finite, or countably, or unaccountably infinite) set of indexes corresponding to $V_j$.

In other words, each $V_j$ may be finite, or countably or unaccountably infinite set of values of the attribute $\alpha_j$ needed by the experts in a specific application or in a specific problem to solve.

On each attribute value set $V_j$, $1 \leq j \leq m$, there may exists a *dominant attribute value* $v_{jk_o} \in V_j$, determined by the experts upon the application or problem to solve.

{However, there are cases when such dominant attribute value may not be taking into consideration or may not exist, or there may be many dominant attribute values. In such cases, either one discards the contradiction degree function, or a different procedure (or multi-contradictory degree function) may be designed.}

The attribute value degrees of contradiction are afterwards calculated between the dominant attribute value and the other attribute values in special, but also among any attribute values.

For example, in the neutrosophic set, the attribute "appurtenance" has three values: membership, indeterminacy, and nonmembership. The dominant attribute value is *membership*, because it is the most important in all neutrosophic applications.

The degrees of contradictions are:

*c(nonmembership, membership)* = 1, and

*c(indeterminacy, membership)* =

*c(indeterminacy, nonmembership)* = 0.5.

In order to fusion (or combine) the results from multiple sources of information, we apply aggregation operators.

On each attribute value $v_{ij} \in V_i, j \in W_i$, one applies some aggregation operators.

We start with the dominant attribute value.

An aggregation operator $O: (P, V_{[m]}) \times (P, V_{[m]}) \rightarrow [0, 1]$ can be extended for the case when working with oversets, undersets, or offsets {see [4]} to

$$O: (P, V_{[m]}) \times (P, V_{[m]}) \rightarrow [\psi, \varphi], \qquad (157)$$

where $\psi < 0$ and $\varphi > 1$.

Let's consider two opposite aggregation operators $O_1$ and $O_2$ (for example the $t_{norm}$ and $t_{conorm}$ respectively). If onto the dominant attribute value we apply $O_1$ and the degree of contradiction between another attribute value $v_{ij}$ and the attribute dominant value is $c_j \subseteq [0, 1]$, then:

$$(1 - c_j)O_1(v_{ij}) + c_j O_2(v_{ij}). \qquad (158)$$

## II.14.3. Classification of Multi-Attribute Plithogenic General Sets

Let $x$ be a generic element, $x \in P$, and all
$$\left(v_{1j}, v_{2j}, \ldots, v_{mj}\right) \in V_1 \times V_2 \times \ldots \times V_m. \qquad (159)$$

### II.14.3.1. Multi-Attribute Plithogenic Single-Valued Set

If $d_{[m]}\left(x_1\left(v_{1j}, v_{2j}, \ldots, v_{mj}\right)\right)$ is a Cartesian product of $m$ single numbers belonging to $[0, 1]^m$, we have a *Single-Valued Multi-Attribute Plithogenic Set*.

### II.14.3.2. Multi-Attribute Plithogenic Interval-Valued Set

If $d_{[m]}\left(x_1\left(v_{1j}, v_{2j}, \ldots, v_{mj}\right)\right)$ is a Cartesian product of $m$ intervals included in $[0, 1]^m$, then we have an *Interval-Valued Multi-Attribute Plithogenic Set*.

### II.14.3.3. Multi-Attribute Plithogenic Hesitant Set

If $d_{[m]}\left(x_1\left(v_{1j}, v_{2j}, \ldots, v_{mj}\right)\right)$ is a Cartesian product of $m$ hesitant sets, each of the form $\left\{n_1, n_2, \ldots, n_{u_j}\right\}$ with $u_j \geq 2$ for each $1 \leq j \leq m$, included in $[0, 1]^m$, then we have a *Hesitant Multi-Attribute Plithogenic Set*.

### II.14.3.4. Multi-Attribute Plithogenic Linguistic Set

If $d_{[m]}\left(x_1\left(v_{1j}, v_{2j}, \ldots, v_{mj}\right)\right) \in L^m$, where

$L = \{l_1, l_2, \ldots, l_h\}$ is a set of $h \geq 2$ labels, then we have a *Linguistic Multi-Attribute Plithogenic Set*.

### II.14.3.5. Multi-Attribute Plithogenic Linguistic-Interval Set

If $d_{[m]}\left(x_1\left(v_{1j}, v_{2j}, \ldots, v_{mj}\right)\right)$ is a Cartesian product of $m$ linguistic-intervals, each linguistic interval of the form $[l_{j1}, l_{j2}]$, with $1 \leq j_1 < j_2 \leq h$, then we have a *Linguistic-Interval Multi-Attribute Plithogenic Set*.

## II.14.4. Multi-Attribute Plithogenic General Number

It is a vector of dimension $\zeta = card(V_{[m]})$, i.e. $(p_1, p_2, \ldots, p_\zeta)$, where all $p_k$ may be:

- crisp (single-valued) numbers in *[0, 1]$^m$*;
- or intervals included in *[0, 1]$^m$*;
- or hesitant subsets of the form $\{n_1, n_2, \ldots, n_u\} \subset [0, 1]^m$;
- or general subsets included in *[0, 1]$^m$*
- or labels included in label set $L = \{l_1, l_2, \ldots, l_h\}^m$;
- or label intervals of the form $[l_{j1}, l_{j2}]^m$, with $1 \leq j_1 < j_2 \leq h$, where the labels $l_{j1}, l_{j2} \in L$;

etc.

Therefore, we may have: *Multi-Attribute Plithogenic Single-Valued / Interval-Valued / Hesitant / General / Linguistic / Interval-Linguistic* etc. *Numbers*.

## II.14.5. Bipolar Multi-Attribute Plithogenic General Set

If $d_{[m]}\left(x_1(v_{1j}, v_{2j}, \ldots, v_{mj})\right) \in \mathcal{P}([-1, 0] \times [0,1])^m$, where $\mathcal{P}([-1, 0] \times [0,1])$ is the power set of $[-1, 0] \times [0,1]$, then we have a *Bipolar Multi-Attribute Plithogenic General Set*, which further may be sub-classified as *Bipolar Multi-Attribute Plithogenic Single-Valued / Interval-Valued / Hesitant / Linguistic / Linguistic-Interval Set*.

# II.15. Example of Uni-Attribute (of 4-Attribute-Values) Plithogenic Single-Valued Fuzzy Set Complement (Negation)

Let's consider that the attribute "size" that has the following values: *small* (the dominant one), *medium*, *big*, *very big*.

| Degrees of contradiction | 0 | 0.50 | 0.75 | 1 |
|---|---|---|---|---|
| Attribute values | small | medium | big | very big |
| Degrees of appurtenance | 0.8 | 0.1 | 0.3 | 0.2 |

*Table 1.*

# II.16. Example of Refinement and Negation of a Uni-Attribute (of 4-Attribute-Values) Plithogenic Single-Valued Fuzzy Set

As a refinement of the above table, let's add the attribute "bigger" as in the below table.

The opposite (negation) of the attribute value "big", which is 75% in contradiction with "small", will be an attribute value which is $1 - 0.75 = 0.25 = 25\%$ in contradiction with "small", so it will be equal to $\frac{1}{2}["small" + "medium"]$. Let's call it "less medium", whose degree of appurtenance is $1 - 0.3 = 0.7$.

If the attribute "size" has other values, small being dominant value:

| Degrees of contradiction | 0 | **0.14** | **0.25** | 0.50 | 0.75 | **0.86** | 1 |
|---|---|---|---|---|---|---|---|
| Attribute values | small | **above small (anti-bigger)** | **less medium (anti-big)** | medium | big | **bigger** | very big |
| Degrees of appurtenance | 0.8 | **0.6** | **0.7** | 0.1 | 0.3 | **0.4** | 0.2 |

*Table 2.*

The opposite (negation) of "bigger" is 1 - 0.86 = 0.14 = 14% in contradiction degree with the dominant attribute value ("small"), so it is in between "small" and "medium", we may say it is included into the attribute-value interval *[small, medium]*, much closer to "small" than to "medium". Let's call is "above small", whose degree of appurtenance is 1 – 0.4 = 0.6.

# II.17. Example of Multi-Attribute (of 24 Attribute-Values) Plithogenic Fuzzy Set Intersection, Union, and Complement

Let $P$ be a plithogenic set, representing the students from a college. Let $x \in P$ be a generic student that is characterized by three attributes:

- <u>altitude</u>, whose values are {tall, short} $\stackrel{\text{def}}{=} \{a_1, a_2\}$;
- <u>weight</u>, whose values are
{obese, fat, medium, thin} $\stackrel{\text{def}}{=} \{w_1, w_2, w_3, w_4\}$;
and
- <u>hair color</u>, whose values are
{blond, reddish, brown} $\stackrel{\text{def}}{=} \{h_1, h_2, h_3\}$.

The multi-attribute of dimension $3$ is
$V_3 = \{(a_i, w_j, h_k) \text{ for all } 1 \leq i \leq 2, 1 \leq j \leq 4, 1 \leq k \leq 3\}$.

The cardinal of $V_3$ is $|V_3| = 2 \times 4 \times 3 = 24$.

The *uni-dimensional attribute contradiction degrees* are:
$c(a_1, a_2) = 1$;
$c(w_1, w_2) = \frac{1}{3}$, $c(w_1, w_3) = \frac{2}{3}$, $c(w_1, w_4) = 1$;
$c(h_1, h_2) = 0.5$, $c(h_1, h_3) = 1$.

Dominant attribute values are: $a_1, w_1$, and $h_1$ respectively for each corresponding uni-dimensional attribute.

Let's use the fuzzy $t_{norm} = a \wedge_F b = ab$, and fuzzy $t_{conorm} = a \vee_F b = a + b - ab$.

## II.17.1. Tri-Dimensional Plithogenic Single-Valued Fuzzy Set Intersection and Union

Let

$$x_A = \begin{Bmatrix} d_A(x, a_i, w_j, h_k), \\ for\ all\ 1 \leq i \leq 2, 1 \leq j \leq 4, 1 \leq k \leq 3 \end{Bmatrix} \quad (160)$$

and

$$x_B = \begin{Bmatrix} d_B(x, a_i, w_j, h_k), \\ for\ all\ 1 \leq i \leq 2, 1 \leq j \leq 4, 1 \leq k \leq 3 \end{Bmatrix}. \quad (161)$$

Then:

$$x_A(a_i, w_j, h_k) \wedge_P x_B(a_i, w_j, h_k) = \begin{cases} (1-c(a_D,a_i)) \cdot [d_A(x,a_D) \wedge_F d_B(x,a_i)] \\ +c(a_D,a_i) \cdot [d_A(x,a_D) \vee_F d_B(x,a_i)], 1 \leq i \leq 2; \\ (1-c(w_D,w_j)) \cdot [d_A(x,w_D) \wedge_F d_B(x,w_j)] \\ +c(w_D,w_j) \cdot [d_A(x,w_D) \vee_F d_B(x,w_j)], 1 \leq j \leq 4; \\ (1-c(h_D,h_k)) \cdot [d_A(x,h_D) \wedge_F d_B(x,h_k)] \\ +c(h_D,h_k) \cdot [d_A(x,h_D) \vee_F d_B(x,h_k)], 1 \leq k \leq 3. \end{cases}$$

(162)

and

$$x_A(a_i, w_j, h_k) \vee_P x_B(a_i, w_j, h_k) = \begin{cases} (1-c(a_D,a_i)) \cdot [d_A(x,a_D) \vee_F d_B(x,a_i)] \\ +c(a_D,a_i) \cdot [d_A(x,a_D) \wedge_F d_B(x,a_i)], 1 \leq i \leq 2; \\ (1-c(w_D,w_j)) \cdot [d_A(x,w_D) \vee_F d_B(x,w_j)] \\ +c(w_D,w_j) \cdot [d_A(x,w_D) \wedge_F d_B(x,w_j)], 1 \leq j \leq 4; \\ (1-c(h_D,h_k)) \cdot [d_A(x,h_D) \vee_F d_B(x,h_k)] \\ +c(h_D,h_k) \cdot [d_A(x,h_D) \wedge_F d_B(x,h_k)], 1 \leq k \leq 3. \end{cases}$$

(163)

Let's have $x_A(d_A(a_1) = 0.8, d_A(w_2) = 0.6, d_A(h_3) = 0.5)$

and

$x_B(d_B(a_1) = 0.4, d_B(w_2) = 0.1, d_B(h_3) = 0.7)$.

We take only one 3-attribute value: $(a_1, w_2, h_3)$, for the other 23 3-attribute values it will be analougsly.

For $x_A \wedge_p x_B$ we calculate for each uni-dimensional attribute separately:

$$[1 - c(a_D, a_1)] \cdot \left[0.8 \wedge_F 0.4\right] + c(a_D, a_1) \cdot \left[0.8 \vee_F 0.4\right]$$
$$= (1 - 0) \cdot [0.8(0.4)] + 0 \cdot \left[0.8 \vee_F 0.4\right] = 0.32;$$

$$\left[1 - c[w_D, w_2]\right] \cdot \left[0.6 \wedge_F 0.1\right] + c(w_D, w_2) \cdot \left[0.6 \vee_F 0.1\right]$$
$$= \left(1 - \frac{1}{3}\right)[0.6(0.1)] + \frac{1}{3}[0.6 + 0.1 - 0.6(0.1)]$$
$$= \frac{2}{3}[0.06] + \frac{1}{3}[0.64] = \frac{0.76}{3} \approx 0.25;$$

$$[1 - c(h_D, h_3)] \cdot \left[0.5 \wedge_F 0.7\right] + c(h_D, h_3) \cdot \left[0.5 \vee_F 0.7\right]$$
$$= [1 - 1] \cdot [0.5(0.7)] + 1 \cdot [0.5 + 0.7 - 0.5(0.7)] = 0 \cdot [0.35] + 0.85$$
$$= 0.85.$$

Whence $x_A \wedge_p x_B(a_1, w_2, h_3) \approx (0.32, 0.25, 0.85)$.

For $x_A \vee_p x_B$ we do similarly:

$$[1 - c(a_D, a_1)] \cdot \left[0.8 \vee_F 0.4\right] + c(a_D, a_1) \cdot \left[0.8 \wedge_F 0.4\right]$$
$$= (1 - 0) \cdot [0.8 + 0.4 - 0.8(0.4)] + 0 \cdot [0.8(0.4)] = 1 \cdot [0.88] + 0 = 0.88;$$

$$\left[1 - c[w_D, w_2]\right] \cdot \left[0.6 \vee_F 0.1\right] + c(w_D, w_2) \cdot \left[0.6 \wedge_F 0.1\right]$$
$$= \left(1 - \frac{1}{3}\right)[0.6 + 0.1 - 0.6(0.1)] + \frac{1}{3}[0.6(0.1)]$$
$$= \frac{2}{3}[0.64] + \frac{1}{3}[0.06] = \frac{1.34}{3} \approx 0.44;$$

$$[1 - c(h_D, h_3)] \cdot \left[0.5 \vee_F 0.7\right] + c(h_D, h_3) \cdot \left[0.5 \wedge_F 0.7\right]$$
$$= [1 - 1] \cdot [0.5 + 0.7 - 0.5(0.7)] + 1 \cdot [0.5(0.7)] = 0 + 0.35 = 0.35.$$

Whence $x_A \vee_p x_B(a_1, w_2, h_3) \approx (0.88, 0.44, 0.35)$.

For $\neg_p x_A(a_1, w_2, h_3) = (d_A(a_2) = 0.8, d_A(w_3) = 0.6, d_A(h_1) = 0.5)$, since the opposite of $a_1$ is $a_2$, the opposite of $w_2$ is $w_3$, and the opposite of $h_3$ is $h_1$.

## II.18. Another Example of Multi-Attribute (of 5 Attribute-Values) Plithogenic Fuzzy Set Complement and Refined Attribute-Value Set

The 5-attribute values plithogenic fuzzy complement (negation) of

$$x \begin{pmatrix} 0 & 0.50 & 0.75 & 0.86 & 1 \\ \text{small,} & \text{medium,} & \text{big,} & \text{bigger,} & \text{very big} \\ 0.8 & 0.1 & 0.3 & 0.4 & 0.2 \end{pmatrix}$$

is

$$\neg_p x \begin{pmatrix} 1-1 & 1-0.86 & 1-0.75 & 1-0.50 & 1-0 \\ \text{anti} - \text{very big,} & \text{anti} - \text{bigger,} & \text{anti} - \text{big,} & \text{anti} - \text{medium,} & \text{anti} - \text{small} \\ 0.2 & 0.4 & 0.3 & 0.1 & 0.8 \end{pmatrix}$$

$$= \neg_p x \begin{pmatrix} 0 & 0.14 & 0.25 & 0.50 & 1 \\ \text{small,} & \text{anti} - \text{bigger,} & \text{anti} - \text{big,} & \text{medium,} & \text{very big} \\ 0.2 & 0.4 & 0.3 & 0.1 & 0.8 \end{pmatrix}$$

$$= \neg_p x \begin{pmatrix} 0 & 0.14 & 0.25 & 0.50 & 1 \\ \text{small,} & \text{above small,} & \text{below medium,} & \text{medium,} & \text{very big} \\ 0.2 & 0.4 & 0.3 & 0.1 & 0.8 \end{pmatrix}.$$

Therefore, the original attribute-value set

$V$ = {small, medium, big, bigger, very big}

has been partially refined into:

$RefinedV$ = {small, *above small, below medium*, medium, very big},

where

*above small, below medium* ∈ *[small, medium]*.

## II.19. Multi-Dimensional Plithogenic Aggregation Set Operators

Let $U$ be a universal set, and $A, B \subset U$ be two plithogenic sets.

Let $\alpha_{[m]} = \alpha_1 \times \alpha_2 \times ... \times \alpha_m$ be an $m$-dimensional attribute, for $m \geq 1$, and each attribute $\alpha_i$, $1 \leq i \leq m$, has $r_i \geq 1$ values:

$$V_i = \{v_{i1}, v_{i2}, ..., v_{ir_i}\}. \quad (164)$$

An element (object) $x \in P$ is characterized by $r_1 \times r_2 \times ... \times r_m \stackrel{\text{def}}{=} r$ values:

$$V = \prod_{i=1}^{n} \{v_{i1}, v_{i2}, ..., v_{ir_i}\} = \{v_{11}, v_{12}, ..., v_{1r_1}\} \times \{v_{21}, v_{22}, ..., v_{2r_2}\} \times ... \times \{v_{m1}, v_{m2}, ..., v_{mr_m}\}$$
$$= \{(v_{1j_1}, v_{2j_2}, ..., v_{mj_m}), 1 \leq j_1 \leq r_1, 1 \leq j_2 \leq r_2, ..., 1 \leq j_m \leq r_m,\}. \quad (165)$$

Let $c(v_{iD}, v_{ik}) = c_{ik} \subseteq [0, 1]^z$ be the degree of contradiction between the attribute $\alpha_i$ dominant value (denoted by $v_{iD}$) and other attribute $\alpha_i$ value (denoted by $v_{ik}$), for $1 \leq i \leq m$ and $1 \leq k \leq r_i$. Where *z = 1* (for fuzzy degree of contradiction), *z = 2* (for intuitionistic fuzzy degree of contradiction), or *z = 3* (for neutrosophic degree of contradiction). And $c_{ik}$, as part of the unit interval [0, 1], may be a subset, or an interval, or a hesitant set, or a single number etc.

We split back the *m*-dimensional attribute into *m* uni-dimensional attributes. And, when applying the plithogenic aggregation operators onto an *m*-uple $(v_{1j_1}, v_{2j_2}, ..., v_{mj_m})$, we separately apply the $t_{norm}$, $t_{conorm}$, or a linear combination of these separately on each of its *m*-components: $v_{1j_1}, v_{2j_2}, ..., v_{mj_m}$.

Let $d_A: P \times V_i \to \mathcal{P}([0, 1])^z$ for each $1 \leq i \leq m$, be the appurtenance fuzzy degree (for $z = 1$), appurtenance intuitionistic fuzzy degree (for $z = 2$), or appurtenance neutrosophic degree (for $z = 3$) function, whereas $\mathcal{P}([0, 1])$ is the power set of the unit interval [0, 1], i.e. all subsets of [0, 1].

Upon the attribute value degree function, the $t_{norm}$, $t_{conorm}$, and their linear combinations are adjusted to the fuzzy sets, intuitionistic fuzzy sets, or neutrosophic sets respectively.

And similarly $d_B: P \times V_i \to \mathcal{P}([0, 1])^z$ for each $1 \leq i \leq m$.

## II.19.1. Multi-Attribute Plithogenic Aggregation Set Operations

Let's use easier notations for two *m*-uple plithogenic numbers:

$$x_A = \{d_A(x, u_1), ..., d_A(x, u_i), ..., d_A(x, u_m)\} \quad (166)$$

and

$$x_B = \{d_B(x, u_1), ..., d_B(x, u_i), ..., d_B(x, u_m)\}. \quad (167)$$

## II.19.2. Multi-Atribute Plithogenic Set Intersection

Let $u_{iD}$ be the attribute $\alpha_i$ dominant value, and $u_i$ be any of the attribute $\alpha_i$ value, $i \in \{1, 2, ..., m\}$.

$$x_A \wedge_p x_B = (1 - c(u_{iD}, u_i)) \cdot \left[ d_A(x, u_{iD}) \wedge_F d_B(x, u_i) \right]$$
$$+ c(u_{iD}, u_i) \cdot \left[ d_A(x, u_{iD}, u_i) \vee_F d_B(x, u_i) \right], \forall i \in \{1, 2, ..., m\}\}.$$

(168)

## II.19.3. Multi-Atribute Plithogenic Set Union

$$x_A \vee_p x_B = (1 - c(u_{iD}, u_i)) \cdot [d_A(x, u_{iD}) \vee_F d_B(x, u_i)]$$
$$+ c(u_{iD}, u_i) \cdot [d_A(x, u_{iD}, u_i) \wedge_F d_B(x, u_i)], \forall i \in \{1, 2, ..., m\}\}.$$
(169)

## II.19.4. Multi-Atribute Plithogenic Set Complement (Negation)

Without losing generality, we assume the attribute value contradiction degrees are:

$$c(u_{1D}, u_1), \dots, c(u_{iD}, u_i), \dots, c(u_{mD}, u_m). \quad (170)$$

The plithogenic element attributes' values:

$$\{u_1, \dots, u_i, \dots, u_m\} \quad (171)$$

The attribute values' appurtenance degree:

$$\{d_A(x, u_1), \dots, d_A(x, u_i), \dots, d_A(x, u_m)\}. \quad (172)$$

Then, the plithogenic complement (negation) is:

$$1 - c(u_{1D}, u_1), \dots, 1 - c(u_{iD}, u_i), \dots, 1 - c(u_{mD}, u_m)$$
$$anti(u_1), \dots, anti(u_i), \quad \dots, \quad anti(u_m)$$
(173)

or

$$\neg_p x_A = \begin{Bmatrix} d_A(x, anti(u_1)) = d_A(x, u_1), \dots, \\ d_A(x, anti(u_i)) = d_A(x, u_i), \dots, d_A(x, anti(u_m)) = d_A(x, u_m) \end{Bmatrix},$$
(174)

where $anti(u_i), 1 \leq i \leq m$, is the attribute $\propto'_i$ opposite value of $u_i$, or

$$c(u_{iD}, anti(u_i)) = [1 - c(u_{iD}, u_i)]. \quad (175)$$

## II.19.5. Multi-Atribute Plithogenic Set Inclusion (Partial Order)

Assuming a relation of partial order has been defined on $\mathcal{P}([0, 1])^z$, for each z = 1, 2, 3. Then:

$x_A \leq_p x_B$ if and only if:
$$d_A(x, u_i) \leq (1 - c(u_{iD}, u_i)) \cdot d_B(x, u_i), \qquad (176)$$
for $0 \leq c(u_{iD}, u_i) < 0.5$, and
$$d_A(x, u_i) \geq (1 - c(u_{iD}, u_i)) \cdot d_B(x, u_i), \qquad (177)$$
for $c(u_{iD}, u_i) \in [0.5, 1]$ for all $1 \leq i \leq m$.

## II.19.6. Multi-Atribute Plithogenic Set Equality

Similarly, assuming a relation of total order has been defined on $\mathcal{P}([0, 1])^z$, for each z =1, 2, 3. Then:

$x_A =_p x_B$ if and only if $x_A \leq_p x_B$ and $x_B \leq_p x_A$.

## II.20. Uni-Attribute Plithogenic Single-Value Number Operations

Let $A = (a_1, a_2, \dots, a_n)$ and $B = (b_1, b_2, \dots, b_n)$ be two single-valued uni-attribute plithogenic numbers, where $a_1, a_2, \dots, a_n, b_1, b_2, \dots, b_n \in [0, 1], n \geq 1$, and
$$0 \leq \sum_{i=1}^n a_i \leq n \qquad (178)$$
and
$$0 \leq \sum_{i=1}^n b_i \leq n. \qquad (179)$$
Let α be the attribute that has *n* values:
$V = \{v_1, v_2, \dots, v_n\},$

where the attribute dominant value $v_D \equiv v_1$, the single-valued contradiction fuzzy degree $c(v_D, v_1) = c_i \in [0, 1]$ for $i \in \{1, 2, \ldots, n\}$, such that $0 = c_1 \leq c_2 \leq \cdots \leq c_n \leq 1$.

We define for the first time the following plithogenic number operations:

### II.20.1. Plithogenic Single-Value Number Summation:

$$A \oplus B = \left\{ \begin{array}{l} (1 - c_i) \cdot [a_i + b_i - a_i \cdot b_i] + \\ c_i \cdot [a_i \cdot b_i], i \in \{1, 2, \ldots, n\} \end{array} \right\}. \qquad (180)$$

### II.20.2. Plithogenic Single-Value Number Multiplication:

$$A \otimes B = \left\{ \begin{array}{l} (1 - c_i) \cdot [a_i \cdot b_i] + c_i \cdot \\ [a_i + b_i - a_i \cdot b_i], \\ i \in \{1, 2, \ldots, n\} \end{array} \right\}. \qquad (181)$$

### II.20.3. Plithogenic Single-Value Number Scalar Multiplication:

$$\lambda \cdot A = \left\{ \begin{array}{l} (1 - c_i) \cdot [1 - (1 - a_i)^\lambda] + c_i \cdot a_i^\lambda, \\ i \in \{1, 2, \ldots, n\} \end{array} \right\}, \qquad (182)$$

where the real number $\lambda > 0$.

### II.20.4. Plithogenic Single-Value Number Power:

$$A^\lambda = \{(1 - c_i) \cdot a_i^\lambda + c_i \cdot [1 - (1 - a_i)^\lambda]\}, \qquad (183)$$
$i \in \{1, 2, \ldots, n\}$,
where the real number $\lambda > 0$.

## II.20.5. Properties of Uni-Attribute Plithogenic Single-Value Number Operations

These plithogenic number operations are extensions of the fuzzy, intuitionistic fuzzy, and neutrosophic number operations. In addition, the plithogenic number operations use the linear combination of these fuzzy, intuitionistic fuzzy, and neutrosophic operations.

The plithogenic single-value number operations can be extended to interval-valued, hesitant-value, or in general subset-value number operations.

## II.21. Distance and Similarity Measures for Single-Valued Uni-Attribute Plithogenic Numbers

### II.21.1. Plithogenic Distance Measure Axioms

Let U be a universe of discourse, and A, B, C three plithogenic sets included in U.

The plithogenic distance measure is defined in the classical way:

$$D: U \times U \to [0, 1] \qquad (184)$$

such that, for any $A, B, C$ in $U$:

i. $0 \leq D(A, B) \leq 1$;
ii. $D(A, B) = D(B, A)$;
iii. $D(A, B) = 0$
if and only if $A \equiv B$ (i.e. $a_i = b_i$ for all $1 \leq i \leq n$);
iiii. $A \subset B \subset C \Longrightarrow D(A, C) \geq \max\{D(A, B), D(B, C)\}$.

## II.21.2. Plithogenic Similarity Measure Axioms

The similarity measure is defined as:
$$S: P \times P \to [0, 1] \tag{185}$$
such that, for any $A, B, C$ in $P$:

i. $0 \leq S(A, B) \leq 1$;
ii. $S(A, B) = S(B, A)$;
iii. $S(A, B) = 1$
if and only if $A \equiv B$ (i.e. $a_i = b_i$ for all $1 \leq i \leq n$);
iiii. $A \subset B \subset C \Longrightarrow S(A, C) \leq \min\{S(A, B), S(B, C)\}$.

There are many neutrosophic measure functions defined in the literature. We extend for the first time several of them from neutrosophic to the plithogenic environment.

### II.21.2.1. Dice Similarity Plithogenic Number Measure (extended from Ye, 2014)

$$D(A, B) = \frac{2 \cdot \sum_{i=1}^{n} a_i b_i}{\sum_{i=1}^{n}(a_i^2 + b_i^2)}. \tag{186}$$

### II.21.2.2. Cosine Similarity Plithogenic Number Measure (extended from Broumi & Smarandache, 2014)

$$\cos(A, B) = \frac{\sum_{i=1}^{n} a_i b_i}{\sqrt{\sum_{i=1}^{n} a_i^2} \cdot \sqrt{\sum_{i=1}^{n} b_i^2}}. \tag{187}$$

### II.21.2.3. Hamming Plithogenic Number Distance

$$HD(A, B) = \frac{1}{n} \sum_{i=1}^{n} |a_i - b_i|. \tag{188}$$

### II.21.2.3.1. Hamming Similarity Plithogenic Number Measure

$$HS(A, B) = 1 - HD(A, B). \tag{189}$$

### II.21.2.4. Euclidean Plithogenic Number Distance

$$ED(A, B) = \sqrt{\frac{1}{n} \sum_{i=1}^{n} (a_i - b_i)^2}. \tag{190}$$

*II.21.2.4.1. Euclidean Similarity Plithogenic Number Measure*
$$ES(A,B) = 1 - ED(A,B). \tag{191}$$

*II.21.2.5. Jaccard Similarity Plithogenic Number Measure*
$$J(A,B) = \frac{\sum_{i=1}^{n} a_i b_i}{\sum_{i=1}^{n}(a_i^2 + b_i^2 - a_i b_i)}. \tag{192}$$

## II.22. Application of Bi-Attribute Plithogenic Single-Valued Set

Let $\mathcal{U}$ be a universe of discourse, and $P \subset \mathcal{U}$ a plithogenic set.

In a plithogenic set $P$, each element (object) $x \in P$ is characterized by $m \geq 1$ attributes $\alpha_1, \alpha_2, \ldots, \alpha_m$, and each attribute $\alpha_i$, $1 \leq i \leq m$, has $r_i \geq 1$ values:

$$V_i = \{v_{i1}, v_{i2}, \ldots, v_{ir_i}\}.$$

Therefore, the element $x$ is characterized by $r = r_1 \times r_2 \times \ldots \times r_m$ attributes' values.

For example, if the attributes are "color" and "height", and their values (required by the application the experts want to do) are:

$Color = \{green, yellow, red\}$

and

$Height = \{tall, medium\}$,

then the object $x \in P$ is characterized by the Cartesian product

$Color \times Height = $
$\begin{Bmatrix} (green, tall), (green, medium), (yellow, tall), \\ (yellow, medium), (red, tall), (red, medium) \end{Bmatrix}.$

Let's consider the dominant (i.e. the most important, or reference) value of attribute "*color*" be "*green*", and of attribute "*height*" be "*tall*".

The attribute value contradiction fuzzy degrees are:
$c(green, green) = 0$,
$c(green, yellow) = \frac{1}{3}$,
$c(green, red) = \frac{2}{3}$,
$c(tall, tall) = 0$,
$c(tall, medium) = \frac{1}{2}$.

Suppose we have two experts A and B.

Further on, we consider (fuzzy, intuitionistic fuzzy, or neutrosophic) degrees of appurtenance of each attribute value to the set $P$ with respect to experts' criteria.

We consider the single value number fuzzy degrees, for simplicity of the example.

Let $v_i$ be a uni-attribute value and its degree of contradiction with respect to the dominant uni-attribute value $v_D$ be $c(v_D, v_i) \stackrel{\text{def}}{=} c_i$.

Let $d_A(x, v_i)$ be the appurtenance degree of the attribute value $v_i$ of the element $x$ with respect to the set A. And similarly for $d_B(x, v_i)$. Then, we recall the plithogenic aggregation operators with respect to this attribute value $v_i$ that will be employed:

## II.22.1. One-Attribute Value Plithogenic Single-Valued Fuzzy Set Intersection

$$d_A(x, v_i) \wedge_p d_B(x, v_i) =$$
$$(1 - c_i) \cdot [d_A(x, v_i) \wedge_F d_B(x, v_i)]$$
$$+ c_i \cdot [d_A(x, v_i) \vee_F d_B(x, v_i)] \quad (193)$$

## II.22.2. One-Attribute Value Plithogenic Single-Valued Fuzzy Set Union

$$d_A(x, v_i) \vee_p d_B(x, v_i) = (1 - c_i) \cdot$$
$$[d_A(x, v_i) \vee_F d_B(x, v_i)] + c_i \cdot [d_A(x, v_i) \wedge_F d_B(x, v_i)] \quad (193)$$

## II.22.3. One Attribute Value Plithogenic Single-Valued Fuzzy Set Complement (Negation)

$$\neg_p v_i = anti(v_i) = (1 - c_i) \cdot v_i \quad (195)$$
$$\neg_p d_A(x, (1 - c_i)v_i) = d_A(x, v_i) \quad (196)$$

## II.23. Singe-Valued Fuzzy Set Degrees of Appurtenance

*According to Expert A:*
$d_A: \{green, yellow, red; tall, medium\} \to [0, 1]$
One has:
$d_A(green) = 0.6$,
$d_A(yellow) = 0.2$,
$d_A(red) = 0.7$;
$d_A(tall) = 0.8$,
$d_A(medium) = 0.5$.
We summarize as follows:

*According to Expert A:*

| Contradiction Degrees | 0 | $\frac{1}{3}$ | $\frac{2}{3}$ |  | 0 | $\frac{1}{2}$ |
|---|---|---|---|---|---|---|
| Attributes' Values | green | yellow | red |  | tall | medium |
| Fuzzy Degrees | 0.6 | 0.2 | 0.7 |  | 0.8 | 0.5 |

*Table 3.*

*According to Expert B:*

| Contradiction Degrees | 0 | $\frac{1}{3}$ | $\frac{2}{3}$ |  | 0 | $\frac{1}{2}$ |
|---|---|---|---|---|---|---|
| Attributes' Values | green | yellow | red |  | tall | medium |
| Fuzzy Degrees | 0.7 | 0.4 | 0.6 |  | 0.6 | 0.4 |

*Table 4.*

The element

$x\{$ *(green, tall), (green, medium), (yellow, tall),*
*(yellow, medium), (red, tall), (red, medium)* $\} \in P$

with respect to the two experts as above is represented as:
$x_A\{(0.6, 0.8), (0.6, 0.5), (0.2, 0.8), (0.2, 0.5), (0.7, 0.8), (0.7, 0.5)\}$
and
$x_B\{(0.7, 0.6), (0.7, 0.4), (0.4, 0.6), (0.4, 0.4), (0.6, 0.6), (0.6, 0.4)\}$.

In order to find the optimal representation of $x$, we need to intersect $x_A$ and $x_B$, each having six duplets. Actually, we separately intersect the corresponding duplets.

In this example, we take the fuzzy $t_{norm}: a \wedge_F b = ab$ and the fuzzy $t_{conorm}: a \vee_F b = a + b - ab$.

## II.23.1. Application of Uni-Attribute Value Plithogenic Single-Valued Fuzzy Set Intersection

Let's compute $x_A \wedge_p x_B$.

$\phantom{()}0 \quad 0 \quad\quad 0 \quad 0$  {*degrees of contradictions*}
$(0.6, 0.8) \wedge_p (0.7, 0.6) = (0.6 \wedge_p 0.7, 0.8 \wedge_p 0.6)$
$\phantom{(0.6, 0.8) \wedge_p (0.7, 0.6)} = (0.6 \cdot 0.7, 0.8 \cdot 0.6) = (0.42, 0.48),$

where above each duplet we wrote the degrees of contradictions of each attribute value with respect to their correspondent dominant attribute value. Since they were zero, $\wedge_p$ coincided with $\wedge_F$.

{*the first raw below 0 ½ and again 0 ½ represents the contradiction degrees*}

$\begin{pmatrix} 0 & \frac{1}{2} \\ 0.6, & 0.5 \end{pmatrix} \wedge_p \begin{pmatrix} 0 & \frac{1}{2} \\ 0.7, & 0.4 \end{pmatrix} = (0.6 \wedge_p 0.7, 0.5 \wedge_p 0.4)$
$\phantom{xxxxxxxxxxx} = (0.6 \cdot 0.7, (1 - 0.5) \cdot [0.5 \wedge_F 0.4] + 0.5$
$\phantom{x} \cdot [0.5 \vee_F 0.4])$
$\phantom{xxxxxxxxxxx} = (0.42, 0.5[0.2] + 0.5[0.5 + 0.4 - 0.5 \cdot 0.4])$
$\phantom{xxxxxxxxxxx} = (0.42, 0.45).$

$\begin{pmatrix} \frac{1}{3}, & 0 \\ 0.2 & 0.8 \end{pmatrix} \wedge_p \begin{pmatrix} \frac{1}{3}, & 0 \\ 0.4 & 0.6 \end{pmatrix} = (0.2 \wedge_p 0.4, 0.8 \wedge_p 0.6)$
$\phantom{xxxxxxxxxxx} = \left(\left\{1 - \frac{1}{3}\right\} \cdot [0.2 \wedge_F 0.4] + \left\{\frac{1}{3}\right\} \cdot [0.2 \vee_F 0.4], 0.8 \right.$
$\phantom{x} \left. \cdot 0.6 \right) \approx (0.23, 0.48).$

$\begin{pmatrix} \frac{1}{3}, & \frac{1}{2} \\ 0.2 & 0.5 \end{pmatrix} \wedge_p \begin{pmatrix} \frac{1}{3}, & \frac{1}{2} \\ 0.4 & 0.4 \end{pmatrix} = (0.2 \wedge_p 0.4, 0.5 \wedge_p 0.4)$

(they were computed above)

$$\approx (0.23, 0.45).$$

$$\begin{pmatrix} \frac{2}{3}, & 0 \\ 0.7 & 0.8 \end{pmatrix} \wedge_p \begin{pmatrix} \frac{2}{3}, & 0 \\ 0.6 & 0.6 \end{pmatrix} = \left(0.7 \wedge_p 0.8, 0.8 \wedge_p 0.6\right)$$

$$= \left(\left\{1 - \frac{2}{3}\right\} \cdot [0.7 \wedge_F 0.6] + \left\{\frac{2}{3}\right\} \right.$$
$$\left. \cdot [0.7 \vee_F 0.6], 0.48\right)$$

(the second component was computed above)

$$= \left(\frac{1}{3}[0.7 \cdot 0.6] + \frac{2}{3}[0.7 + 0.6 - 0.7 \cdot 0.6], 0.48\right) \approx (0.73, 0.48).$$

And the last duplet:

$$\begin{pmatrix} \frac{2}{3}, & \frac{1}{2} \\ 0.7 & 0.5 \end{pmatrix} \wedge_p \begin{pmatrix} \frac{2}{3}, & \frac{1}{2} \\ 0.6 & 0.4 \end{pmatrix} = \left(0.7 \wedge_p 0.6, 0.5 \wedge_p 0.4\right)$$
$$\approx (0.73, 0.45)$$

(they were computed above).

Finally:

$$x_A \wedge_p x_B$$
$$\approx \left\{ \begin{matrix} (0.42, 0.48), (0.42, 0.45), (0.23, 0.48), (0.23, 0.45), \\ (0.73, 0.48), (0.73, 0.45) \end{matrix} \right\},$$

or, after the intersection of the experts' opinions $A \wedge_P B$, we summarize the result as:

| Contradiction Degrees | 0 | $\frac{1}{3}$ | $\frac{2}{3}$ | | 0 | $\frac{1}{2}$ |
|---|---|---|---|---|---|---|
| Attributes' Values | green | yellow | red | | tall | medium |
| Fuzzy Degrees of Expert A for x | 0.6 | 0.2 | 0.7 | | 0.8 | 0.5 |
| Fuzzy Degrees of Expert B for x | 0.7 | 0.4 | 0.6 | | 0.6 | 0.4 |
| Fuzzy Degrees of $x_A \wedge_p x_B$ | 0.42 | 0.23 | 0.73 | | 0.48 | 0.45 |
| Fuzzy Degrees of $x_A \vee_p x_B$ | 0.88 | 0.37 | 0.57 | | 0.92 | 0.45 |

*Table 5.*

## II.23.2. Application of Uni-Attribute Value Plithogenic Single-Valued Fuzzy Set Union

We separately compute for each single attribute value:
$d_A^F(x, green) \vee_p d_B^F(x, green) = 0.6 \vee_p 0.7$
$= (1 - 0) \cdot [0.6 \vee_F 0.7] + 0 \cdot [0.6 \wedge_F 0.7]$
$= 1 \cdot [0.6 + 0.7 - 0.6 \cdot 0.7] + 0 = 0.88.$
$d_A^F(x, yellow) \vee_p d_B^F(x, yellow) = 0.2 \vee_p 0.4$
$= \left(1 - \frac{1}{3}\right) \cdot [0.2 \vee_F 0.4] + \frac{1}{3} \cdot [0.2 \wedge_F 0.4]$
$= \frac{2}{3} \cdot (0.2 + 0.4 - 0.2 \cdot 0.4) + \frac{1}{3}(0.2 \cdot 0.4)$
$\approx 0.37.$

$$d_A^F(x, red) \vee_p d_B^F(x, red) = 0.7 \vee_p 0.6$$
$$= \left\{1 - \frac{2}{3}\right\} \cdot [0.7 \vee_F 0.6] + \frac{2}{3} \cdot [0.7 \wedge_F 0.6]$$
$$= \frac{1}{3} \cdot (0.7 + 0.6 - 0.7 \cdot 0.6) + \frac{2}{3}(0.7 \cdot 0.6)$$
$$\approx 0.57.$$
$$d_A^F(x, tall) \vee_p d_B^F(x, tall) = 0.8 \vee_p 0.6$$
$$= (1 - 0) \cdot (0.8 + 0.6 - 0.8 \cdot 0.6) + 0$$
$$\cdot (0.8 \cdot 0.6) = 0.92.$$
$$d_A^F(x, medium) \vee_p d_B^F(x, medium) = 0.5 \vee_p 0.4$$
$$= \frac{1}{2}(0.5 + 0.4 - 0.5 \cdot 0.4) + \frac{1}{2} \cdot (0.5 \cdot 0.4)$$
$$= 0.45.$$

### II.23.3. Properties of Plithogenic Single-Valued Set Operators in Applications

1) When the attribute value contradiction degree with respect to the corresponding dominant attribute value is 0 (zero), one simply use the fuzzy intersection:
$$d_{A \wedge_p B}(x, green) = d_A(x, green) \wedge_F d_B(x, green) = 0.6 \cdot 0.7 = 0.42,$$
and
$$d_{A \wedge_p B}(x, tall) = d_A(x, tall) \wedge_F d_B(x, tall) = 0.8 \cdot 0.6 = 0.48.$$

2) But, if the attribute value contradiction degree with respect to the corresponding dominant attribute value is different from 0 and from 1, the result of the plithogenic intersection is between the results of fuzzy $t_{norm}$ and fuzzy $t_{conorm}$. Examples:

$$d_A(x, yellow) \wedge_F d_B(x, yellow) = 0.2 \wedge_F 0.4 = 0.2 \cdot 0.4$$
$$= 0.08 \ (t_{norm}),$$
$$d_A(x, yellow) \vee_F d_B(x, yellow) = 0.2 \vee_F 0.4$$
$$= 0.2 + 0.4 - 0.2 \cdot 0.4 = 0.52 \ (t_{conorm});$$

while

$$d_A(x, yellow) \wedge_p d_B(x, yellow) = 0.23 \in [0.08, 0.52]$$

{or 0.23 ≈ 0.2266… = (2/3)×0.08 + (1/3)×0.52, i.e. a linear combination of $t_{norm}$ and $t_{conorm}$}.

Similarly:

$$d_A(x, red) \wedge_p d_B(x, red) = 0.7 \wedge_F 0.6 = 0.7 \cdot 0.6$$
$$= 0.42 \ (t_{norm}),$$
$$d_A(x, red) \vee_p d_B(x, red) = 0.7 \vee_F 0.6 = 0.7 + 0.6 - 0.7 \cdot$$
$$0.6 = 0.88 \ (t_{conorm});$$

while

$$d_A(x, red) \wedge_p d_B(x, red) = 0.57 \in [0.42, 0.88]$$

{linear combination of $t_{norm}$ and $t_{conorm}$}.

And

$$d_A(x, medium) \wedge_F d_B(x, medium) = 0.5 \wedge_F 0.4 = 0.5 \cdot 0.4$$
$$= 0.20,$$
$$d_A(x, medium) \vee_F d_B(x, medium) = 0.5 \vee_F 0.4$$
$$= 0.5 + 0.4 - 0.5 \cdot 0.4 = 0.70,$$

while

$d_A(x, medium) \wedge_p d_B(x, medium) = 0.45$, which is just in the middle (because "medium" contradiction degree is $\frac{1}{2}$) of the interval $[0.20, 0.70]$.

## II.24. Single-Valued Intuitionistic Fuzzy Set Degree of Appurtenance

| Contradiction Degrees | | 0 | $\frac{1}{3}$ | $\frac{2}{3}$ | | 0 | $\frac{1}{2}$ |
|---|---|---|---|---|---|---|---|
| Attributes' Values | | green | yellow | red | | tall | medium |
| Intuitionistic Fuzzy Degrees | Expert A | (0.4, 0.5) | (0.1, 0.2) | (0, 0.3) | | (0.8, 0.2) | (0.4, 0.5) |
| | Expert B | (0.6, 0.3) | (0.4, 0.3) | (0.2, 0.5) | | (0.6, 0.1) | (0.5, 0.3) |
| | Experts $A \wedge_p B$ | (0.24, 0.65) | (0.18, 0.31) | (0.13, 0.32) | | (0.48, 0.28) | (0.45, 0.40) |
| | Experts $A \vee_p B$ | (0.76, 0.15) | (0.32, 0.19) | (0.07, 0.48) | | (0.92, 0.02) | (0.45, 0.40) |

*Table 6.*

## II.24.1. One-Attribute Value Plithogenic Single-Valued Intuitionistic Fuzzy Set Intersection

$$\{degrees\ of\ contradictions\}$$

$$d_A^{IF}(x, green) \wedge_p d_B^{IF}(x, green) = \overset{0}{(0.4, 0.5)} \wedge_p \overset{0}{(0.6, 0.3)}$$
$$= \left(0.4 \wedge_p 0.6, 0.5 \vee_p 0.3\right) =$$
$$= (1 \cdot [0.4 \cdot 0.6] + 0 \cdot [0.4 + 0.6 - 0.4 \cdot 0.6], 0$$
$$\cdot [0.5 \cdot 0.3] + 1 \cdot [0.5 + 0.3 - 0.5 \cdot 0.3])$$
$$= (0.24, 0.65).$$

$$d_A^{IF}(x, yellow) \wedge_p d_B^{IF}(x, yellow) = \overset{1/3}{(0.1, 0.2)} \wedge_p \overset{1/3}{(0.4, 0.3)}$$
$$= \left(0.1 \wedge_p 0.4, 0.2 \vee_p 0.3\right)$$
$$= \left(\left\{1 - \frac{1}{3}\right\} \cdot [0.1 \wedge_F 0.4] + \left\{\frac{1}{3}\right\}\right.$$
$$\cdot [0.1 \vee_F 0.4], \left\{1 - \frac{1}{3}\right\} \cdot [0.2 \vee_F 0.3]\right) + \left\{\frac{1}{3}\right\}$$
$$\cdot [0.2 \wedge_F 0.3]$$
$$= \left(\frac{2}{3} \cdot [0.1 \cdot 0.4] + \frac{1}{3} \cdot [0.1 + 0.4 - 0.1 \cdot 0.4],\right.$$
$$\left.\frac{2}{3} \cdot [0.2 + 0.3 - 0.2 \cdot 0.3]\right) + \frac{1}{3} \cdot [0.2 \cdot 0.3]$$
$$\approx (0.18, 0.31).$$

$$d_A^{IF}(x, red) \wedge_p d_B^{IF}(x, red) = (0, 0.3) \wedge_p^{2/3} (0.2, 0.5)^{2/3}$$
$$= \left(0 \wedge_p 0.2, 0.3 \vee_p 0.5\right)$$
$$= \left(\left\{1 - \frac{2}{3}\right\} \cdot [0 \wedge_F 0.2] + \left\{\frac{2}{3}\right\} \cdot [0 \vee_F 0.2], \left\{1 - \frac{2}{3}\right\}\right.$$
$$\left. \cdot [0.3 \vee_F 0.5] + \left\{\frac{2}{3}\right\} \cdot [0.3 \wedge_F 0.5] + \right)$$
$$= \left(\frac{1}{3} \cdot [0 \cdot 0.2] + \frac{2}{3} \cdot [0 + 0.2 - 0 \cdot 0.2], \frac{1}{3}\right.$$
$$\left. \cdot [0.3 + 0.5 - 0.3 \cdot 0.5] + \frac{2}{3} \cdot [0.3 \cdot 0.5]\right)$$
$$\approx (0.13, 0.32).$$

$$d_A^{IF}(x, tall) \wedge_p d_B^{IF}(x, tall) = (0.8, 0.2) \wedge_p^{0} (0.6, 0.1)^{0}$$
$$= \left(0.8 \wedge_p 0.6, 0.2 \vee_p 0.1\right)$$
$$= (\{1 - 0\} \cdot [0.8 \wedge_F 0.6] + \{0\}$$
$$\cdot [0.8 \vee_F 0.6], \{1 - 0\} \cdot [0.2 \vee_F 0.1] + \{0\}$$
$$\cdot [0.2 \wedge_F 0.1])$$
$$= (1 \cdot [0.8 \cdot 0.6] + 0 \cdot [0.8 + 0.6 - 0.8 \cdot 0.6], 1$$
$$\cdot [0.2 + 0.1 - 0.2 \cdot 0.1] + 0 \cdot [0.2 \cdot 0.1])$$
$$= (0.48, 0.28).$$

$$d_A^{IF}(x, medium) \wedge_p d_B^{IF}(x, medium) = (0.4, 0.5) \wedge_p (0.5, 0.3)$$
$$= \left(0.4 \wedge_p 0.5, 0.5 \vee_p 0.3\right)$$
$$= \left(\left\{1 - \frac{1}{2}\right\} \cdot [0.4 \wedge_F 0.5] + \left\{\frac{1}{2}\right\}\right.$$
$$\cdot [0.4 \vee_F 0.5], \left\{1 - \frac{1}{2}\right\} \cdot [0.5 \vee_F 0.3] + \left\{\frac{1}{2}\right\}$$
$$\left.\cdot [0.5 \wedge_p 0.3]\right)$$
$$= \left(\frac{1}{2} \cdot [0.4 \cdot 0.5] + \frac{1}{2} \cdot [0.4 + 0.5 - 0.4 \cdot 0.5], \frac{1}{2}\right.$$
$$\left.\cdot [0.5 + 0.3 - 0.5 \cdot 0.3] + \frac{1}{2} \cdot [0.5 \cdot 0.3]\right)$$
$$= (0.45, 0.40).$$

## II.24.2. One-Attribute Value Plithogenic Single-Valued Intuitionistic Fuzzy Set Union

$$d_A^{IF}(x, green) \vee_p d_B^{IF}(x, green) = (0.4, 0.5) \vee_p (0.6, 0.3)$$
$$= \left(0.4 \vee_p 0.6, 0.5 \wedge_p 0.3\right)$$
$$= (\{1 - 0\} \cdot [0.4 \vee_F 0.6] + \{0\}$$
$$\cdot [0.4 \wedge_F 0.6], \{1 - 0\} \cdot [0.5 \wedge_F 0.3] + \{0\}$$
$$\cdot [0.5 \vee_F 0.3])$$
$$= (1 \cdot [0.4 + 0.6 - 0.4 \cdot 0.6] + 0 \cdot [0.4 \cdot 0.6], 0$$
$$\cdot [0.5 \cdot 0.3] + 1 \cdot [0.5 + 0.3 - 0.5 \cdot 0.3])$$
$$= (0.76, 0.15).$$

$$d_A^{IF}(x, yellow) \vee_p d_B^{IF}(x, yellow) = (0.1, 0.2) \overset{1/3}{\vee_p} (0.4, 0.3)$$
$$= \left(0.1 \vee_p 0.4,\, 0.2 \wedge_p 0.3\right)$$
$$= \left(\left\{1 - \frac{1}{3}\right\} \cdot [0.1 \vee_F 0.4] + \left\{\frac{1}{3}\right\} \right.$$
$$\cdot [0.1 \wedge_F 0.4], \left\{1 - \frac{1}{3}\right\} \cdot [0.2 \wedge_F 0.3] + \left\{\frac{1}{3}\right\}$$
$$\left. \cdot [0.2 \vee_F 0.3]\right)$$
$$= \left(\frac{2}{3} \cdot [0.1 + 0.4 - 0.1 \cdot 0.4] + \frac{1}{3} \cdot [0.1 \cdot 0.4],\, \frac{2}{3}\right.$$
$$\left. \cdot [0.2 \cdot 0.3] + \frac{1}{3} \cdot [0.2 + 0.3 - 0.2 \cdot 0.3]\right)$$
$$\approx (0.32, 0.19).$$

$$d_A^{IF}(x, red) \vee_p d_B^{IF}(x, red) = (0, 0.3) \overset{2/3}{\vee_p} (0.2, 0.5)$$
$$= \left(0 \vee_p 0.2,\, 0.3 \wedge_p 0.5\right)$$
$$= \left(\left\{1 - \frac{2}{3}\right\} \cdot [0 \vee_p 0.2] + \left\{\frac{2}{3}\right\} \cdot [0 \wedge_p 0.2], \left\{1 - \frac{2}{3}\right\} \right.$$
$$\left. \cdot [0.3 \wedge_p 0.5] + \left\{\frac{2}{3}\right\} \cdot [0.3 \vee_p 0.5]\right)$$
$$= \left(\frac{1}{3}[0 + 0.2 - 0 \cdot 0.2] + \frac{2}{3}[0 \cdot 0.2],\, \frac{1}{3}[0.3 \cdot 0.5]\right.$$
$$\left. + \frac{2}{3} \cdot [0.3 + 0.5 - 0.3 \cdot 0.5]\right) \approx (0.07, 0.48).$$

$$d_A^{IF}(x, tall) \vee_p d_B^{IF}(x, tall) = (0.8, 0.2) \overset{0}{\vee_p} (0.6, 0.1)$$
$$= \left(0.8 \vee_p 0.6, 0.2 \wedge_p 0.1\right)$$
$$= (\{1 - 0\} \cdot [0.8 \vee_F 0.6] + \{0\}$$
$$\cdot [0.8 \wedge_F 0.6], \{1 - 0\} \cdot [0.2 \wedge_F 0.1] + \{0\}$$
$$\cdot [0.2 \vee_F 0.1])$$
$$= (1 \cdot [0.8 + 0.6 - 0.8 \cdot 0.6] + 0 \cdot [0.8 \cdot 0.6], 1$$
$$\cdot [0.2 \cdot 0.1] + 0 \cdot [0.2 + 0.1 - 0.2 - 0.1])$$
$$= (0.92, 0.02).$$

$$d_A^{IF}(x, medium) \vee_p d_B^{IF}(x, medium) = (0.4, 0.5) \overset{1/2}{\vee_p} (0.5, 0.3)$$
$$= \left(0.4 \vee_p 0.5, 0.5 \wedge_p 0.3\right)$$
$$= \left(\left\{1 - \frac{1}{2}\right\} \cdot [0.4 \vee_p 0.5] + \left\{\frac{1}{2}\right\}\right.$$
$$\left. \cdot [0.4 \wedge_F 0.5], \left\{1 - \frac{1}{2}\right\} \cdot [0.5 \wedge_F 0.3] + \left\{\frac{1}{2}\right\}\right.$$
$$\left. \cdot [0.5 \vee_F 0.3]\right)$$
$$= \left(\frac{1}{2} \cdot [0.4 + 0.5 - 0.4 \cdot 0.5] + \frac{1}{2} \cdot [0.4 \cdot 0.5], \frac{1}{2}\right.$$
$$\left. \cdot [0.5 \cdot 0.3] + \frac{1}{2} \cdot [0.5 + 0.3 - 0.5 - 0.3]\right)$$
$$= (0.45, 0.40).$$

## II.25. Single Valued Neutrosophic Set Degree of Appurtenance

| Contradiction | Degrees | 0 | $\frac{1}{3}$ | $\frac{2}{3}$ | | 0 | $\frac{1}{2}$ |
|---|---|---|---|---|---|---|---|
| Attributes' | Values | green | yellow | red | | tall | medium |
| Neutrosophic Degrees | Expert A | 0  $\frac{1}{2}$  1 (0.4, 0.1, 0.5) | (0.3, 0.6, 0.2) | (0.2, 0.1, 0.4) | | (0.8, 0.3, 0.1) | (0.6, 0.2, 0.3) |
| | Expert B | (0.5, 0.2, 0.4) | (0.4, 0.1, 0.3) | (0.3, 0.4, 0.2) | | (0.7, 0.1, 0.6) | (0.5, 0.1, 0.3) |
| | Experts $A \wedge_p B$ | (0.20, 0.15, 0.70) | (0.27, 0.35, 0.31) | (0.31, 0.25, 0.23) | | (0.56, 0.20, 0.64) | (0.55, 0.15, 0.30) |
| | Experts $A \vee_p B$ | (0.70, 0.15, 0.20) | (0.43, 0.35, 0.19) | (0.19, 0.25, 0.37) | | (0.94, 0.20, 0.06) | (0.55, 0.15, 0.30) |

*Table 7.*

## II.25.1. One-Attribute Value Plithogenic Single-Valued Neutrosophic Set Intersection

$d_A^N(x, green) \wedge_p d_B^N(x, green)$
$$= (0.4, 0.1, 0.5) \wedge_p (0.5, 0.2, 0.4)$$
$$= \left(0.4 \wedge_p 0.5, \left\{1 - \frac{1}{2}\right\} \cdot (0.1 \wedge_p 0.2) + \left\{\frac{1}{2}\right\} \right.$$
$$\left. \cdot (0.1 \vee_p 0.2), 0.5 \vee_p 0.4\right)$$
$$= \left(0.4 \wedge_p 0.5, \frac{1}{2} \cdot [0.1 \wedge_p 0.2] + \frac{1}{2} \right.$$
$$\left. \cdot [0.1 \vee_p 0.2], 0.4 \vee_p 0.5\right)$$

{ Using first the *interior neutrosophic contradiction degrees* (between the neutrosophic components *T*, *I*, and *F*):

$$\begin{matrix} 0 & \frac{1}{2} & 1 \\ T, & I, & F \end{matrix} \}$$

$$= \left(\{1-0\} \cdot [0.4 \wedge_F 0.5] + \{0\} \cdot [0.4 \vee_F 0.5], \frac{1}{2} \cdot [0.1 \wedge_p 0.2] \right.$$
$$+ \frac{1}{2} \cdot [0.1 \vee_p 0.2], \{1-0\} \cdot [0.5 \vee_F 0.4] + \{0\}$$
$$\left. \cdot [0.5 \wedge_F 0.4]\right) =$$
$$= \left(1 \cdot [0.4 \cdot 0.5] + 0 \cdot [0.4 + 0.5 - 0.4 \cdot 0.5], \frac{1}{2} \cdot [0.1 \wedge_p 0.2] \right.$$
$$+ \frac{1}{2} \cdot [0.1 \vee_p 0.2], (1-0) \cdot [0.5 + 0.4 - 0.5 \cdot 0.4]$$
$$\left. + 0 \cdot [0.5 \cdot 0.4]\right)$$

$$= \left(0.20, \frac{1}{2} \cdot [0.1 \wedge_F 0.2] + \frac{1}{2} \cdot [0.1 \vee_F 0.2], 0.70\right)$$

$$= \left(0.20, \frac{1}{2}(0.1 \cdot 0.2) + \frac{1}{2}\right.$$
$$\left. \cdot [0.1 + 0.2 - 0.1 \cdot 0.2], 0.70\right)$$
$$= (0.20, 0.15, 0.70).$$

$d_A^N(x, yellow) \wedge_p d_B^N(x, yellow)$

$$= \left(0.3, \begin{matrix}\frac{1}{3}\\ 0.6,\end{matrix} 0.2\right) \wedge_p \left(0.4, \begin{matrix}\frac{1}{3}\\ 0.1,\end{matrix} 0.3\right)$$

$$= \left(0.3 \wedge_p 0.4, \frac{1}{2} \cdot [0.6 \wedge_p 0.1] + \frac{1}{2}\right.$$
$$\left. \cdot [0.6 \vee_p 0.1], 0.2 \vee_p 0.3\right)$$

$$\left\{\begin{matrix}\text{one firstly used the interior neutrosophic contradiction degrees:}\\ c(T, I) = \frac{1}{2}, c(T, F) = 1.\end{matrix}\right\}$$

$$= \left(\left\{1 - \frac{1}{3}\right\} \cdot [0.3 \wedge_F 0.4] + \left\{\frac{1}{3}\right\} \cdot [0.3 \vee_F 0.4], \frac{1}{2} \cdot [0.6 \cdot 0.1] + \frac{1}{2}\right.$$
$$\cdot [0.6 + o.1 - 0.6 \cdot 0.1], \left\{1 - \frac{1}{3}\right\} \cdot [0.2 \vee_F 0.3]$$
$$\left. + \left\{\frac{1}{3}\right\} \cdot [0.2 \wedge_F 0.3]\right)$$

$$= \left(\frac{2}{3} \cdot [0.3 \cdot 0.4] + \frac{1}{3}\right.$$
$$\cdot [0.3 + 0.4 - 0.3 \cdot 0.4], 0.35, \frac{2}{3}$$
$$\left. \cdot [0.2 + 0.3 - 0.2 \cdot 0.3] + \frac{1}{3} \cdot [0.2 \cdot 0.3]\right)$$
$$\approx (0.27, 0.35, 0.31).$$

$$d_A^N(x, red) \wedge_p d_B^N(x, red) = (0.2, 0.1, 0.4) \wedge_p^{2/3} (0.3, 0.4, 0.2)$$
$$= \left(0.2 \wedge_p 0.3, 0.1 \vee_p 0.4, 0.4 \vee_p 0.2\right)$$
$$= \left(\left\{1 - \frac{2}{3}\right\} \cdot [0.2 \wedge_F 0.3] + \left\{\frac{2}{3}\right\} \cdot [0.2 \vee_p 0.3]\right), \frac{1}{2}$$
$$\cdot [I_1 \wedge_F I_2 + I_1 \vee_F I_2], \{\text{according to Theorem 5}\}$$
$$\left\{1 - \frac{2}{3}\right\} \cdot [0.4 \vee_F 0.2] + \left\{\frac{2}{3}\right\} \cdot [0.4 \wedge_F 0.2])$$
$$= \left(\frac{1}{3} \cdot [0.2 \cdot 0.3] + \frac{2}{3} \cdot [0.2 + 0.3 - 0.2 \cdot 0.3], \frac{1}{2}\right.$$
$$\cdot [0.1 \cdot 0.4 + 0.1 + 0.4 - 0.1 \cdot 0.4], \frac{1}{3}$$
$$\left. \cdot [0.4 + 0.2 - 0.4 \cdot 0.2] + \frac{2}{3} \cdot [0.4 \cdot 0.2]\right)$$
$$\approx (0.31, 0.25, 0.23).$$

{The degree of contradiction is $2/3 > 0.5$.}

$$d_A^N(x, tall) \wedge_p d_B^N(x, tall) = (0.8, 0.3, 0.1) \wedge_p (0.7, 0.1, 0.6)$$
$$= \left(0.8 \wedge_p 0.7, 0.3 \vee_p 0.1, 0.1 \vee_p 0.6\right)$$
$$= \left(0.8 \wedge_F 0.7, \frac{1}{2}\right.$$
$$\left. \cdot (0.3 \wedge_F 0.1 + 0.3 \vee_F 0.1), 0.1 \vee_F 0.6\right)$$

(since the exterior degree of contradiction is zero)

$$= \left(0.8 \cdot 0.7, \frac{1}{2} \cdot (0.3 \cdot 0.1 + 0.3 + 0.1 - 0.3 \cdot 0.1), 0.1 + 0.6\right.$$
$$\left. - 0.1 \cdot 0.6\right) = (0.56, 0.20, 0.64).$$

$$d_A^N(x, medium) \wedge_p d_B^N(x, medium)$$
$$= (0.6, 0.2, 0.3) \wedge_p (0.5, 0.1, 0.3)$$
$$= (0.6 \wedge_p 0.5, 0.2 \vee_p 0.1, 0.3 \vee_p 0.3)$$
$$= \left(\frac{1}{2} \cdot [0.6 \cdot 0.5] + \frac{1}{2} \cdot [0.6 + 0.5 - 0.6 \cdot 0.5], \frac{1}{2}\right.$$
$$\cdot [0.2 \cdot 0.1 + 0.2 + 0.1 - 0.2 \cdot 0.1], \frac{1}{2} \cdot [0.3 \cdot 0.3]$$
$$\left. + \frac{1}{2} \cdot [0.3 + 0.3 - 0.3 \cdot 0.3]\right)$$
$$= (0.55, 0.15, 0.30).$$

{Since the degree of contradiction is $1/2$.}

## II.25.2. One-Attribute Value Plithogenic Single-Valued Neutrosophic Set Union

$$d_A^N(x, green) \vee_p d_B^N(x, green)$$
$$= \left(\underset{0.4,}{0} \ 0.1, \ 0.5\right) \vee_p \left(\underset{0.5,}{0} \ 0.2, \ 0.4\right)$$
$$= (0.4 \vee_p 0.5, 0.1 \wedge_p 0.2, 0.5 \wedge_p 0.4)$$

{since the degree of contradiction is zero}
$$= \left(0.4 + 0.5 - 0.4\right.$$
$$\cdot 0.5, \frac{1}{2}(0.1 \cdot 0.2 + 0.1 + 0.2 - 0.1 \cdot 0.2), 0.5$$
$$\left. \cdot 0.4\right) = (0.70, 0.15, 0.20).$$

$d_A^N(x, yellow) \vee_p d_B^N(x, yellow)$

$$= \begin{pmatrix} \frac{1}{3} \\ 0.3, \quad 0.6, \quad 0.2 \end{pmatrix} \vee_p \begin{pmatrix} \frac{1}{3} \\ 0.4, \quad 0.1, \quad 0.3 \end{pmatrix}$$

$$= \left( 0.3 \vee_p 0.4, 0.6 \wedge_p 0.1, 0.2 \wedge_p 0.3 \right)$$

$$= \left( \left\{ 1 - \frac{1}{3} \right\} \cdot [0.3 \vee_F 0.4] + \left\{ \frac{1}{3} \right\} \cdot [0.3 \wedge_F 0.4], \frac{1}{2} \right.$$
$$\cdot [0.6 \wedge_F 0.1 + 0.6 \vee_F 0.1], \left\{ 1 - \frac{1}{3} \right\} \cdot [0.2 \wedge_F 0.3]$$
$$\left. + \left\{ \frac{1}{3} \right\} \cdot [0.2 \vee_F 0.3] \right)$$

$$= \left( \frac{2}{3} \cdot [0.3 + 0.4 - 0.3 \cdot 0.4] + \frac{1}{3} \cdot [0.3 \cdot 0.4], \frac{1}{2} \right.$$
$$\cdot [0.6 \cdot 0.1 + 0.6 + 0.1 - 0.6 \cdot 0.1], \frac{2}{3} \cdot [0.2 \cdot 0.3]$$
$$\left. + \frac{1}{3} \cdot [0.2 + 0.3 - 0.2 \cdot 0.3] \right)$$

$$\approx (0.43, 0.35, 0.19).$$

$$d_A^N(x, red) \vee_p d_B^N(x, red)$$
$$= \begin{pmatrix} & \frac{2}{3} & \\ 0.2, & 0.1, & 0.4 \end{pmatrix} \vee_p \begin{pmatrix} & \frac{2}{3} & \\ 0.3, & 0.4, & 0.2 \end{pmatrix}$$
$$= (0.2 \vee_p 0.3, 0.1 \wedge_p 0.4, 0.4 \wedge_p 0.2)$$
$$= \left(\left\{1 - \frac{2}{3}\right\} \cdot [0.2 \vee_p 0.3] + \left\{\frac{2}{3}\right\} \cdot [0.2 \wedge_p 0.3]\right), \frac{1}{2}$$
$$\cdot [0.1 \wedge_F 0.4 + 0.1 \vee_F 0.4], \left\{1 - \frac{2}{3}\right\} \cdot [0.4 \wedge_F 0.2]$$
$$+ \left\{\frac{2}{3}\right\} \cdot [0.4 \vee_F 0.2]$$
$$= \left(\frac{1}{3} \cdot [0.2 + 0.3 - 0.2 \cdot 0.3] + \frac{2}{3} \cdot [0.2 \cdot 0.3], \frac{1}{2}\right.$$
$$\cdot [0.1 \cdot 0.4 + 0.4 + 0.1 - 0.1 \cdot 0.4], \frac{1}{3} \cdot [0.4 \cdot 0.2]$$
$$\left. + \frac{2}{3} \cdot [0.4 + 0.2 - 0.4 \cdot 0.2]\right)$$
$$\approx (0.19, 0.25, 0.37).$$

{The degree of contradiction is $\frac{2}{3} > 0.5$.}

$d_A^N(x, tall) \vee_p d_B^N(x, tall)$

$$= \begin{pmatrix} & 0 & \\ 0.8, & 0.3, & 0.1 \end{pmatrix} \vee_p \begin{pmatrix} & 0 & \\ 0.7, & 0.1, & 0.6 \end{pmatrix}$$

$$= (0.8 \vee_p 0.7, 0.3 \wedge_p 0.1, 0.1 \wedge_p 0.6)$$

$$= \left(0.8 \vee_F 0.7, \frac{1}{2}(0.3 \wedge_F 0.1 + 0.3 \vee_F 0.1), 0.1 \wedge_F 0.6\right)$$

$$= \left(0.8 + 0.7 - 0.8 \cdot 0.7, \frac{1}{2}(0.3 \cdot 0.1 + 0.3 + 0.1 - 0.3 \cdot 0.1), 0.1 \cdot 0.6\right) = (0.94, 0.20, 0.06).$$

$d_A^N(x, medium) \vee_p d_B^N(x, medium)$

$$= \begin{pmatrix} & \frac{1}{2} & \\ 0.6, & 0.2, & 0.3 \end{pmatrix} \vee_p \begin{pmatrix} & \frac{1}{2} & \\ 0.5, & 0.1, & 0.3 \end{pmatrix}$$

$$= (0.6 \vee_p 0.5, 0.2 \vee_p 0.1, 0.3 \vee_p 0.3)$$

$$= \left(\left\{1 - \frac{1}{2}\right\} \cdot [0.6 \vee_p 0.5] + \left\{\frac{1}{2}\right\} \cdot [0.6 \wedge_F 0.5]\right), \frac{1}{2} \cdot [0.2 \wedge_F 0.1 + 0.2 \vee_p 0.1],$$

$$\left\{1 - \frac{1}{2}\right\} \cdot [0.3 \wedge_F 0.3] + \left\{\frac{1}{2}\right\} \cdot [0.3 \vee_p 0.3]$$

$$= \left(\frac{1}{2} \cdot [0.6 + 0.5 - 0.6 \cdot 0.5] + \frac{1}{2} \cdot [0.6 \cdot 0.5], \frac{1}{2} \cdot [0.2 \cdot 0.1 + 0.2 + 0.1 - 0.2 \cdot 0.1], \frac{1}{2} \cdot [0.3 \cdot 0.3] + \frac{1}{2} \cdot [0.3 + 0.3 - 0.3 \cdot 0.3]\right)$$

$$= (0.55, 0.15, 0.30).$$

# III. PLITHOGENIC LOGIC

We now trace the previous study on plithogenic set and adjust it to the plithogenic logic. In order for the chapter to be self-contained, we recopy the previous main plithogenic set formulas and ideas, but we correlate and adapt them to the logic field.

## III.1. Informal Definition of Plithogenic Logic

A *plithogenic logical proposition P* is a proposition that is characterized by <u>many degrees of truth-values</u> with respect to the corresponding attributes' values that characterize *P*.

For each attribute's value *v* there is a corresponding *degree of truth-value d(P, v)* of *P* with respect to the attribute value *v*.

In order to obtain a better accuracy for the plithogenic aggregation logical operators, a *contradiction (dissimilarity) degree* is defined between each attribute value and the dominant (most important) attribute value.

{However, there are cases when such dominant attribute value may not be taking into consideration or may not exist [and then by default the contradiction degree is taken as zero], or there may be many dominant attribute values. In such cases, either the contradiction degree function is suppressed, or another relationship function between attribute values should be established.}

The plithogenic aggregation logical operators (conjunction, disjunction, negation, inclusion, equality) are based on contradiction degrees between attributes' values, and the first two are linear combinations of the fuzzy logical operators $t_{norm}$ and $t_{conorm}$.

Plithogenic logic is a generalization of the classical logic, fuzzy logic, intuitionistic fuzzy logic, and neutrosophic logic, since these four types of logics are characterized by a single attribute value (*truth-value*): which has one value (*truth*) – for the classical logic and fuzzy logic, two values (*truth,* and *falsehood*) – for intuitionistic fuzzy logic, or three values (*truth, falsehood,* and *indeterminacy*) – for neutrosophic logic.

A plithogenic logic proposition *P*, in general, may be characterized by four or more degrees of truth-values resulted from the number of attribute-values that characterize *P*. The number of attribute-values is established by the experts.

## III.2. Formal Definition of Single (Uni-Dimensional) Attribute Plithogenic Set

Let *U* be a logical universe of discourse, and *P* a logical proposition, $P \in U$.

### III.2.1. Attribute Value Spectrum

Let $\mathcal{A}$ be a non-empty set of uni-dimensional attributes $\mathcal{A} = \{\alpha_1, \alpha_2, \ldots, \alpha_m\}$,

$m \geq 1$; and $\alpha \in \mathcal{A}$ be a given attribute whose spectrum of all possible values (or states) is the non-empty set *S*, where *S* can be a finite discrete set, $S = \{s_1, s_2, \ldots, s_l\}$, $1 \leq l < \infty$, or infinitely countable set $S = \{s_1, s_2, \ldots, s_\infty\}$, or infinitely uncountable (continuum) set $S = ]a, b[$, $a < b$, where $]\ldots[$ is any open, semi-open, or closed interval from the set of real numbers or from other general set.

### III.2.2. Attribute Value Range

Let $V$ be a non-empty subset of $S$, where $V$ is the *range of all attribute's values* needed by the experts for their logical application. Each logical proposition $P$ is characterized by all attribute's values in $V = \{v_1, v_2, …, v_n\}$, for $n \geq 1$.

### III.2.3. Dominant Attribute Value

Into the attribute's value set $V$, in general, there is a <u>dominant attribute value</u>, which is determined by the experts upon their application. Dominant attribute value means the most important attribute value that the experts are interested in.

{However, there cases when such dominant attribute value may not be taking into consideration or not exist [in such case it is zero by default], or there may be many dominant (important) attribute values - when different approach should be employed.}

### III.2.4. Attribute Value Truth-Value Degree Function

With respect to each attributes value $v \in V$ the proposition $P$ has a corresponding degree of truth-value: $d(P, v)$.

The degree of truth-value may be: a fuzzy degree of truth-value, or intuitionistic fuzzy degree of truth-value, or neutrosophic degree of truth-value of the proposition $P$ with respect to $v$.

Therefore, the <u>attribute value truth-value degree function</u> is:
$$\forall P \in U, d: U \times V \to \mathcal{P}([0, 1]^z), \tag{197}$$
so $d(P, v)$ is a subset of $[0, 1]^z$, where

$\mathcal{P}([0, 1]^z)$ is the power set of the $[0, 1]^z$, where $z = 1$ (for fuzzy degree of truth-value), $z = 2$ (for intuitionistic fuzzy degree of truth-value), or $z = 3$ (for neutrosophic degree de truth-value).

## III.2.5. Attribute Value Contradiction Degree Function

Let the cardinal $|V| \geq 1$.

Let $c: V \times V \rightarrow [0, 1]$ be the <u>attribute value contradiction degree function</u> (that we introduce now for the first time) between any two attribute values $v_1$ and $v_2$, denoted by

$c(v_1, v_2)$, and satisfying the following axioms:

$c(v_1, v_1) = 0$, the contradiction degree between the same attribute values is zero;

$c(v_1, v_2) = c(v_2, v_1)$, commutativity.

For simplicity, we use a <u>fuzzy attribute value contradiction degree function</u> ($c$ as above, that we may denote by $c_F$ in order to distinguish it from the next two), but an <u>intuitionistic attribute value contradiction function</u> ($c_{IF} : V \times V \rightarrow [0, 1]^2$), or more general a <u>neutrosophic attribute value contradiction function</u> ($c_N : V \times V \rightarrow [0, 1]^3$) may be utilized increasing the complexity of calculation but the accuracy as well.

We mostly compute the contradiction degree between *uni-dimensional attribute values*. For *multi-dimensional attribute values* we split them into corresponding uni-dimensional attribute values.

The attribute value contradiction degree function helps the plithogenic aggregation logical operators, and the plithogenic logical inclusion (partial order) relationship to obtain a more accurate result.

The attribute value contradiction degree function is designed in each field where plithogenic logic is used in accordance with the application to solve. If it is ignored, the aggregations still work, but the result may lose accuracy.

Then $(U, a, V, d, c)$ is called a plithogenic logic:

where "$U$" is a logical universe of discourse of many logical generic propositions $P$, "$a$" is a (multi-dimensional in general) attribute, "$V$" is the range of the attribute's values, "$d$" is the degree of truth-value of each logical proposition $P \in U$ with respect to each attribute value - and "$d$" stands for "$d_F$" or "$d_{IF}$" or "$d_N$", when dealing with fuzzy degree of truth-value, intuitionistic fuzzy degree of truth-value, or neutrosophic degree of truth-value respectively of a plithogenic logical proposition $P$; and "$c$" stands for "$c_F$" or "$c_{IF}$" or "$c_N$", when dealing with fuzzy degree of contradiction (dissimilarity), intuitionistic fuzzy degree of contradiction, or neutrosophic degree of contradiction between attribute values respectively.

The functions $d(\cdot,\cdot)$ and $c(\cdot,\cdot)$ are defined in accordance with the logical applications the experts need to solve.

One uses the notation:

$P(d(P,V))$,

where $d(P,V) = \{d(P,v), \text{for all } v \in V\}, \forall P \in U$.

## III.2.6. About the Plithogenic Aggregation Logical Operators

The attribute value contradiction degree is calculated between each attribute value with respect to the dominant attribute value (denoted $v_D$) in special, and with respect to other attribute values as well.

The attribute value contradiction degree function $c$ between the attribute's values is used into the definition of <u>plithogenic aggregation logical operators</u> {*conjunction (AND), disjunction (OR), Negation ($\neg$), Implication ($\Rightarrow$), Equivalence ($\Leftrightarrow$), and other plithogenic aggregation operators that combine two or more attribute value degrees - that $t_{norm}$ and $t_{conorm}$ act upon*}.

Several of the plithogenic aggregation logical operators are linear combinations of the fuzzy $t_{norm}$ *(denoted $\wedge_F$)*, and fuzzy $t_{conorm}$ *(denoted $\vee_F$)*, but non-linear combinations may as well be constructed.

If one applies the $t_{norm}$ on dominant attribute value denoted by $v_D$, and the contradiction between $v_D$ and $v_2$ is $c(v_D, v_2)$, then onto attribute value $v_2$ one applies:

*[1 − c(v_D, v_2)]·t_{norm}(v_D, v_2) + c(v_D, v_2)·t_{conorm}(v_D, v_2)*, (198)

Or, by using symbols:

*[1 − c(v_D, v_2)]·(v_D$\wedge_F$v_2) + c(v_D, v_2)·(v_D$\vee_F$v_2).* (199)

Similarly, if one applies the $t_{conorm}$ on dominant attribute value denoted by $v_D$, and the contradiction between $v_D$ and $v_2$ is $c(v_D, v_2)$, then onto attribute value $v_2$ one applies:

*[1 − c(v_D, v_2)]·t_{conorm}(v_D, v_2) + c(v_D, v_2)·t_{norm}(v_D, v_2)*, (200)

Or, by using symbols:

*[1 − c(v_D, v_2)]·(v_D$\vee_F$v_2) + c(v_D, v_2)·(v_D$\wedge_F$v_2).* (201)

## III.3. Plithogenic Logic as Generalization of other Logics

The plithogenic set is an extension of all: classical logic, fuzzy logic, intuitionistic fuzzy logic, and neutrosophic logic.

For examples:

Let *U* be a logical universe of discourse, and a generic logical proposition P ∈ U. Then:

## III.3.1. Classical (Crisp) Logic (CCL)

The *attribute* is α = "truth-value";

the *set of attribute values V = {truth, falsehood},* with cardinal |V| = 2;

the *dominant attribute value = truth*;

the *attribute value truth-value degree function*:

$$d: U \times V \rightarrow \{0, 1\}, \qquad (202)$$

$d(P, truth) = 1, \ d(P, falsehood) = 0,$

and the *attribute value contradiction degree function*:

$$c: V \times V \rightarrow \{0, 1\}, \qquad (203)$$

$c(truth, truth) = c(falsehood, falsehood) = 0,$

$c(truth, falsehood) = 1.$

### III.3.1.2. Crisp (Classical) Intersection

$$a \wedge b \in \{0, 1\} \qquad (204)$$

### III.3.1.3. Crisp (Classical) Union

$$a \vee b \in \{0, 1\} \qquad (205)$$

### III.3.1.4. Crisp (Classical) Complement (Negation)

$$\neg a \in \{0, 1\}. \qquad (206)$$

## III.3.2. Single-Valued Fuzzy Logic (SVFL)

The *attribute* is α = "truth-value";

*set of attribute values V = {truth},* whose cardinal |V| = 1;

the *dominant attribute value = truth*;

the *attribute value truth-value degree function:*

$$d: U \times V \rightarrow [0, 1], \qquad (207)$$

with $d(P, truth) \in [0, 1]$;

and the *attribute value contradiction degree function*:

$$c: V \times V \rightarrow [0, 1], \tag{208}$$

$c(truth, truth) = 0.$

### III.3.2.1. Fuzzy Intersection

$a \wedge_F b \in [0, 1]$

### III.3.2.2. Fuzzy Union

$a \vee_F b \in [0, 1]$

### III.3.2.3. Fuzzy Complement (Negation)

$$\neg_F a = 1 - a \in [0, 1]. \tag{209}$$

## III.3.3. Single-Valued Intuitionistic Fuzzy Logic (SVIFL)

The *attribute* is α = "truth-value";

the *set of attribute values* $V = \{truth, falsehood\}$, whose cardinal $|V| = 2$;

the *dominant attribute value* = *truth*;

the *attribute-value truth-value degree function*:

$$d: U \times V \rightarrow [0, 1], \tag{210}$$

$d(P, truth) \in [0, 1]$, $d(P, falsehood) \in [0, 1]$,

with $d(P, truth) + d(P, falsehood) \leq 1$,

and the *attribute value contradiction degree function*:

$$c: V \times V \rightarrow [0, 1], \tag{211}$$

$c(truth, truth) = c(falsehood, falsehood) = 0$,

$c(truth, falsehood) = 1$,

which means that for SVIFL aggregation operators' conjunction (AND) and disjunction (OR), if one applies the $t_{norm}$ on truth degree, then one has to apply the $t_{conorm}$ on falsehood degree – and reciprocally.

Therefore:

### III.3.3.1. Intuitionistic Fuzzy Intersection

$$(a_1, a_2) \wedge_{IFS} (b_1, b_2) = (a_1 \wedge_F b_1, a_2 \vee_F b_2) \qquad (212)$$

### III.3.3.2. Intuitionistic Fuzzy Union

$$(a_1, a_2) \vee_{IFS} (b_1, b_2) = (a_1 \vee_F b_1, a_2 \wedge_F b_2), \qquad (213)$$

and

### III.3.3.3. Intuitionistic Fuzzy Complement (Negation)

$$\neg_{IFS}(a_1, a_2) = (a_2, a_1). \qquad (214)$$

where $\wedge_F$ and $\vee_F$ are the fuzzy $t_{norm}$ and fuzzy $t_{conorm}$ respectively.

## III.3.4. Single-Valued Neutrosophic Set (SVNS)

The *attribute* is α = "truth-value";

the *set of attribute values* V = *{truth, indeterminacy, falsehood}*, whose cardinal |V| = 3;

the *dominant attribute value = truth*;

the *attribute-value truth-value degree function*:

$$d: U \times V \rightarrow [0, 1], \qquad (215)$$

$d(P, truth) \in [0, 1]$, $d(P, indeterminacy) \in [0, 1]$, $d(P, falsehood) \in [0, 1]$,

with

$0 \leq d(P, truth) + d(P, indeterminacy) + d(P, falsehood) \leq 3$;

and the *attribute-value contradiction degree function*:

$$c: V \times V \rightarrow [0, 1], \qquad (216)$$

$c(truth, truth) = c(indeterminacy, indeterminacy) = c(falsehood, falsehood) = 0$, $c(truth, falsehood) = 1$,

$c(truth, indeterminacy) = c(falsehood, indeterminacy) = 0.5$,

which means that for the SVNL aggregation operators (conjunction, disjunction, negation etc.), if one applies the $t_{norm}$ on truth, then one has to apply the $t_{conorm}$ on falsehood {and reciprocally), while on indeterminacy one applies the average of $t_{norm}$ and $t_{conorm}$, as follows:

### III.3.4.1. Neutrosophic Conjunction

*III.3.4.1.1. Simple Neutrosophic conjunction (the most used by the neutrosophic community):*

$$(a_1, a_2, a_3) \wedge_{NL} (b_1, b_2, b_3) = \left( a_1 \wedge_F b_1, a_2 \vee_F b_2, a_3 \vee_F b_3 \right)$$
(217)

*III.3.4.1.2. Plithogenic Neutrosophic Conjunction*

$$(a_1, a_2, a_3) \wedge_P (b_1, b_2, b_3)$$
$$= \left( a_1 \wedge_F b_1, \frac{1}{2}\left[ (a_2 \wedge_F b_2) + (a_2 \vee_F b_2) \right], a_3 \vee_F b_3 \right), \quad (218)$$

### III.3.4.2. Neutrosophic Disjunction

*III.3.4.2.1. Simple Neutrosophic Disjunction (the most used by the neutrosophic community):*

$$(a_1, a_2, a_3) \vee_{NL} (b_1, b_2, b_3) = \left( a_1 \vee_F b_1, a_2 \wedge_F b_2, a_3 \wedge_F b_3 \right)$$
(219)

*III.3.4.2.2. Plithogenic Neutrosophic Disjunction*

$$(a_1, a_2, a_3) \vee_P (b_1, b_2, b_3) =$$
$$\left( a_1 \vee_F b_1, \frac{1}{2}\left[ (a_2 \wedge_F b_2) + (a_2 \vee_F b_2) \right], a_3 \wedge_F b_3 \right)$$
(220)

In other way, with respect to what one applies on the truth, one applies the opposite on falsehood, while on indeterminacy one applies the average between them.

### III.3.4.3. Neutrosophic Negation:

$$\neg_{NS} (a_1, a_2, a_3) = (a_3, a_2, a_1).$$
(221)

## III.4. One-Attribute-Value Plithogenic Single-Valued Fuzzy Logic Operators

We consider the single-value number degrees, for simplicity of the example.

Let $v_i$ be a uni-attribute value and its degree of contradiction with respect to the dominant uni-attribute value $v_D$ be

$c(v_D, v_i) \stackrel{\text{def}}{=} c_i$.

Let $d_A(P, v_i)$ be the truth-value degree of the attribute-value $v_i$ of the logical proposition $P$ with respect to the Expert A. And similarly for $d_B(P, v_i)$. Then, we recall the plithogenic aggregation logical operators with respect to this attribute value $v_i$ that will be employed:

### III.4.1. One-Attribute-Value Plithogenic Single-Valued Fuzzy Logic Conjunction

$$d_A(P, v_i) \wedge_p d_B(P, v_i)$$
$$= (1 - c_i) \cdot [d_A(P, v_i) \wedge_F d_B(P, v_i)] + c_i$$
$$\cdot [d_A(P, v_i) \vee_F d_B(P, v_i)]$$

(222)

### III.4.2. One-Attribute Value Plithogenic Single-Valued Fuzzy Logic Disjunction

$$d_A(P, v_i) \vee_p d_B(P, v_i) =$$
$$= (1 - c_i) \cdot [d_A(P, v_i) \vee_F d_B(P, v_i)] + c_i$$
$$\cdot [d_A(P, v_i) \wedge_F d_B(P, v_i)]$$

(223)

## III.4.3. One Attribute Value Plithogenic Single-Valued Fuzzy Logic Negation

$$\neg_p v_i = anti(v_i) = (1 - c_i) \cdot v_i \qquad (224)$$
$$\neg_p d_A(P, (1 - c_i)v_i) = d_A(P, v_i) \qquad (225)$$

## III.5. n-Attribute-Values Plithogenic Single-Valued Logic Operators

The easiest way to apply the plithogenic logic operators on a multi-attribute plithogenic logic is to split back the *m*-dimensional attribute, $m \geq 1$, into *m* uni-dimensional attributes, and separately apply the plithogenic logic operators on the set of all values (needed by the application to solve) of each given attribute.

Therefore, let α be a given attribute, characterizing each element $x \in P$, whose set of values are:

$$V = \{v_1, v_2, \dots, v_n\} \equiv \{v_D, v_2, \dots, v_n\}, n \geq 1, \qquad (226)$$

where $v_D$ = dominant attribute value, and $c(v_D, v_i) = c_i \in [0, 1]$ the contradiction degrees. Without restricting the generality, we consider the values arranged in an increasing order with respect to their contradiction degrees, i.e.:

$$c(v_D, v_D) = 0 \leq c_1 \leq c_2 \leq \cdots \leq c_{i_0}$$
$$< \frac{1}{2} \leq c_{i_0+1} \leq \cdots \leq c_n \leq 1. \qquad (227)$$

## III.5.1. n-Attribute-Values Plithogenic Single-Valued Fuzzy Logic Operators

Let's consider two experts, A and B, which evaluate a logical proposition *P*, with respect to the fuzzy degrees of the values

$v_1, \ldots, v_n$ of truth-values of the logical proposition $P$, upon some given criteria:

$$d_A^F: P \times V \to [0, 1], \quad d_A^F(x, v_i) = a_i \in [0, 1], \qquad (228)$$
$$d_B^F: P \times V \to [0, 1], \quad d_B^F(x, v_i) = b_i \in [0, 1], \qquad (229)$$

for all $i \in \{1, 2, \ldots, n\}$.

### III.5.2. n-Attribute-Values Plithogenic Single-Valued Fuzzy Logic Conjunction

$$(a_1, a_2, \ldots, a_{i_0}, a_{i_0+1}, \ldots, a_n) \wedge_p (b_1, b_2, \ldots, b_{i_0}, b_{i_0+1}, \ldots, b_n)$$
$$= (a_1 \wedge_p b_1, a_2 \wedge_p b_2, \ldots, a_{i_0} \wedge_p b_{i_0}, a_{i_0+1} \wedge_p b_{i_0+1}, \ldots, a_n \wedge_p b_n) \qquad (230)$$

The first $i_0$ conjunctions are proper plithogenic conjunctions (the weights onto the $t_{norm}$'s are bigger than onto $t_{conorm}$'s):

$$a_1 \wedge_p b_1, a_2 \wedge_p b_2, \ldots, a_{i_0} \wedge_p b_{i_0} \qquad (231)$$

whereas the next $n - i_0$ conjunctions

$$a_{i_0+1} \wedge_p b_{i_0+1}, \ldots, a_n \wedge_p b_n \qquad (232)$$

are improper plithogenic disjunctions (since the weights onto the $t_{norm}$'s are less than onto $t_{conorm}$'s):

### III.5.3. n-Attribute-Values Plithogenic Single-Valued Fuzzy Logic Disjunction

$$(a_1, a_2, \ldots, a_{i_0}, a_{i_0+1}, \ldots, a_n) \vee_p (b_1, b_2, \ldots, b_{i_0}, b_{i_0+1}, \ldots, b_n) \qquad (233)$$
$$= (a_1 \vee_p b_1, a_2 \vee_p b_2, \ldots, a_{i_0} \vee_p b_{i_0}, a_{i_0+1} \vee_p b_{i_0+1}, \ldots, a_n \vee_p b_n)$$

The first $i_0$ disjunctions are proper plithogenic disjunctions (the weights onto the $t_{conorm}$'s are bigger than onto $t_{norm}$'s):

$$a_1 \vee_p b_1, a_2 \vee_p b_2, \ldots, a_{i_0} \vee_p b_{i_0} \qquad (234)$$

whereas the next $n - i_0$ disjunctions

$$a_{i_0+1} \vee_p b_{i_0+1}, \ldots, a_n \vee_p b_n \qquad (235)$$

are improper plithogenic conjunctions (since the weights onto the *t*<sub>conorm</sub>'s are less than onto *t*<sub>norm</sub>'s):

## III.5.4. n-Attribute-Values Plithogenic Single-Valued Fuzzy Logic Negations

In general, for a generic logical proposition $P$, characterized by the uni-dimensional attribute α, whose attribute values are $V = (v_D, v_2, ..., v_n), n \geq 2$, and with attribute value contradiction degrees (with respect to the dominant attribute value $v_D$) are respectively: $0 \leq c_2 \leq \cdots \leq c_{n-1} \leq c_n \leq 1$, and their attribute value degrees of truth-values with respect to the set $P$ are respectively $a_D, a_2, ..., a_{n-1}, a_n \in [0, 1]$, then the plithogenic fuzzy logic negation of $P$ is:

$$\neg_p[\,P\begin{pmatrix} 0 & c_2 & & c_{n-1} & c_n \\ v_D, & v_2, & ..., & v_{n-1}, & v_n \\ a_D & a_2 & & a_{n-1} & a_n \end{pmatrix}] =$$

$$\neg_p P\begin{pmatrix} 1-c_n & 1-c_{n-1} & & 1-c_2 & 1-c_D \\ anti(v_n) & anti(v_{n-1}) & ... & anti(v_2) & anti(v_D) \\ a_n & a_{n-1} & & a_2 & a_D \end{pmatrix}. \quad (236)$$

Some $anti(V_i)$ may coincide with some $V_j$, whereas other $anti(V_i)$ may fall in between two consecutive $[v_k, v_{k+1}]$ or we may say that they belong to the *Refined set* V;

or

$$= \begin{Bmatrix} v_n & v_{n-1} & ... & ... & v_1 \\ a_1, & a_2, & ..., & ..., & a_n \end{Bmatrix} \quad (237)$$

{This version gives an exact result when the contradiction degrees are equi-distant (for example: *0, 0.25, 0.50, 0.75, 1*) or symmetric with respect to the center *0.5* (for example: *0, 0.4, 0.6, 1*), and an approximate result when they are not equi-distant and not symmetric to the center (for example: *0, 0.3, 0.8, 0.9, 1*);}

or

$$\begin{Bmatrix} v_1 & v_2 & \dots & v_{i_0} & v_{i_0+1} & \dots & v_n \\ 1-a_1 & 1-a_2 & \dots & 1-a_{i_0} & 1-a_{i_0+1} & \dots & 1-a_n \end{Bmatrix} \quad (238)$$

where $anti(v_i) \in V$ or $anti(v_i) \in RefinedV$, for all $i \in \{1, 2, \dots, n\}$.

## III.6. Multi-Attribute Plithogenic General Logic

### III.6.1. Definition of Multi-Attribute Plithogenic General Logic

Let $U$ be a logical universe of discourse, and a plithogenic logical proposition $P \in U$.

Let $\mathcal{H}$ be a set of $m \geq 2$ attributes: $\alpha_1, \alpha_2, \dots, \alpha_m$, whose corresponding spectra of values are the non-empty sets $S_1, S_2, \dots, S_m$ respectively.

Let $V_1 \subseteq S_1, V_2 \subseteq S_2, \dots, V_m \subseteq S_m$ be subsets of attribute values of the attributes $\alpha_1, \alpha_2, \dots, \alpha_m$ respectively needed by experts in their given application.

For each $j \in \{1, 2, \dots, m\}$, the *set of attribute values $V_j$* means the *range of attribute $\alpha_j$'s values*, needed by the experts in a specific application or in a specific problem to solve.

Each logical proposition $P \in U$ is characterized by all $m$ attributes.

Let the m-dimensional attribute value degree function be:
$$d_{[m]}: (U, V_1 \times V_2 \times \dots \times V_m) \to \mathcal{P}([0,1])^m. \quad (239)$$

For any $P \in U$, and any $v_j \in V_j$ with $j \in \{1, 2, \dots, m\}$, one has:
$$d_{[m]}(P(v_1, v_2, \dots, v_m)) \subseteq \mathcal{P}([0,1])^m. \quad (240)$$

## III.6.2. Example of Plithogenic Logic

Let

*P = "John is a knowledgeable person"*

be a logical proposition.

The three attributes under which this proposition has to be evaluated about - according to the experts - are:

<u>Science</u> (whose attribute values are: *mathematics, physics, anatomy*), <u>Literature</u> (whose attribute values are: *poetry, novel*), and <u>Arts</u> (whose only attribute value is: *sculpture*).

Assume that one has: fuzzy truth-value degrees, and fuzzy contradiction (dissimilarity) degrees – for simpler calculation.

The experts consider that the attributes' values contradiction (dissimilarity) degrees, determined by the experts that study this problem, are:

    0       0.3    0.8    0    0.9    0

*mathematics, physics, anatomy; poetry, novels; sculpture*

Let's assume that mathematics, poetry, sculpture are <u>dominant [most important] attribute values</u> for the attributes Science, Literature, and Arts respectively.

The degree of contradiction (dissimilarity) between *physics* and *mathematics* is 0.3, between *anatomy* and *mathematics* is 0.8;

while the degree of contradiction (dissimilarity) between *novels* and *poetry* is 0.9;

there is no degree of contradiction (dissimilarity) between attribute values from different attribute classes (for example between *sculpture* and *anatomy*, etc.).

According to Expert A(lexander), the truth-values of plithogenic proposition *P* are:

$P_A$(0.7, 0.6, 0.4; 0.9, 0.2; 0.5),

which means that John's degree of truth (knowledge) in mathematics is *0.7*, degree of truth (knowledge) in physics is *0.6*, degree of truth (knowledge) in anatomy is *0.4*; degree of truth (knowledge) in poetry is *0.9*, degree of truth (knowledge) in novels is *0.2*; degree of truth (knowledge) in sculpture is *0.5*.

But, according to Expert B(arbara), the truth-values of plithogenic proposition *P* are:

$P_B(0.9, 0.6, 0.2;\ 0.8, 0.7;\ 0.3)$.

We use the *6*-attribute-values plithogenic single-valued logical intersection, taken as before

$$t_{norm}(a, b) = a \wedge_F b = a \cdot b \qquad (241)$$

and $t_{conorm}(a, b) = a \vee_F b = a + b - a \cdot b$

$$(242)$$

whence we get:

$P_A(0.7, 0.6, 0.4;\ 0.9, 0.2;\ 0.5) \wedge_P P_B(0.9, 0.6, 0.2;\ 0.8, 0.7;\ 0.3)$
$= P_{A \wedge_P B}(0.7 \wedge_P 0.9, 0.6 \wedge_P 0.6, 0.4 \wedge_P 0.2;\ 0.9 \wedge_P 0.8, 0.2 \wedge_P 0.7;\ 0.5 \wedge_P 0.3) = P_{A \wedge_P B}(0.7 \wedge_F 0.9, 0.6 \wedge_P 0.6, 0.4 \wedge_P 0.2;\ 0.9 \wedge_F 0.8, 0.2 \wedge_P 0.7; 0.5 \wedge_F 0.3) = P_{A \wedge_P B}(0.7 \cdot 0.9, (1-0.3) \cdot 0.6 \wedge_F 0.6 + 0.3 \cdot 0.6 \vee_F 0.6, (1-0.8) \cdot 0.4 \wedge_F 0.2 + 0.8 \cdot 0.4 \vee_F 0.2;\ 0.9 \cdot 0.8, (1-0.9) \cdot 0.2 \wedge_F 0.7 + 0.9 \cdot 0.2 \vee_F 0.7;\ 0.5 \cdot 0.3) =$

$P_{A \wedge_P B}(0.630, 0.504, 0.432;\ 0.720, 0.698;\ 0.150)$.

# IV. PLITHOGENIC PROBABILITY

We again trace both previous studies on plithogenic set and plithogenic logic respectively, and adjust them to the plithogenic probability. In order for the chapter to be self-contained, we recopy as well the previous plithogenic set and plithogenic logic main formulas and ideas, but we correlate and adapt them to the probabilistic field.

## IV.1. Informal Definition of Plithogenic Probability

In the *plithogenic probability* each event $E$ from a probability space $U$ is characterized by <u>many chances</u> of the event to occur [<u>not only one chance</u> of the event to occur: as in classical probability, imprecise probability, and neutrosophic probability], chances of occurrence calculated with respect to the corresponding attributes' values that characterize the event $E$. The attributes' values that characterize the event are established by experts with respect to the application or problem they need to solve.

A *discrete finite n-attribute-values plithogenic probability space $U_{nk}$*, of $k$ events, each event together with its $n$ chances of occurring, is displayed below:

$U_{nk} = \{E_1(d_{11}, d_{21}, ..., d_{n1}), E_2(d_{12}, d_{22}, ..., d_{n2}), ...,$
$E_k(d_{1k}, d_{2k}, ..., d_{nk})\}.$  (243)

With respect to each attribute's value $v_j$, $j \in \{1, 2, ..., n\}$, $n \geq 1$, there is a corresponding degree of chance $d(E_i, v_j) = d_{ij}$ of the event $E_i$ to occur, for $i \in \{1, 2, ..., k\}$, $k \geq 1$.

A *discrete infinite n-attribute-values plithogenic probability space* $U_{n\infty}$, of infinitely many events, each event together with its $n$ chances of occurring, is displayed below:

$U_{n\infty} = \{E_1(d_{11}, d_{21}, ..., d_{n1}), E_2(d_{12}, d_{22}, ..., d_{n2}), ...,$
$E_\infty(d_{1\infty}, d_{2\infty}, ..., d_{n\infty})\}.$ (244)

While a continuous *n-attribute-values* plithogenic probability space $U_{nI}$ is:

$U_{nI} = \{E_i(d_{1i}, d_{2i}, ..., d_{ni}), i \in I,$ (245)

where $I$ is a continuous set of indices}.

And a *continuous infinite-attribute-values plithogenic probability space* $U_{I_1 I_2 \infty}$ is:

$U_{I_1 I_2 \infty} = \{E_i(d_{1i}, d_{2i}, ..., d_{ji}), i \in I_1, j \in I_2,$ (246)

where $I_1$ and $I_2$ are continuous sets of indices}.

In order to obtain a better accuracy for the plithogenic aggregation probabilistic operators, a *contradiction (dissimilarity) degree* is defined between each attribute value and the dominant (most important) attribute value.

{However, there are cases when such dominant attribute value may not be taking into consideration or may not exist [and then by default the contradiction degree is taken as zero], or there may be many dominant attribute values. In such cases, either the contradiction degree function is suppressed, or another relationship function between attribute values should be established.}

The plithogenic aggregation probabilistic operators (conjunction, disjunction, negation, inclusion, equality) are based on contradiction degrees between attributes' values, and the first two are linear combinations of the fuzzy logical operators' $t_{norm}$ and $t_{conorm}$.

## IV.2. Plithogenic Probability as Generalization of other Probabilities

Plithogenic probability is a generalization of the <u>classical probability</u> [ since a single event may have more crisp-probabilities of occurrence ], <u>imprecise probability</u> [ since a single event may have more subunitary subset-probabilities of occurrence ], and <u>neutrosophic probability</u> [ since a single event may have more triplets of:  subunitary subset-probabilities of occurrence, subunitary subset-probabilities of indeterminacy (not clear if occurring or not occurring), and  subunitary subset-probabilities of nonoccurring ].

## IV.3. Formal Definition of Single (Uni-Dimensional) Attribute Plithogenic Probability

Let $U$ be a probability space, and an event $E \in U$.

### IV.3.1. Attribute Value Spectrum

Let $\mathcal{A}$ be a non-empty set of uni-dimensional attributes

$\mathcal{A} = \{\alpha_1, \alpha_2, ..., \alpha_m\}$, $m \geq 1$; and $\alpha \in \mathcal{A}$ be a given attribute whose spectrum of all possible values (or states) is the non-empty set $S$, where $S$ can be a finite discrete set, $S = \{s_1, s_2, ..., s_l\}$, $1 \leq l < \infty$, or infinitely countable set $S = \{s_1, s_2, ..., s_\infty\}$, or infinitely uncountable (continuum) set $S = ]a, b[$, $a < b$, where $]...[$ is any open, semi-open, or closed interval from the set of real numbers or from other general set.

## IV.3.2. Attribute Value Range

Let $V$ be a non-empty subset of $S$, where $V$ is the *range of all attribute's values* needed by the experts for their probabilistic application. Each probabilistic event $E$ is characterized by all attribute's values in $V = \{v_1, v_2, …, v_n\}$, for $n \geq 1$.

## IV.3.3. Dominant Attribute Value

Into the attribute's value set $V$, in general, there is a dominant attribute value, which is determined by the experts upon their application. Dominant attribute value means the most important attribute value that the experts are interested in.

{However, there cases when such dominant attribute value may not be taking into consideration or not exist [in such case it is zero by default], or there may be many dominant (important) attribute values - when different approach should be employed.}

## IV.3.4. Attribute-Value Chance-of-Occurrence Degree-Function

With respect to each attributes value $v \in V$ the event $E$ has a corresponding degree of occurring: $d(E, v)$.

The degree of occuring may be: a *fuzzy degree of occuring*, or *intuitionistic fuzzy degree of occurring-nonoccurring*, or *neutrosophic degree of occurring- indeterminacy-nonoccuring*.

Therefore, the attribute-value chance-of-occurrence degree-function is:

$\forall E \in U, d: U \times V \to \mathcal{P}([0, 1]^z)$,

(247)

so $d(P, v)$ is a subset of $[0, 1]^z$, where

$\mathcal{P}([0, 1]^z)$ is the power set of $[0, 1]^z$, where $z = 1$ (for fuzzy degree of occurrence), $z = 2$ (for intuitionistic fuzzy degree of occurrence-nonoccurrence), or $z = 3$ (for neutrosophic degree de occurrence-indeterminacy-nonoccurence).

## IV.3.5. Attribute-Value Contradiction (Dissimilarity) Degree Function

Let the cardinal $|V| \geq 1$.

Let $c: V \times V \to [0, 1]$ be the <u>attribute value contradiction degree function</u> (that we introduce now for the first time) between any two attribute values $v_1$ and $v_2$, denoted by

$c(v_1, v_2)$, and satisfying the following axioms:

$c(v_1, v_1) = 0$, the contradiction degree between the same attribute values is zero;

$c(v_1, v_2) = c(v_2, v_1)$, commutativity.

For simplicity, we use a <u>fuzzy attribute value contradiction degree function</u> ($c$ as above, that we may denote by $c_F$ in order to distinguish it from the next two), but an <u>intuitionistic attribute value contradiction function</u> ($c_{IF} : V \times V \to [0, 1]^2$), or more general a <u>neutrosophic attribute value contradiction function</u> ($c_N : V \times V \to [0, 1]^3$) may be utilized increasing the complexity of calculation but the accuracy as well.

We mostly compute the contradiction degree between *uni-dimensional attribute values*. For *multi-dimensional attribute values* we split them back into corresponding uni-dimensional attribute values.

The attribute value contradiction degree function helps the plithogenic aggregation probability operators, and the plithogenic probability inclusion (partial order) relationship to obtain a more accurate result.

The attribute value contradiction degree function is designed in each field where plithogenic probability is used in accordance with the application to solve. If it is ignored, the aggregations still work, but the result may lose accuracy.

Then $(U, a, V, d, c)$ is called a *plithogenic probability*:

where "$U$" is the probability space of all events $E$, "$a$" is a (multi-dimensional in general) attribute with respect to which the chances of occurrences of $E$ are calculated, "$V$" is the range of the attribute's values, "$d$" is the degree of chance-of-occurrence of each event $E \in U$ with respect to each attribute value - and "$d$" stands for "$d_F$" or "$d_{IF}$" or "$d_N$", when dealing with fuzzy degree of occurrence, intuitionistic fuzzy degree of occurrence-nonoccurrence, or neutrosophic degree of occurrence-indeterminacy-nonoccurence respectively of a plithogenic probabilistic event $E$; and "$c$" stands for "$c_F$" or "$c_{IF}$" or "$c_N$", when dealing with fuzzy degree of contradiction (dissimilarity), intuitionistic fuzzy degree of contradiction (dissimilarity), or neutrosophic degree of contradiction (dissimilarity) between attribute values respectively.

The functions $d(\cdot,\cdot)$ and $c(\cdot,\cdot)$ are defined in accordance with the probabilistic applications the experts need to solve.

One uses the notation:

$E\big(d(E,V)\big),$

where $d(E,V) = \{d(E,v), \text{for all } v \in V\}, \forall E \in U.$

## IV.3.6. About the Plithogenic Aggregation Probabilistic Operators

The attribute value contradiction degree is calculated between each attribute value with respect to the dominant attribute value (denoted $v_D$) in special, and with respect to other attribute values as well.

The attribute value contradiction degree function $c$ between the attribute's values is used into the definition of <u>plithogenic aggregation probabilistic operators</u> {*conjunction (AND), disjunction (OR), Negation ($\neg$), Implication ($\Rightarrow$), Equivalence ($\Leftrightarrow$),* and other plithogenic aggregation probabilistic operators that combine two or more attribute value degrees - that $t_{norm}$ and $t_{conorm}$ act upon}.

Several plithogenic aggregation probabilistic operators are linear combinations of the fuzzy $t_{norm}$ *(*denoted $\wedge_F$*),* and fuzzy $t_{conorm}$ *(*denoted $\vee_F$*)*, but non-linear combinations may as well be constructed.

If one applies the $t_{norm}$ on dominant attribute value denoted by $v_D$, and the contradiction between $v_D$ and $v_2$ is $c(v_D, v_2)$, then onto attribute value $v_2$ one applies:

$[1 - c(v_D, v_2)] \cdot t_{norm}(v_D, v_2) + c(v_D, v_2) \cdot t_{conorm}(v_D, v_2)$,  (248)

Or, by using symbols:

$[1 - c(v_D, v_2)] \cdot (v_D \wedge_F v_2) + c(v_D, v_2) \cdot (v_D \vee_F v_2)$.  (249)

Similarly, if one applies the $t_{conorm}$ on dominant attribute value denoted by $v_D$, and the contradiction between $v_D$ and $v_2$ is $c(v_D, v_2)$, then onto attribute value $v_2$ one applies:

$[1 - c(v_D, v_2)] \cdot t_{conorm}(v_D, v_2) + c(v_D, v_2) \cdot t_{norm}(v_D, v_2)$,  (250)

Or, by using symbols:

$$[1 - c(v_D, v_2)] \cdot (v_D \vee_F v_2) + c(v_D, v_2) \cdot (v_D \wedge_F v_2). \qquad (251)$$

## IV.4. One-Attribute-Value Plithogenic Single-Valued Fuzzy Probabilistic Operators

We consider the single-value number degrees of chances of occurrence, for simplicity of the example.

Let $v_i$ be a uni-attribute value and its degree of contradiction with respect to the dominant uni-attribute value $v_D$ be

$c(v_D, v_i) \stackrel{\text{def}}{=} c_i$.

Let $d_A(E, v_i)$ be the degree of occurrence of event $E$ with respect to the attribute-value $v_i$ given by Expert $A$. And similarly for $d_B(E, v_i)$. Then, we recall the plithogenic aggregation probabilistic operators with respect to this attribute value $v_i$ that will be employed:

## IV.4.1. One-Attribute-Value Plithogenic Single-Valued Fuzzy Probabilistic Conjunction

$$\begin{aligned} d_A(E, v_i) \wedge_p d_B(E, v_i) &= (1 - c_i) \cdot [d_A(E, v_i) \wedge_F d_B(E, v_i)] + c_i \\ &\quad \cdot [d_A(E, v_i) \vee_F d_B(E, v_i)] \end{aligned}$$
(252)

## IV.4.2. One-Attribute Value Plithogenic Single-Valued Fuzzy Probabilistic Disjunction

$$d_A(E, v_i) \vee_p d_B(E, v_i) == (1 - c_i) \cdot [d_A(E, v_i) \vee_F d_B(E, v_i)] + c_i \cdot [d_A(E, v_i) \wedge_F d_B(E, v_i)] \qquad (253)$$

## IV.4.3. One Attribute Value Plithogenic Single-Valued Fuzzy Probabilistic Negations

$$\neg_p d_A(E, v_i) = d_A(E, anti(v_i)) = d_A(E, (1 - c_i)v_i) \quad (254)$$

or

$$\neg_p d_A(E, v_i) = 1 - d_A(E, v_i). \quad (255)$$

## IV.5. n-Attribute-Values Plithogenic Single-Valued Probabilistic Operators

The easiest way to apply the plithogenic probabilistic operators on a multi-attribute plithogenic probability is to split back the *m*-dimensional attribute, $m \geq 1$, into *m* uni-dimensional attributes, and separately apply the plithogenic probabilistic operators on the set of all attribute values (needed by the application to solve) of each uni-dimensional attribute.

Therefore, let α be a given uni-dimensional attribute, characterizing each event $E \in U$, whose set of values are:

$$V = \{v_1, v_2, \ldots, v_n\} \equiv \{v_D, v_2, \ldots, v_n\}, n \geq 1, \quad (256)$$

where $v_D$ = dominant attribute value, and

$c(v_D, v_i) = c_i \in [0, 1]$ the contradiction degrees. Without restricting the generality, we consider the uni-dimensional attribute values arranged in an increasing order with respect to their contradiction degrees, i.e.:

$$c(v_D, v_D) = 0 \leq c_1 \leq c_2 \leq \cdots \leq c_{i_0}$$
$$< \frac{1}{2} \leq c_{i_0+1} \leq \cdots \leq c_n \leq 1. \quad (257)$$

## IV.5.1. n-Attribute-Values Plithogenic Single-Valued Fuzzy Probabilistic Operators

Let's consider two experts, A and B, which evaluate a probabilistic event $E$, with respect to the fuzzy degrees of the values $v_1, \ldots, v_n$ representing the chances of occurrence of event $E$, upon some given criteria:

$$d_A^F: U \times V \to [0,1], d_A^F(x, v_i) = a_i \in [0,1], \quad (258)$$
$$d_B^F: U \times V \to [0,1], d_B^F(x, v_i) = b_i \in [0,1], \quad (259)$$

for all $i \in \{1, 2, \ldots, n\}$.

## IV.5.2. n-Attribute-Values Plithogenic Single-Valued Fuzzy Probabilistic Conjunction

$$\left(a_1, a_2, \ldots, a_{i_0}, a_{i_0+1}, \ldots, a_n\right) \wedge_p \left(b_1, b_2, \ldots, b_{i_0}, b_{i_0+1}, \ldots, b_n\right) =$$
$$\begin{pmatrix} a_1 \wedge_p b_1, a_2 \wedge_p b_2, \ldots, a_{i_0} \wedge_p \\ b_{i_0}, a_{i_0+1} \wedge_p b_{i_0+1}, \ldots, a_n \wedge_p b_n \end{pmatrix} \quad (260)$$

The first $i_0$ probabilistic conjunctions are proper plithogenic probabilistic conjunctions (the weights onto the $t_{norm}$'s are bigger than onto $t_{conorm}$'s):

$$a_1 \wedge_p b_1, a_2 \wedge_p b_2, \ldots, a_{i_0} \wedge_p b_{i_0}$$
(261)

whereas the next $n - i_0$ probabilistic conjunctions

$$a_{i_0+1} \wedge_p b_{i_0+1}, \ldots, a_n \wedge_p b_n \quad (262)$$

are improper plithogenic probabilistic disjunctions (since the weights onto the $t_{norm}$'s are less than onto $t_{conorm}$'s):

## IV.5.3. n-Attribute-Values Plithogenic Single-Valued Fuzzy Probabilistic Disjunction

$$(a_1, a_2, \ldots, a_{i_0}, a_{i_0+1}, \ldots, a_n) \vee_p (b_1, b_2, \ldots, b_{i_0}, b_{i_0+1}, \ldots, b_n) = \begin{pmatrix} a_1 \vee_p b_1, a_2 \vee_p b_2, \ldots, a_{i_0} \vee_p \\ b_{i_0}, a_{i_0+1} \vee_p b_{i_0+1}, \ldots, a_n \vee_p b_n \end{pmatrix} \quad (263)$$

The first $i_0$ probabilistic disjunctions are proper plithogenic probabilistic disjunctions (the weights onto the $t_{conorm}$'s are bigger than onto $t_{norm}$'s):

$$a_1 \vee_p b_1, a_2 \vee_p b_2, \ldots, a_{i_0} \vee_p b_{i_0} \quad (264)$$

whereas the next $n - i_0$ probabilistic disjunctions

$$a_{i_0+1} \vee_p b_{i_0+1}, \ldots, a_n \vee_p b_n \quad (265)$$

are improper plithogenic probabilistic conjunctionss (since the weights onto the $t_{conorm}$'s are less than onto $t_{norm}$'s):

## IV.5.4. n-Attribute-Values Plithogenic Single-Valued Fuzzy Probabilistic Negations

In general, for a generic probabilistic event $E$, characterized by the uni-dimensional attribute α, whose attribute values are $V = (v_D, v_2, \ldots, v_n), n \geq 2$, and whose attribute value contradiction degrees (with respect to the dominant attribute value $v_D$) are respectively: $0 \leq c_2 \leq \cdots \leq c_{n-1} \leq c_n \leq 1$, and their attribute value degrees of occurrence of the event $E$ are respectively $a_D, a_2, \ldots, a_{n-1}, a_n \in [0, 1]$, then the plithogenic fuzzy probabilistic negation of $E$ is:

$$\neg_p [\, E \begin{pmatrix} 0 & c_2 & & c_{n-1} & c_n \\ v_D, v_2, & \ldots, & v_{n-1}, & v_n \\ a_D & a_2 & & a_{n-1} & a_n \end{pmatrix} ] =$$

$$\neg_p E \begin{pmatrix} 1-c_n & 1-c_{n-1} & & 1-c_2 & 1-c_D \\ anti(v_n) & anti(v_{n-1}) & \ldots & anti(v_2) & anti(v_D) \\ a_n & a_{n-1} & & a_2 & a_D \end{pmatrix}. \quad (266)$$

Some $anti(V_i)$ may coincide with some $V_j$, whereas other $anti(V_i)$ may fall in between two consecutive $[v_k, v_{k+1}]$ or we may say that they belong to the *Refined* set V;

or

$$= \begin{Bmatrix} v_n & v_{n-1} & \ldots & \ldots & v_1 \\ a_1, & a_2, & \ldots,, & \ldots, & a_n \end{Bmatrix} \quad (267)$$

{this version gives an exact result when the contradiction degrees are equi-distant (for example: *0, 0.25, 0.50, 0.75, 1*) or symmetric with respect to the center *0.5* (for example: *0, 0.4, 0.6, 1*), and an approximate result when they are not equi-distant and not symmetric to the center (for example: *0, 0.3, 0.8, 0.9, 1*);}

or

$$\begin{Bmatrix} v_1 & v_2 & \ldots & v_{i_0} & v_{i_0+1} & \ldots & v_n \\ 1-a_1 & 1-a_2 & \ldots & 1-a_{i_0} & 1-a_{i_0+1} & \ldots & 1-a_n \end{Bmatrix} \quad (268)$$

where $anti(v_i) \in V$ or $anti(v_i) \in RefinedV$, for all $i \in \{1, 2, \ldots, n\}$.

## IV.6. Multi-Attribute Plithogenic General Probability

## IV.6.1. Definition of Multi-Attribute Plithogenic General Probabilistic

Let *U* be a probability space, and a plithogenic probabilistic event $E \in U$.

Let $\mathcal{A}$ be a set of $m \geq 2$ attributes: α₁, α₂, …, αₘ, whose corresponding spectra of values are the non-empty sets $S_1$, $S_2$, …, $S_m$ respectively.

Let $V_1 \subseteq S_1$, $V_2 \subseteq S_2$, …, $V_m \subseteq S_m$ be subsets of attribute values of the attributes α₁, α₂, …, αₘ respectively needed by experts in their given probabilistic application.

For each $j \in \{1, 2, ..., m\}$, the *set of attribute values $V_j$* means the *range of attribute α_j's values*, needed by the experts in a specific application or in a specific problem to solve.

Each probabilistic event $E \in U$ is characterized by all $m$ attributes.

Let the *m-dimensional attribute value degree of chance of occurrence function* be:

$$d_{[m]}: (U, V_1 \times V_2 \times ... \times V_m) \to \mathcal{P}([0, 1])^m. \quad (269)$$

For any $E \in U$, and any $v_j \in V_j$ with $j \in \{1, 2, ..., m\}$, one has:

$$d_{[m]}\big(E(v_1, v_2, ..., v_m)\big) \subseteq \mathcal{P}([0, 1])^m. \quad (270)$$

## IV.6.2. Example of Plithogenic Probabilistic

What is the plithogenic probability that Jenifer will graduate at the end of this semester in her program of electrical engineer, given that she is enrolled in and has to pass two courses of *Mathematics* (*Second-Order Differential Equations*, and *Stochastic Analysis*), and two courses of *Mechanics* (*Fluid Mechanics,* and *Solid Mechanics)* ?

We have a 4-attribute values plithogenic probability.

### IV.6.2.1. Plithogenic Fuzzy Probability

According to her adviser, Jenifer's *plithogenic single-valued fuzzy probability* of graduating at the end of this semester is:

*J*( 0.5, 0.6;  0.8, 0.4 ),

which means *50%* chance of passing the Second-Order Differential Equations class, *60%* chance of passing the Stochastic Analysis class (as part of Mathematics), and *80%* of passing the Fluid Mechanics class and *40%* of passing the Solid Mechanics class (as part of Physics).

Using a larger approximation (less accuracy), the adviser predicts that Jenifer's *plithogenic interval-valued fuzzy probability* of graduating at the end of this semester is:

*J*( [0.4, 0.6], [0.3, 0.7];  [0.8, 0.9], [0.2, 0.5] ),

which means that between 40%-50% are Jenifer's chances to pass the class of Second-Order Differential Equations; and similarly for the other three classes.

### IV.6.2.2. Plithogenic Intuitionistic Fuzzy Probability

Jenifer's *plithogenic single-valued intuitionistic fuzzy probability* of graduating at the end of this semester is:

*J*( (0.5, 0.2), (0.6, 0.4);  (0.8, 0.1), (0.4, 0.5) ),

which mean that 50% is chance that Jenifer passes the Second-Order Differential Equations class and 20% chance that she fails this class; and similarly for the other three classes.

Jenifer's *plithogenic interval-valued intuitionistic fuzzy probability* of graduating at the end of this semester is:

*J*( ([0.5, 0.6], [0.1, 0.2]), ([0.6, 0.8], [0.2,0.4]);  ([0.8, 0.9], [0.0, 0.1]), ([0.3, 0.6], [0.3, 0.5]) ),

which mean that between 50% - 60% is chance that Jenifer passes the Second-Order Differential Equations class and between 10% - 20% is the chance that she fails this class; and similarly for the other three classes.

### IV.6.2.3. Plithogenic Neutrosophic Probability

*Jenifer's plithogenic single-valued neutrosophic probability* of graduating at the end of this semester is:

$J((0.5, 0.1, 0.2), (0.6, 0.2, 0.4);\ (0.8, 0.0, 0.1), (0.4, 0.3, 0.5))$,

which mean that there is 50% chance, 10% indeterminate-chance, and 20% nonchance that Jenifer passes the class of Second-Order Differential Equation; and similarly for the other three classes.

*Jenifer's plithogenic interval-valued neutrosophic probability* of graduating at the end of this semester is:

J( ([0.5, 0.6], [0.0, 0.1], [0.2., 0.4]), ([0.6, 0.8], [0.1, 0.2], [0.3, 0.5]); ([0.8, 0.9], [0.0, 0.2], [0.1, 0.3]), (0.4, 0.3, 0.5) ),

which mean that there is between 50% - 60% chance, between 0% - 10% indeterminate-chance, and between 20% - 405 nonchance that Jenifer passes the class of Second-Order Differential Equation; and similarly for the other three classes.

## IV.6.3. Plithogenic Probability as Probability of Probabilities

Plithogenic probability is a probability of (classical, imprecise, intuitionistic fuzzy, or neutrosophic) probabilities – depending on the choice of the chance function. Or plithogenic probability is a *refined probability*.

# V. PLITHOGENIC STATISTICS

As a generalization of classical statistics and neutrosophic statistics, the *Plithogenic Statistics* is the analysis of events described by the plithogenic probability.

Since in plithogenic probability each event $E$ from a probability space $U$ is characterized by <u>many chances</u> of the event to occur [<u>not only one chance</u> of the event $E$ to occur: as in classical probability, imprecise probability, and neutrosophic probability], a **plithogenic probability distribution function**, $PP(x)$, of a random variable $x$, is described by <u>many</u> *plithogenic probability distribution <u>sub-functions</u>*, where each sub-function represents the chance (with respect to a given attribute value) that value $x$ occurs, and these chances of occurrence can be represented by classical, imprecise, or neutrosophic probabilities (depending on the type of degree of a chance).

[More study is to be done in this subject…]

## Future Research

As generalization of dialectics and neutrosophy, plithogeny will find more use in blending diverse philosophical, ideological, religious, political and social ideas.

After the extension of fuzzy set, intuitionistic fuzzy set, and neutrosophic set to the plithogenic set;

the extension of classical logic, fuzzy logic, intuitionistic fuzzy logic and neutrosophic logic to plithogenic logic;

and the extension of classical probability, imprecise probability, and neutrosophic probability to plithogenic probability – more applications of the plithogenic set/logic/probability/statistics in various fields should follow.

The classes of *plithogenic implication operators* and their corresponding sets of *plithogenic rules* are to be constructed in this direction.

Also, exploration of <u>non-linear combinations</u> of $t_{norm}$ and $t_{conorm}$, or of other norms and conorms, in constructing of more *sophisticated plithogenic set, logic* and *probabilistic aggregation operators*, for a better modeling of real life applications.

More study, development, and utilizations should be done and proved into the field of *plithogenic statistics*.

whole issue of this journal is dedicated to Neutrosophy and Neutrosophic Logic.

11. F. Smarandache, "Definitions Derived from Neutrosophics", <Multiple Valued Logic / An International Journal>, USA, ISSN 1023-6627, Vol. 8, No. 5-6, pp. 591-604, 2002.

In this book we introduce for the first time, as generalization of dialectics and neutrosophy, the philosophical concept called *plithogeny*. And as its derivatives: the *plithogenic set* (as generalization of crisp, fuzzy, intuitionistic fuzzy, and neutrosophic sets), *plithogenic logic* (as generalization of classical, fuzzy, intuitionistic fuzzy, and neutrosophic logics), *plithogenic probability* (as generalization of classical, imprecise, and neutrosophic probabilities), and *plithogenic statistics* (as generalization of classical, and neutrosophic statistics).

Plithogeny is the genesis or origination, creation, formation, development, and evolution of *new entities* from dynamics and organic fusions of contradictory and/or neutrals and/or non-contradictory *multiple old entities*.

Plithogenic Set is a set whose elements are characterized by one or more attributes, and each attribute may have many values.

An attribute's value $v$ has a corresponding (fuzzy, intuitionistic fuzzy, or neutrosophic) *degree of appurtenance $d(x, v)$* of the element $x$, to the set $P$, with respect to some given criteria.

In order to obtain a better accuracy for the plithogenic aggregation operators in the plithogenic set/logic/probability and for a more exact inclusion (partial order), a (fuzzy, intuitionistic fuzzy, or neutrosophic) *contradiction (dissimilarity) degree* is defined between each attribute value and the dominant (most important) attribute value.

The *plithogenic intersection and union* are linear combinations of the fuzzy operators $t_{norm}$ and $t_{conorm}$, while the *plithogenic complement/inclusion/equality* are influenced by the attribute values' contradiction (dissimilarity) degrees.

Formal definitions of plithogenic set/logic/probability/statistics are presented into the book, followed by plithogenic aggregation operators, various theorems related to them, and afterwards examples and applications of these new concepts in our everyday life.

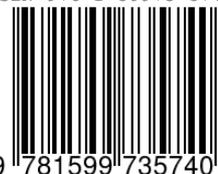

ISBN 978-1-59973-574-0

9 781599 735740